\documentclass[10pt,twocolumn,letterpaper]{article}

\usepackage[pagenumbers]{cvpr} % To force page numbers, e.g. for an arXiv version

\usepackage{times}
\usepackage{amsmath,amssymb}
\usepackage{amsthm}            % after amsmath

\usepackage{graphicx}
\usepackage{subcaption}
\usepackage{xcolor}
\usepackage{colortbl}          % after xcolor
\usepackage{booktabs}
\usepackage{multirow}
\usepackage{cuted}
\usepackage{capt-of}
\usepackage{tikz}
\usetikzlibrary{arrows.meta,positioning,calc}

\usepackage{enumitem}
\usepackage{algorithm}
\usepackage{algpseudocode}
\usepackage[most]{tcolorbox}
\usepackage{comment}
\usepackage{bibunits}
\usepackage{tcolorbox}
\usepackage{tabularx}

\definecolor{bestgreen}{RGB}{226,247,222}
\definecolor{refgray}{gray}{0.94}
\definecolor{ourteal}{HTML}{1D9E75}
\definecolor{trainpurple}{HTML}{534AB7}
\definecolor{trainfill}{HTML}{EEEDFE}
\definecolor{datagray}{HTML}{B4B2A9}
\definecolor{framegray}{HTML}{888780}
\definecolor{thmred}{HTML}{7B1113}
\definecolor{propblue}{HTML}{1A2A80}

\DeclareMathOperator*{\argmin}{arg\,min}
\newcommand{\R}{\mathbb{R}}
\newcommand{\E}{\mathbb{E}}
\newcommand{\g}{\mathbf{g}}
\newcommand{\M}{\mathcal{M}}
\newcommand{\dd}{\,\mathrm d}
\newcommand{\Lcal}{\mathcal{L}}
\newcommand{\best}[1]{\cellcolor{bestgreen}\textbf{\boldmath$#1$}}

\newtheorem{theorem}{Theorem}
\newtheorem{proposition}{Proposition}
\newtheorem{lemma}{Lemma}
\newtheorem{corollary}{Corollary}
\newtheorem{definition}{Definition}

\tcbset{
  mfmbox/.style={
    enhanced, breakable,
    colback=white, colframe=#1, coltitle=white,
    boxrule=0.5pt, arc=1.5pt, outer arc=1.5pt,
    left=4pt, right=4pt, top=2pt, bottom=2pt,
    toptitle=1pt, bottomtitle=1pt,
    fonttitle=\bfseries\small\raggedright, fontupper=\normalsize\itshape,
    before skip=6pt, after skip=6pt,
  },
}
\newtcolorbox[use counter=theorem]{mfmtheorem}[2]{mfmbox=thmred,
  title={Theorem~\thetcbcounter: #1}, label={thm:#2}}
\newtcolorbox[use counter=proposition]{mfmproposition}[2]{mfmbox=propblue,
  title={Proposition~\thetcbcounter: #1}, label={prop:#2}}

\extrafloats{100}

\definecolor{refgray}{gray}{0.94}
\usepackage{amsmath,amssymb,amsthm}
\usepackage{placeins}
\usepackage{adjustbox} % max width avoids unreadable table scaling
\DeclareMathOperator{\supp}{supp}
\DeclareMathOperator{\Var}{Var}
\DeclareMathOperator{\Cov}{Cov}
\DeclareMathOperator{\Cut}{Cut}

\newcommand{\Lip}{\ell}
\newcommand{\F}{\mathcal{F}}

\newtheorem{remark}{Remark}
\makeatletter
\@ifundefined{proposition}{\newtheorem{proposition}{Proposition}}{}
\makeatother

\renewcommand{\thetable}{S\arabic{table}}   % tables S1, S2, ... (cited in the main paper)
\renewcommand{\thefigure}{S\arabic{figure}} % figures S1, S2, ...

\extrafloats{100}

\definecolor{cvprblue}{rgb}{0.21,0.49,0.74}
\usepackage[pagebackref,breaklinks,colorlinks,allcolors=cvprblue]{hyperref}

\def\confName{CVPR}
\def\confYear{2027}

\title{Morphological Flow Matching: One-Step Generation via Hopf-Lax Formula in Riemannian Manifolds}

\author{Thierno Fall$^{1, 2}$, El Hadji S. Diop$^{1,2}$,  Mohamed Daoudi$^{3,4}$ \\
$^{1}$NAGIP-Nonlinear Analysis and Geometric Information Processing Research Group\\
$^{2}$Department of Mathematics, University Iba Der Thiam, BP $967$, Thies, Senegal\\
$^{3}$Univ. Lille, CNRS, Centrale Lille, Institut Mines-Télécom, UMR 9189 CRIStAL, F-59000 Lille, France\\
$^{4}$IMT Nord Europe, Institut Mines-Télécom, Univ. Lille, Centre for Digital Systems, F-59000 Lille, France\\
{\tt\small thiernofall571@gmail.com, elhadjisamba.diop@univ-thies.sn,}\\{\tt\small 
 mohamed.daoudi@imt-nord-europe.fr}
}
\begin{document}
%\maketitle
\twocolumn[{%
\renewcommand\twocolumn[1][]{#1}%
\maketitle
\vspace{-1.5em}
\begin{center}
\includegraphics[width=0.8\textwidth]{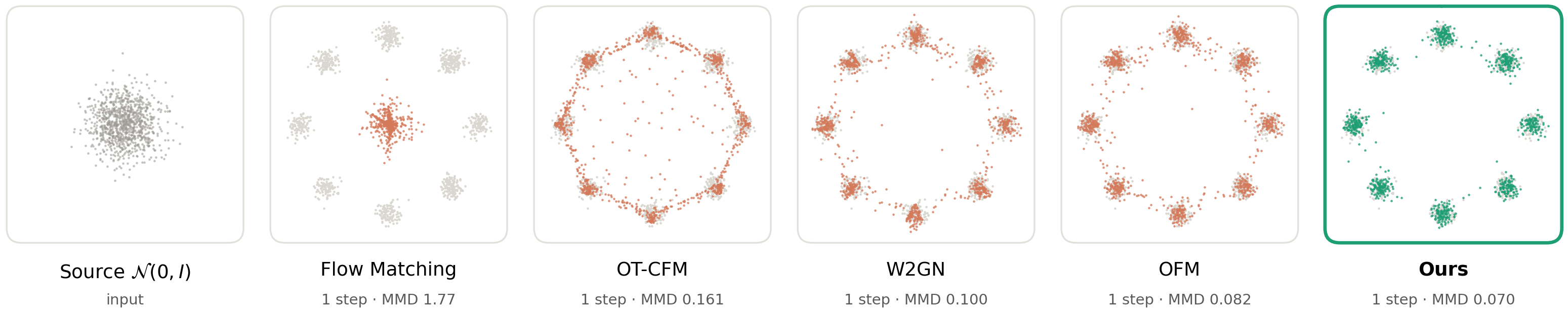}
\captionof{figure}{\textbf{One-step generation on 8 Gaussians.} All methods use a single network evaluation (NFE~$=1$); data are shown in gray (samples generated with seed 0; MMD averaged over 5 seeds). FM collapses to the center, and OT-CFM, W2GN and OFM, whose one-step maps are continuous, connect the modes with bridges of points. MorphoFM, driven by the erosion of a ReLU potential whose gradient can jump, separates the modes in a single step and reaches the sampling-noise floor (MMD 0.070 vs.\ 0.067 between two data samples).}
\label{fig:teaser}
\end{center}
\vspace{0.5em}
}]

\begin{abstract}
Flow matching (FM) samples by integrating a learned velocity field over many steps, and a continuous one-step map
cannot separate disconnected modes. We introduce \emph{morphological flow matching} (MorphoFM), which defines the
velocity field by the morphological erosion of a learned scalar potential, given by the Hopf--Lax formula in Euclidean
spaces and on Riemannian manifolds. At the optimal potential, trajectories are constant-speed geodesics, so one step is
exact. Since the conditional flow matching loss cannot learn such a potential in general, we derive a loss from an
optimality gap of the erosion: its terminal term is, up to sign, the semi-dual Kantorovich objective, and its
minimizer cancels the flow matching residual on the optimal coupling without computing that coupling. ReLU potentials,
whose gradients can jump, separate disconnected modes in one step, and batch candidates with memorized extremal points
keep the inner problem tractable. On synthetic and spherical Earth data, MorphoFM matches the data reference in one
step; on CIFAR-10, the same algorithm scales to images, with a one-step FID of $31.3$ versus $231.5$ for OT-CFM and
$350.6$ for FM.
\end{abstract}    

\section{Introduction}
\label{sec:intro}

Flow matching (FM)~\cite{lipman2023flow,liu2023flow} has become a standard approach to generative modeling: a velocity
field is learned by regression, and samples are generated by integrating an ODE. Curved trajectories, however,
require many network evaluations. Existing remedies either straighten the trajectories, through
reflow~\cite{liu2023flow}, optimal transport couplings (OT-CFM~\cite{tong2024improving}) or convex potentials
(OFM~\cite{kornilov2024optimal}), or learn direct jumps along them (consistency models~\cite{song2023consistency},
shortcut models~\cite{frans2025shortcut}, MeanFlow~\cite{geng2025mean}). Yet any \emph{continuous} one-step generator
faces a topological obstruction: since a continuous map sends a connected set onto a connected set, it cannot push a
Gaussian onto a distribution with disconnected support, and approximating one requires a large Lipschitz
constant~\cite{salmona2022pushforward}. Fig.~\ref{fig:teaser} shows the consequence on a mixture of eight Gaussians:
the one-step samples of OT-CFM, and those of W2GN~\cite{korotin2021wasserstein} and OFM, whose smooth convex
potentials have continuous gradients, connect the modes with bridges of points, \ie, they put mass where the data has
none. Separating the modes in a single step, therefore, requires a \emph{discontinuous} transport map.

Multiscale morphological erosions can be computed using partial differential equations
(PDEs)~\cite{brockett1994evolution,boomgaard1994morphological}. These PDEs belong to the family of first-order
Hamilton--Jacobi (HJ) equations, whose viscosity solutions~\cite{Fleming2006,Barles2008} are explicitly given by the
Hopf--Lax formula. We propose \emph{Morphological Flow Matching} (MorphoFM) to overcome the topological obstruction.
MorphoFM reverses the usual viewpoint: instead of eroding a given function, we \emph{learn} a scalar potential
$f_\theta$ and define the generative velocity field through its Hopf--Lax erosion, using a generalized family of
structuring functions in Euclidean spaces and on Riemannian manifolds.

At the optimal potential, optimal transport theory~\cite{gangbo1996,mccann2001,fathi2010} guarantees that trajectories
are geodesics traversed at constant speed, so a single step is exact. Moreover, ReLU potentials have discontinuous
gradients: their one-step map separates the modes (Fig.~\ref{fig:teaser}, right), which no smooth potential can do.
Learning the potential requires a dedicated objective, since the conditional flow matching (CFM) loss, applied directly
to the morphological field, cannot learn it in general. We instead measure the CFM residual through an optimality gap
of the erosion, whose terminal term is, up to sign, the semi-dual Kantorovich objective~\cite{villani2009}. Minimizing
it under the independent coupling cancels this residual on the optimal coupling, without computing that coupling.

Experiments on synthetic distributions in $\R^2$ and geographic data on $\mathbb S^2$ show that a single step reaches
the sampling-noise floor or the training-data reference; on the sphere, sampling is about 50 times faster than
Riemannian flow matching~\cite{chen2024flow} with 100 steps. On CIFAR-10, the same algorithm scales to images: its
one-step FID is $31.3$, versus $231.5$ for OT-CFM and $350.6$ for FM. Our main contributions are:
\begin{itemize}
\item \textbf{A morphological flow and its training loss.} A velocity field defined by the extremal points of a
Hopf--Lax erosion, and a loss derived from an erosion optimality gap, which links the CFM residual to the classical
semi-dual Kantorovich objective.
\item \textbf{Non-convex ReLU potentials.} Potentials with discontinuous gradients, whose one-step map separates
disconnected modes.
\item \textbf{A training algorithm for non-convex potentials.} Batch candidates that bound the estimated loss from
below, memorized extremal points and Danskin gradients (no Hessian inversion).
\item \textbf{One-step results} at the reference level on synthetic and spherical data, and a scaling test on
CIFAR-10.
\end{itemize}

\section{Related Work}
\label{sec:related}

\noindent
\textbf{Flow matching and one-step generation.}
Flow matching~\cite{lipman2023flow,liu2023flow} learns velocity fields integrated through ODEs. Straightening
strategies include reflow~\cite{liu2023flow} and optimal transport
couplings~\cite{tong2024improving,pooladian2023multisample}, while consistency models~\cite{song2023consistency},
shortcut models~\cite{frans2025shortcut} and MeanFlow~\cite{geng2025mean} enable few- or one-step generation. Action
Matching~\cite{neklyudov2023action} and OT-Flow~\cite{onken2021otflow} use scalar potentials but still require
numerical integration. Optimal Flow Matching (OFM)~\cite{kornilov2024optimal} uses quadratic optimal transport and a
convex ICNN potential to obtain straight trajectories, at the cost of an inner inversion problem and a Hessian
inversion in the gradient. MorphoFM instead learns a single time-independent scalar potential, whose erosion defines
the velocity and whose optimum yields straight trajectories, hence one-step generation; unlike OFM's, this potential
is non-convex, its gradient can jump, and training needs no Hessian inversion (Prop.~\ref{prop:danskin}).

\noindent
\textbf{Neural optimal transport and Hamilton--Jacobi.}
Neural transport methods learn Kantorovich potentials through semi-dual
objectives~\cite{seguy2018large,taghvaei2019wasserstein,makkuva2020optimal}, with convex
potentials~\cite{huang2021convex,korotin2021wasserstein} or adversarially trained
generators~\cite{rout2022generative,choi2023generative}, which can reach low one-step FID on
images~\cite{choi2023generative} but replace the $c$-transform with a second network trained by a min-max game; the
$c$-transform can also be approximated by amortized prediction and refinement~\cite{amos2023amortizing}. Neural
Hamilton--Jacobi Characteristic Flows (NCF)~\cite{park2026ncf} learn a space-time HJ solution through characteristic consistency and distribution matching, enabling transport without ODE integration. Closest to our training procedure, Park
et al.~\cite{park2026implicitot} learn a single Kantorovich potential whose $c$-transform is a proximal problem solved
by fixed-point iterations, with gradients that do not differentiate through the solver. MorphoFM shares this classical
semi-dual foundation~\cite{villani2009} and also avoids differentiating through the solver
(Prop.~\ref{prop:danskin}), but differs in three ways: its time-independent potential defines, through its erosion, a
velocity field whose optimality gap measures the CFM residual; its costs are powers of the geodesic distance on
Riemannian manifolds, whereas these methods use the quadratic Euclidean cost; and its non-convex inner problem is
solved from batch candidates that keep the estimated loss bounded from below (Prop.~\ref{prop:borne}).

\noindent
\textbf{Generative models on manifolds.}
Riemannian CNFs~\cite{mathieu2020riemannian}, Moser Flow~\cite{rozen2021moser}, CNF
Matching~\cite{benhamu2022matching}, Riemannian diffusion models~\cite{debortoli2022riemannian,huang2022riemannian}
and Riemannian Flow Matching~\cite{chen2024flow} rely on multi-step integration. Riemannian convex potential
maps~\cite{cohen2021riemannian} instead build transport maps from the exponential map of $c$-concave
potentials~\cite{mccann2001}, trained as normalizing flows by maximum likelihood. MorphoFM also generates with the
exponential map, without ODE integration, but learns a neural potential from samples with the semi-dual objective,
without Jacobian determinants.

\noindent
\textbf{Mathematical morphology and transport regularity.}
Multiscale morphology is closely related to Hamilton--Jacobi
equations~\cite{brockett1994evolution,boomgaard1994morphological} and to Legendre--Fenchel
duality~\cite{dorst1994slope,heijmans1997slope}; it has been extended to Riemannian manifolds~\cite{diop2021extension}
and used in network layers~\cite{smets2023pde}, including in a GAN generator~\cite{diop2024gmgan}. MorphoFM uses the
erosion of a learned potential not as a layer but as the transport mechanism itself; to our knowledge, erosion has not
been used before to define a generative flow. The topological obstruction to continuous
generators~\cite{dupont2019augmented,cornish2020relaxing,salmona2022pushforward} concerns continuity, not convexity:
convex potentials with smooth activations, as in W2GN and OFM, yield continuous maps. 
MorphoFM instead uses ReLU potentials, which admit
a tropical rational representation~\cite{zhang2018tropical},
closely related to the max-plus algebra underlying
mathematical morphology. Their gradients can be
discontinuous, enabling one-step mode separation
(Fig.~\ref{fig:teaser}).

\section{Morphological Flow Matching (MorphoFM)}\label{sec:mfm}
%================================================================

\begin{figure}[!t]
\centering
\includegraphics[width=\linewidth]{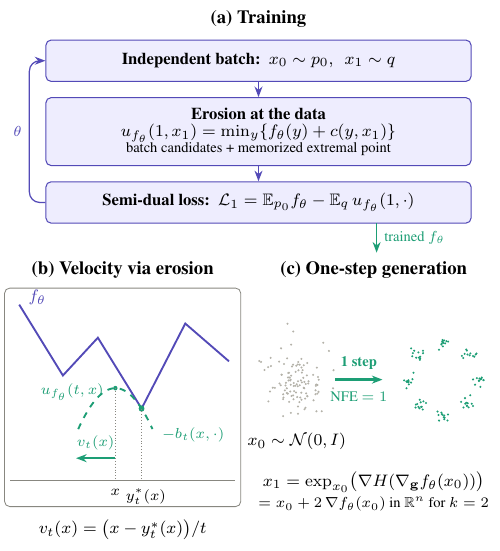}
\caption{\textbf{MorphoFM overview.}
(a) Training a scalar potential through the semi-dual
loss, with batch-based initialization and local
refinement of extremal points.
(b) Morphological velocity derived from the
extremal point of the Hopf--Lax erosion.
(c) Direct one-step generation from the learned
potential, without ODE integration or inner optimization.}
\label{fig:overview}
\end{figure}

\noindent
\textbf{Classical flow matching.}
Flow matching (FM)~\cite{lipman2023flow,liu2023flow} trains a network $v_\theta:\R^n\times[0,1]\to\R^n$ and
generates a sample by integrating the ODE $\dot x_t=v_\theta(x_t,t)$ from $x_0\sim p_0$ with a numerical solver
(Euler, Runge--Kutta); the cost of generation is the number of steps times the cost of one network evaluation. Since
source and target points are drawn independently, the straight conditional paths cross with conflicting velocities;
the learned marginal field averages them, potentially producing curved trajectories, and accurate generation requires many steps.
MorphoFM instead learns a \emph{single scalar potential} $f_\theta:\M\to\R$, defines the velocity field \emph{from the
morphological erosion} of $f_\theta$, and trains it with a Hopf--Lax objective tied to optimal transport. At the
optimal potential, trajectories are straight (geodesics on a manifold), so a single step generates exactly, without
ODE integration (Fig.~\ref{fig:overview}).

\noindent
\textbf{Setting.}
$(\M,\g)$ is a connected Riemannian manifold, compact or equal to $\R^n$, endowed with the metric $\g$. $\nabla_\g$
denotes the gradient under $\g$ and $\exp$ is the exponential map. For $k>1$, $k^\ast=k/(k-1)$, and
$c_k=(k-1)/k^{k^\ast}$. $H:T^\star\!\M\longrightarrow\R$ is a Hamiltonian defined in the cotangent bundle $T^\star\!\M$
of $\M$, which we identify with the tangent bundle through $\g$. For a convex and superlinear $H$, the solution of the
HJ equation
\begin{equation}\label{eq:HJ}
\partial_t u+H(\nabla_\g u)=0,\qquad u(0,\cdot)=f
\end{equation}
is given in the viscosity sense by the Hopf--Lax formula~\cite{evans2010pde,diop2021extension}. Setting
$H(p)=\|p\|_\g^k$, with gradient $\nabla H(p)=k\|p\|_\g^{k-2}p$, establishes a link between the Hopf--Lax formula and
the multiscale \emph{morphological erosion} of $f$ at scale $t$ using the family $(b_t)_{t>0}$ of structuring functions
(SFs) $b_t(x,y)=c_k\,d_\g(x,y)^{k^\ast}/t^{1/(k-1)}$ defined on $\M\times\M$:
\begin{equation}\label{eq:HLO}
u_f(t,x)=\min_{y\in \M}\bigl\{f(y)+b_t(x,y)\bigr\},
\end{equation}
where $d_\g(x,y)$ is the geodesic distance between points $x$ and $y$ in $\M$. We define
$Y_t(x)=\argmin_{y\in\M}\{f(y)+b_t(x,y)\}$ as the set of \emph{extremal points} of $x$ at time $t$. Extremal points
exist if $\M$ is compact and $f$ is continuous, or if $\M=\R^n$ and $f$ is Lipschitz; when there is only one, we denote it by $y^\ast_t(x)$. The
transport cost is $c=b_1=c_k\,d_\g^{k^\ast}$, and the one-step map of $f$ is $T_f(x)=\exp_x(\nabla H(\nabla_\g f(x)))$.
For $k=2$, $b_t=d_\g^2/(4t)$, $c=d_\g^2/4$ and $T_f(x)=x+2\nabla f(x)$ in $\R^n$. 
The construction applies to every $k>1$ in
$\mathbb{R}^n$ and on compact Riemannian
manifolds, while exact one-step generation
is guaranteed at the optimal potential $f^*$
(Thm.~\ref{thm:gen}). The proofs
are in the supplementary material. Experiments use $k=2$; in our applications, the ablation on $k$ shows no measurable
effect on generation (see the supplementary material).

The rest of this section proceeds in four steps: we define the morphological velocity (Sec.~\ref{sec:vit}),
derive the loss (Sec.~\ref{sec:pot}) and its training (Sec.~\ref{sec:num}), and show that one-step generation is exact
(Sec.~\ref{sec:gen}); Algorithm~\ref{alg} summarizes the method.

%-------------------------------------------------------------------
\subsection{Morphological Velocity Field}\label{sec:vit}
%-------------------------------------------------------------------

\begin{definition}[Morphological velocity]\label{def:vit}
For $t>0$, if the extremal point $y^\ast_t(x)$ is unique and joined to $x$ by a unique minimizing geodesic, the
morphological velocity is $v_t(x)=-\frac1t\exp_x^{-1}\!\big(y^\ast_t(x)\big)$, \ie, $\big(x-y^\ast_t(x)\big)/t$ in
$\R^n$ (Fig.~\ref{fig:overview}(b)).
\end{definition}

\begin{proposition}[Velocity is the gradient of the erosion]\label{prop:vit}
Let $t>0$, with $\M$ compact and $f$ continuous, or $\M=\R^n$ and $f$ Lipschitz. Then $u_f(t,\cdot)$ is Lipschitz, hence differentiable almost everywhere. At every point $x$ where
it is differentiable, the extremal point $y^\ast_t(x)$ is unique, joined to $x$ by a unique minimizing geodesic, and
\[
\begin{aligned}
v_t(x)=\nabla H(p)=k\,\|p\|_\g^{k-2}\,p,\qquad p=\nabla_\g u_f(t,x),
\end{aligned}
\]
\ie, $v_t=2\nabla_\g u_f(t,\cdot)$ for $k=2$. (In $\R^n$, this is the classical characterization of the Hopf--Lax
formula~\cite{evans2010pde,cannarsa2004semiconcave}; the Riemannian case is proved in the supplementary material.)
\end{proposition}

\emph{In words,} the velocity is the characteristic velocity of the HJ equation~\eqref{eq:HJ}, \ie, the
velocity with which the characteristic issued from $y^\ast_t(x)$, a geodesic traversed at constant speed, reaches $x$ at time $t$. The initial
field $\nabla H(\nabla_\g f)$ of the one-step map $T_f$ has the same form, with $f$ in place of $u_f(t,\cdot)$; the
morphological velocity itself is defined only for $t>0$.

\noindent
\textbf{Why not the CFM loss?}
It asks the source to be the extremal point of the interpolated point, which the independent coupling leaves
ambiguous; above all, in general no morphological flow with straight, non-crossing trajectories generates its marginal path (see the
supplementary material).

%-------------------------------------------------------------------
\subsection{What We Minimize, and Why}\label{sec:pot}
%-------------------------------------------------------------------

For $(x_0,x_1)$ joined by a unique minimizing geodesic, let $x_t=\exp_{x_0}\!\big(t\exp_{x_0}^{-1}(x_1)\big)$ denote
the geodesic interpolation, with velocity $\dot x_t$.
\vspace{-3mm}
\begin{lemma}[The erosion gap measures the CFM residual]\label{lem:gap}
For $t\in(0,1]$, $G_t(f;x_0,x_1)=f(x_0)+t\,c(x_0,x_1)-u_f(t,x_t)\ge0$, with equality if and only if $x_0$ is an
extremal point of $x_t$. If $x_t$ has a unique extremal point, $G_t=0$ is equivalent to $v_t(x_t)=\dot x_t$.
\end{lemma}
\vspace{-3mm}
\emph{In words,} $G_t$ vanishes exactly when the morphological velocity matches the flow matching target, but it
involves $f$ directly, through its values and its erosion. For any coupling of $p_0$ and $q$, its mean at time $1$ is,
up to a constant, the loss we minimize:
\begin{equation}\label{eq:loss}
\mathcal L_1(f)=\E_{x_0\sim p_0}f(x_0)-\E_{x_1\sim q}\,u_f(1,x_1).
\end{equation}
Since $u_f(1,\cdot)$ is the $c$-transform of $f$~\cite{villani2009}, $-\mathcal L_1(f)$ is the semi-dual Kantorovich objective.
Averaging $G_t$ over $t<1$ would give other losses, but their optimal potential then depends on the coupling, and
under the independent coupling generation becomes overly concentrated (see the supplementary material): only the
$t=1$ term does not depend on the coupling.

\begin{theorem}[The loss is the Kantorovich semi-dual, up to sign]\label{thm:dual}
Assume that $\M$ is compact, or that $\M=\R^n$ and $p_0,q$ have finite moments of order $k^\ast$, and let
$W_c(p_0,q)=\min_{\pi\in\Pi(p_0,q)}\int c\dd\pi$. For every $f:\M\to\R\cup\{+\infty\}$ with $f\in L^1(p_0)$ and
$u_f(1,\cdot)\in L^1(q)$, $\mathcal L_1(f)\ge-W_c(p_0,q)$, and this lower bound is attained by a $c$-convex function
$f^\ast$, \ie, one of the form $f^\ast(x)=\sup_y\{\zeta(y)-c(x,y)\}$. The loss depends only on the marginal
distributions, not on the coupling used to draw $(x_0,x_1)$. (Classical result~\cite[Thm.~5.10]{villani2009}; exact
class of potentials: supplementary material.)
\end{theorem}
\vspace{-3mm}
\emph{In words,} training on independent pairs is enough, and the optimal coupling is recovered without being
computed:

\begin{corollary}[The optimal coupling comes for free]\label{cor:otimplicite}
For every optimal plan $\pi^\ast$, $G_1(f^\ast;x_0,x_1)=0$ for $\pi^\ast$-almost every $(x_0,x_1)$: $x_0$ is an extremal
point of $x_1$. (Classical consequence of duality~\cite[Thm.~5.10(ii)]{villani2009}.)
\end{corollary}

With Lemma~\ref{lem:gap}, the minimizer of $\mathcal L_1$, learned under the independent coupling, therefore cancels
the CFM residual at $t=1$ on the optimal coupling, whereas OT-CFM~\cite{tong2024improving} approximates this coupling
with minibatch solvers. For the quadratic cost and convex potentials, OFM~\cite{kornilov2024optimal} similarly relates
its loss to the Kantorovich dual.

%-------------------------------------------------------------------
\subsection{Training in Practice}\label{sec:num}
%-------------------------------------------------------------------

The term $u_{f_\theta}(1,x_1)=\min_yJ(y)$, with $J(y)=f_\theta(y)+c(y,x_1)$, has no closed form; we compute it at each
iteration for every $x_1$ in the batch (Fig.~\ref{fig:overview}(a)). An approximate point $\hat y$ gives $J(\hat y)\ge u_{f_\theta}(1,x_1)$, hence
an estimated loss $\hat{\mathcal L}_1$ below $\mathcal L_1(f_\theta)$ in expectation, which may collapse where the
search fails.

\begin{proposition}[Batch candidates bound the estimated loss]\label{prop:borne}
Let $B$ be a batch of $|B|$ pairs, with sources $(x_0^i)_{1\le i\le|B|}\sim p_0$ and targets
$(x_1^j)_{1\le j\le|B|}\sim q$; let $J_j(y)=f_\theta(y)+c(y,x_1^j)$, and let $\hat y_j$ be the best point found among
candidates that include all $x_0^i$. Then
$\hat{\mathcal L}_1=\frac1{|B|}\sum_if_\theta(x_0^i)-\frac1{|B|}\sum_jJ_j(\hat y_j)\ge-W_c(\hat p_0,\hat q)$,
where $\hat p_0,\hat q$ are the empirical distributions of the batch.
\end{proposition}

\emph{In words,} the bound does not depend on $\theta$; it does not control the bias, but it prevents the estimate
from collapsing, whatever the quality of the search.
\\
\noindent
\textbf{Solving the inner problem.}
Since the inner problem is guaranteed to be convex only at the optimal potential (for $k=2$ in $\R^n$, Brenier's
condition), we start from the best of several candidates: $x_1$, the best point of the batch and, when available, the
extremal point memorized for the same data point at a previous pass. We then run $K$ steps of a local solver, keeping
the best point: L-BFGS with Armijo line search in $\R^n$ ($K=5$), Riemannian gradient descent on $\mathbb S^2$
($K=10$). An amortized predictor of the extremal point~\cite{amos2023amortizing} can also serve as a candidate; its
ablation shows no measurable effect (see the supplementary material), and it becomes unnecessary once extremal points
are memorized. For
$k=2$ and a ReLU MLP, a Newton step is the fixed-point iteration $y\leftarrow x_1-2\nabla f_\theta(y)$, which inverts
$T_{f_\theta}$; on CIFAR-10, L-BFGS reaches a better minimum, twice as fast (see the supplementary material).
\\
\indent Computing the gradient of $\hat{\mathcal L}_1$ seems to require differentiating $\hat y_j$ with respect to $\theta$,
either through the $K$ solver steps or implicitly, which involves the inverse Hessian of $J_j$, as in
OFM~\cite{kornilov2024optimal}. Danskin's theorem shows that this is unnecessary:

\begin{proposition}[Gradient via Danskin's theorem~\cite{danskin1967}]\label{prop:danskin}
If the extremal point $y^\ast$ of $x_1$ at time $1$ is unique and $\nabla_\theta f_\theta(y)$ exists and is continuous in $(\theta,y)$ near
$(\theta,y^\ast)$ (and, in $\R^n$, $f_\theta$ is Lipschitz with a constant locally bounded in $\theta$), then
$\nabla_\theta u_{f_\theta}(1,x_1)=\nabla_\theta f_\theta(y)\big|_{y=y^\ast}$. For a ReLU network, the regularity
assumption holds as soon as no neuron is exactly at its threshold at $y^\ast$.
\end{proposition}
\emph{In words,} once the extremal point is found, it is held fixed, and it suffices to backpropagate through
$f_\theta$ at that point. 
With an approximate point $\hat y_j$, we apply the same rule; this neglects a term proportional to the gradient of
$J_j$ at $\hat y_j$, which vanishes at an exact minimizer. Each iteration therefore costs, on top of flow matching, the choice of candidates
and $K$ solver steps (see the supplementary material); generation involves no inner problem.\\
\noindent
\textbf{Architecture of the potential.}
If $p_0$ has connected support and $q$ does not, no continuous map pushes $p_0$ onto $q$: $T_{f^\ast}$ has no
continuous version, and a $C^1$ potential cannot generate exactly in one step. We therefore use a ReLU network, whose
discontinuous gradient separates modes along the boundaries between linear regions. For a ReLU MLP in $\R^n$, as in
our experiments in $\R^2$, this choice comes at a price: $\nabla f_\theta$ is piecewise constant, so $T_{f_\theta}$
translates pieces of $p_0$ without deforming them and concentrates mass only by stacking them. More generally, the map
is not injective, which limits exact likelihoods (Sec.~\ref{sec:expsphere}). The potential used on images (a U-Net
encoder with normalization and attention, Sec.~\ref{sec:expcifar}) is not piecewise linear, and its one-step map
contracts mass. Algorithm~\ref{alg} summarizes training: at each iteration, the extremal point of every target is searched from
batch candidates and held fixed, and a gradient step is taken on the estimated loss $\hat{\mathcal L}_1$; generation
then applies $T_{f_\theta}$ once.
\begin{algorithm}[t]
\caption{MorphoFM: training}\label{alg}
\begin{algorithmic}[1]
\Require distributions $p_0,q$; batch size $|B|$; network $f_\theta$ (scalar output); number of solver steps $K$
\Ensure trained potential $f_\theta$
\Repeat
  \State draw a batch $B$: $x_0^i\sim p_0$, $x_1^j\sim q$, $1\le i,j\le|B|$
  \For{$j=1,\dots,|B|$} \Comment{inner problem, no gradient}
    \State $J_j(y)\gets f_\theta(y)+c(y,x_1^j)$
    \State $C_j\gets\{x_1^j,x_0^1,\dots,x_0^{|B|}\}\cup\{\text{memorized point}\}$\label{alg:cand}
    \State $y_j^{(0)}\gets\argmin_{y\in C_j}J_j(y)$
    \For{$s=0,\dots,K-1$}
      \State $y_j^{(s+1)}\gets$ local solver step on $J_j$ from $y_j^{(s)}$
    \EndFor
    \State $\hat y_j\gets$ best point of $y_j^{(0)},\dots,y_j^{(K)}$ \Comment{\textbf{Prop.~\ref{prop:borne}}}\label{alg:best}
  \EndFor
  \State $\ell_f\gets\frac1{|B|}\sum_i f_\theta(x_0^i)-\frac1{|B|}\sum_j J_j(\hat y_j)$ \Comment{\textbf{Prop.~\ref{prop:borne}}}\label{alg:loss}
  \Statex \hspace{\algorithmicindent}\textit{$\hat y_j$ held fixed (\textbf{Prop.~\ref{prop:danskin}}).}
  \State gradient step on $\ell_f$ w.r.t.\ $\theta$
\Until{convergence}
\end{algorithmic}
\end{algorithm}

%-------------------------------------------------------------------
\subsection{Generation: A Single Euler Step Is Exact}\label{sec:gen}
%-------------------------------------------------------------------

\begin{theorem}[One-step generation is exact]\label{thm:gen}
Let $k>1$ and let $p_0$ be absolutely continuous; assume either that $\M=\R^n$ and $p_0,q$ have finite moments of
order $k^\ast$, or that $\M$ is compact. Let $f^\ast$ be a $c$-convex minimizer of $\mathcal L_1$ (Thm.~\ref{thm:dual}),
and
\[
T(x)=\exp_x\!\big(v_0(x)\big),\qquad v_0=\nabla H(\nabla_\g f^\ast),
\]
\ie, $T(x)=x+2\nabla f^\ast(x)$ in $\R^n$ for $k=2$. Then:
\begin{enumerate}[label=(\roman*),leftmargin=*]
\item $(\mathrm{id},T)_\#p_0$ is the unique optimal transport plan from $p_0$ to $q$; in particular, $T_\#p_0=q$;
\item for $p_0$-almost every $x_0$, the trajectory $t\mapsto\psi_t(x_0)=\exp_{x_0}(t\,v_0(x_0))$ is the minimizing
geodesic from $x_0$ to $T(x_0)$, traversed at constant speed; trajectories do not cross; and, for every $t\in(0,1)$,
$x_0$ is the unique extremal point of $\psi_t(x_0)$, hence $v_t(\psi_t(x_0))=\frac{\mathrm d}{\mathrm dt}\psi_t(x_0)$;
\item for every $N\ge1$, the geodesic Euler scheme $x^{(m+1)}=\exp_{x^{(m)}}\!\big(\tfrac1N\,v_{m/N}(x^{(m)})\big)$,
$x^{(0)}=x_0$, with $v_0$ at the first step and the morphological velocity afterwards, yields
$x^{(m)}=\psi_{m/N}(x_0)$ for $p_0$-almost every $x_0$; in particular, a single step gives $T(x_0)\sim q$.
\end{enumerate}
(Item (i) is classical: Gangbo--McCann~\cite{gangbo1996} in $\R^n$, McCann~\cite{mccann2001} for $k=2$ and
Fathi--Figalli~\cite{fathi2010} for any $k$ on a manifold. Full statement and self-contained proof: supplementary
material.)
\end{theorem}
\vspace{-3mm}
\emph{In words,} at the optimal potential, particles follow geodesics at constant speed without crossing, and a single
step integrates the flow exactly (Fig.~\ref{fig:overview}(c)). Generation costs one gradient evaluation of the potential; more steps numerically
give the same samples at 100--1,000 times the computational cost (see the supplementary material).

\noindent
\textbf{What the morphological flow is for.}
The flow is not integrated at generation time, but it plays three roles. It explains why a single step is exact
(Thm.~\ref{thm:gen}). It relates the loss to CFM: by item (ii), for $(x_0,x_1)$ drawn from the optimal plan,
$v_t(x_t)=\dot x_t$ at every $t\in(0,1)$, so that at the optimal potential the morphological field cancels the OT-CFM loss
with the exact plan, although training only uses independent pairs. Finally, averaging its gap $G_t$ over $t$ is what
identified the $t=1$ term as the only one that does not depend on the coupling (see the supplementary material).

%===================================================================

%=================================================================
\section{Experiments}\label{sec:experiences}
%=================================================================
We evaluate our method on three families of problems: synthetic distributions in $\mathbb{R}^2$
(Section~\ref{sec:exp2d}), real-world geographic data on the sphere $\mathbb{S}^2$ (Section~\ref{sec:expsphere}),
and images, CIFAR-10 (Section~\ref{sec:expcifar}). Explicit formulas for $\mathbb{R}^n$ and $\mathbb{S}^2$ and all
reproduction details (datasets, metrics, implementation, and computational cost) are given in the supplementary material.

\noindent
\textbf{Protocol principles.}
To avoid comparing against weakened baselines, we use the authors' \emph{official code} whenever it is available,
and we state explicitly every reimplementation. Experiments in $\mathbb{R}^2$ and on $\mathbb{S}^2$ are repeated over
5 seeds (mean $\pm$ standard deviation). Metrics are computed on test data never seen during training. For each
method, we report the number of network function evaluations at sampling time (NFE): our method, W2GN and OFM
generate with $\mathrm{NFE}=1$, while ODE-based methods are evaluated with several numbers of steps (Euler scheme).

\noindent
\textbf{Same budget.}
Unless otherwise stated, methods are compared \emph{at equal budget} (\emph{same budget} in the tables): same network,
same batch size, and same number of training iterations. The equal budget refers to the number of iterations, not to
the computation time: one iteration of our method is more expensive, since it solves the inner problem; training
times and a comparison at equal training time are given in the supplementary material. Exceptions are stated explicitly: W2GN and OFM follow their authors' protocols,
2-Rectified Flow uses two training runs, and the official RFM as well as the official CIFAR-10 checkpoints are used as
released.

OFM and W2GN are the baselines conceptually closest to ours: they also generate in one step through the gradient of a
potential derived from Kantorovich duality, but this potential is convex and, with their smooth activations, its
gradient is continuous.

% ---------------------------------------------------------------------
\subsection{Synthetic Distributions in \texorpdfstring{$\mathbb{R}^2$}{R2}}
\label{sec:exp2d}

\noindent
\textbf{Data and setup.}
We use four standard distributions: 8 Gaussians (a mixture of 8 isotropic Gaussians with variance $0.1$ placed on a
circle of radius 5), Two Moons, Checkerboard, and Spirals; the source is $\mathcal{N}(0,I_2)$. Unless otherwise
stated, all networks are $3\times256$ MLPs trained for 5{,}000 iterations with a batch size of 512, Adam with a
learning rate of $10^{-3}$ and cosine decay (MeanFlow keeps its authors' configuration). The MMD (kernel bandwidths
from $0.1$ to $5$) and the W2 distance are computed between 2{,}000 generated samples and 2{,}000 test samples;
ODE-based methods are evaluated with 1, 2, 5, 10, and 100 steps.

\begin{table}[!tbp]
\centering
\small
\setlength{\tabcolsep}{4pt}
\caption{MMD ($\downarrow$) between 2{,}000 generated samples and 2{,}000 held-out data samples (mean $\pm$ std over
5 seeds). NFE: number of network evaluations at sampling time. The gray row is the MMD between two independent data
samples and indicates the sampling-noise level. Best value per column highlighted; best one-step value underlined. 
Results for 2 and 5 steps are given in the supplementary material.}
\label{tab:2d_mmd}
\resizebox{\linewidth}{!}{%
\begin{tabular}{l|c|cccc}
\toprule
\multirow{2}{*}{Method} & \multirow{2}{*}{NFE} & \multicolumn{4}{c}{MMD $\downarrow$} \\
\cmidrule(lr){3-6}
 & & 8 Gaussians & Two Moons & Checkerboard & Spirals \\
\midrule
\rowcolor{refgray}
Data (independent sample) & -- & 0.0672 & 0.0599 & 0.0691 & 0.0615 \\
\midrule
Flow Matching~\cite{lipman2023flow} & 1 & $1.7671 \pm 0.0093$ & $1.6853 \pm 0.0257$ & $1.6237 \pm 0.0160$ & $1.6125 \pm 0.0326$ \\
 & 10 & $0.1106 \pm 0.0085$ & $0.1507 \pm 0.0068$ & $0.1240 \pm 0.0061$ & $0.1515 \pm 0.0073$ \\
 & 100 & $0.0737 \pm 0.0106$ & $0.0646 \pm 0.0053$ & $0.0757 \pm 0.0049$ & $0.0785 \pm 0.0057$ \\
\midrule
OT-CFM~\cite{tong2024improving} & 1 & $0.1610 \pm 0.0073$ & $0.0972 \pm 0.0045$ & $0.1032 \pm 0.0035$ & $0.0921 \pm 0.0043$ \\
 & 10 & $0.0804 \pm 0.0083$ & $0.0658 \pm 0.0050$ & $0.0814 \pm 0.0037$ & $0.0797 \pm 0.0043$ \\
 & 100 & $0.0729 \pm 0.0087$ & $0.0641 \pm 0.0050$ & $0.0801 \pm 0.0039$ & $0.0789 \pm 0.0043$ \\
\midrule
2-Rectified Flow~\cite{liu2023flow} & 1 & $0.1168 \pm 0.0048$ & $0.0752 \pm 0.0044$ & $0.0811 \pm 0.0050$ & $0.0832 \pm 0.0053$ \\
 & 10 & $0.0845 \pm 0.0088$ & $0.0675 \pm 0.0050$ & $0.0765 \pm 0.0048$ & $0.0795 \pm 0.0056$ \\
 & 100 & $0.0769 \pm 0.0097$ & $0.0662 \pm 0.0051$ & $0.0762 \pm 0.0049$ & $0.0792 \pm 0.0057$ \\
\midrule
MeanFlow~\cite{geng2025mean} & 1 & $0.2494 \pm 0.0091$ & $0.1332 \pm 0.0062$ & $0.1230 \pm 0.0052$ & $0.1586 \pm 0.0126$ \\
 & 10 & $0.2634 \pm 0.0060$ & $0.1667 \pm 0.0145$ & $0.1934 \pm 0.0143$ & $0.1242 \pm 0.0121$ \\
 & 100 & $0.2833 \pm 0.0052$ & $0.1853 \pm 0.0146$ & $0.2149 \pm 0.0150$ & $0.1505 \pm 0.0124$ \\
\midrule
W2GN~\cite{korotin2021wasserstein} & 1 & $0.0999 \pm 0.0132$ & $0.0992 \pm 0.0109$ & $0.0858 \pm 0.0067$ & $0.0923 \pm 0.0096$ \\
\midrule
OFM~\cite{kornilov2024optimal} & 1 & $0.0821 \pm 0.0142$ & $0.1088 \pm 0.0136$ & $0.0875 \pm 0.0054$ & $0.0873 \pm 0.0060$ \\
\midrule
\textbf{MorphoFM (ours)} & 1 & \best{\underline{0.0695 \pm 0.0112}} & \best{\underline{0.0579 \pm 0.0064}} & \best{\underline{0.0701 \pm 0.0052}} & \best{\underline{0.0618 \pm 0.0054}} \\
\bottomrule
\end{tabular}}
\end{table}

\noindent
\textbf{Results.}
Table~\ref{tab:2d_mmd} shows that our method, with a single step, achieves the best MMD on all four datasets, not
only among one-step methods but also against all ODE-based methods integrated with 100 steps. It outperforms the best
competitor, across all numbers of steps, on 19 out of 20 seeds (the only exception being one seed of 8 Gaussians
against OT-CFM with 100 steps). Its values are at the sampling floor: on Two Moons and Spirals, they are
indistinguishable from those obtained between two independent samples of the true data. With a single step,
ODE-based methods either degrade severely (Flow Matching) or remain well above the floor (OT-CFM, 2-Rectified Flow),
and so do the one-step methods based on convex potentials (W2GN, OFM). Against OFM, the closest baseline, the gap
exceeds three standard deviations on Two Moons, Checkerboard, and Spirals. MeanFlow, although designed for one-step
generation, lags behind all other methods on this problem; with the official code, its quality even decreases as the
number of steps increases, on three of the four datasets. In Wasserstein-2 distance (see the supplementary material),
our one-step method reaches or falls below the noise floor on all four datasets; on Two Moons, Checkerboard, and
Spirals, it is the best one-step method, $0.04$ to $0.05$ below the best one-step competitor (2-Rectified Flow).

\noindent
\noindent
\textbf{Qualitative analysis.}
Figure~\ref{fig:teaser} illustrates the theoretical argument: a continuous transport map cannot push a Gaussian onto
a distribution with disjoint supports without leaving mass between the modes. This is what we observe for W2GN and
OFM, whose convex potentials with smooth activations have a continuous gradient, and for OT-CFM with one step, which
produces a polygon connecting the modes. Our ReLU potential allows a discontinuous gradient: the modes are separated
in a single step, with only a few isolated points between them. The samples of all methods, including 2-Rectified
Flow, at 1, 10, and 100 steps, and the MMD as a function of the number of steps, are shown in the supplementary
material: ODE-based methods only reach our level of quality after 10 to 100 integration steps, when they reach it at
all.

\subsection{Geographic Data on the Sphere \texorpdfstring{$\mathbb{S}^2$}{S2}}
\label{sec:expsphere}

\noindent
\textbf{Dataset and setup.}
We use the four standard Earth datasets of~\cite{mathieu2020riemannian}: Volcano (827 points), Earthquake (6{,}120),
Flood (4{,}875), and Fire (12{,}809). We follow the protocol of the official RFM code~\cite{chen2024flow}, with
identical data processing and the same 80/10/10 splits over five seeds. The source distribution is uniform on
$\mathbb{S}^2$. At equal budget, RFM uses the same $3\times256$ MLP, batches, and number of training iterations as
ours. We also evaluate the official RFM model, trained for approximately 808k iterations on Volcano. MMD is computed
on 2{,}000 generated samples and kernel NLL on 20{,}000 samples, with kernel parameters selected on validation data.
Further protocol details and published exact-NLL baselines are provided in the supplementary material.

\begin{table*}[!t]
\centering
\small
\setlength{\tabcolsep}{3pt}
\caption{Earth datasets on $\mathbb{S}^2$ (mean $\pm$ std over 5 seeds, official 80/10/10 splits of~\cite{chen2024flow}).
MMD ($\downarrow$) and kernel NLL ($\downarrow$, von Mises--Fisher kernel density estimate fitted on 20{,}000 generated
samples, bandwidth selected on the validation split) are computed on the test split. The gray ``Reference'' row
replaces generated samples by training samples. RFM (same budget) uses the same MLP and number of iterations as our
method. Best generative value per column highlighted.}
\label{tab:sphere}
\resizebox{\textwidth}{!}{%
\begin{tabular}{l|c|cc|cc|cc|cc}
\toprule
\multirow{2}{*}{Method} & \multirow{2}{*}{NFE} & \multicolumn{2}{c|}{Volcano} & \multicolumn{2}{c|}{Earthquake} & \multicolumn{2}{c|}{Flood} & \multicolumn{2}{c}{Fire} \\
\cmidrule(lr){3-4}\cmidrule(lr){5-6}\cmidrule(lr){7-8}\cmidrule(lr){9-10}
 & & MMD $\downarrow$ & NLL $\downarrow$ & MMD $\downarrow$ & NLL $\downarrow$ & MMD $\downarrow$ & NLL $\downarrow$ & MMD $\downarrow$ & NLL $\downarrow$ \\
\midrule
\rowcolor{refgray}
Reference (training data) & -- & $0.246 \pm 0.023$ & $0.04 \pm 3.55$ & $0.102 \pm 0.007$ & $0.39 \pm 0.23$ & $0.117 \pm 0.004$ & $0.77 \pm 0.14$ & $0.073 \pm 0.009$ & $-0.90 \pm 0.23$ \\
\midrule
RFM~\cite{chen2024flow}, same budget & 1 & $0.594 \pm 0.042$ & $2.49 \pm 0.05$ & $0.493 \pm 0.008$ & $2.40 \pm 0.01$ & $0.421 \pm 0.014$ & $2.43 \pm 0.03$ & $0.645 \pm 0.011$ & $2.54 \pm 0.02$ \\
 & 10 & $0.328 \pm 0.026$ & $0.25 \pm 0.21$ & $0.155 \pm 0.004$ & $0.94 \pm 0.04$ & $0.151 \pm 0.004$ & $1.43 \pm 0.02$ & $0.181 \pm 0.009$ & $0.15 \pm 0.05$ \\
 & 100 & $0.318 \pm 0.022$ & $0.33 \pm 0.16$ & $0.147 \pm 0.004$ & $0.91 \pm 0.05$ & $0.148 \pm 0.003$ & $1.40 \pm 0.02$ & $0.145 \pm 0.005$ & $0.07 \pm 0.06$ \\
\midrule
\textbf{MorphoFM (ours)} & 1 & \best{0.252 \pm 0.019} & \best{-1.39 \pm 0.56} & \best{0.107 \pm 0.008} & \best{0.27 \pm 0.10} & \best{0.116 \pm 0.006} & \best{0.83 \pm 0.03} & \best{0.073 \pm 0.006} & \best{-0.62 \pm 0.09} \\
\bottomrule
\end{tabular}}
\end{table*}

\noindent
\textbf{Results at equal budget.}
Table~\ref{tab:sphere} shows that our one-step method outperforms RFM on all four datasets and for both metrics at
equal training budget, even against RFM with 100 integration steps. Our MMD reaches the training-data reference on all
datasets (\eg, $0.073$ vs.\ $0.073$ on Fire and $0.116$ vs.\ $0.117$ on Flood). The unusually high variance of the
Volcano reference kernel NLL ($0.04\pm3.55$) is caused by a single seed containing isolated test points that are
assigned near-zero density by the training-data kernel estimator.

\noindent
\textbf{Training budget.}
To verify that this conclusion does not stem from an undertrained competitor, we trained RFM with its authors' full
protocol on Volcano (see the supplementary material), with about 2{,}600 times more training points seen and a much
larger network. With 1{,}000 integration steps, this model achieves a better likelihood than ours, but a worse MMD
($0.264$ vs.\ $0.247$, for a reference of $0.244$).

\noindent\textbf{Likelihood estimation.}
As detailed in the supplementary material, our
learned transport map is not necessarily injective
and does not cover the entire sphere.
We evaluate a partial inverse-based log-density
estimate on the 35--47\% of test points for which
numerical inversion succeeds.
Since the estimate accounts for only one preimage
and excludes non-inverted points, it does not
represent the exact NLL of the full distribution.
This highlights a limitation of MorphoFM for
density estimation, despite its efficient one-step
generation.

\begin{figure}[th]
\centering
 \includegraphics[width=0.95\linewidth]{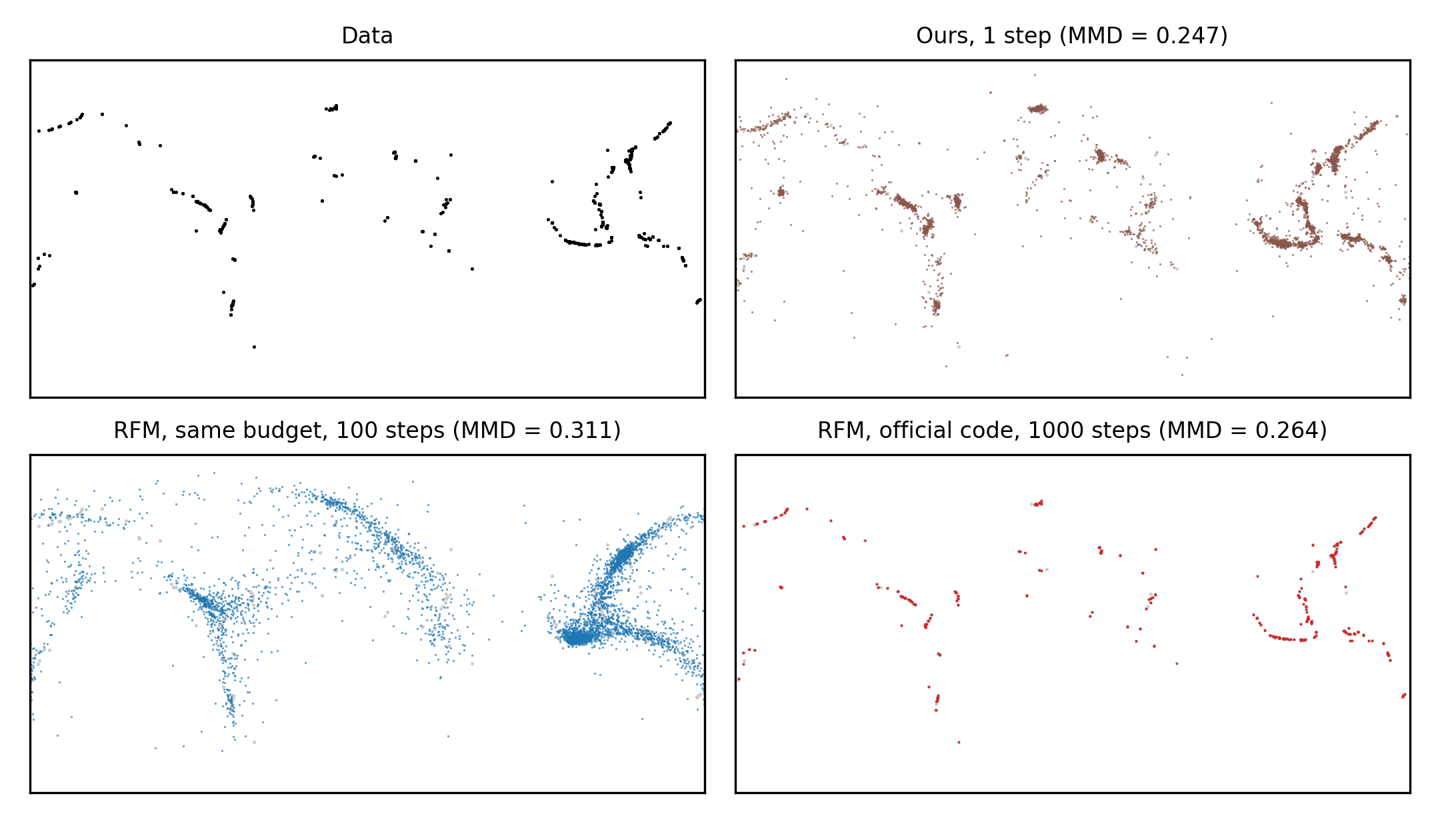}
\caption{Generated samples on the Volcano dataset (seed 0, longitude--latitude projection; gray: data). Our method
with one step; RFM at the same budget with 100 steps; official RFM with 1{,}000 steps. The reported MMD is that of
seed 0. Our method reproduces the volcanic arcs in a single step, whereas RFM at the same budget spreads out the mass
and creates arcs absent from the data. The four Earth datasets are shown in the supplementary material.}
\label{fig:volcano}
\end{figure}

\noindent
\textbf{Qualitative analysis and cost.}
Figure~\ref{fig:volcano} shows that our one-step method recovers the volcanic arcs, whereas RFM at equal budget
produces diffuse samples and spurious structures. Similar results are observed on the other Earth datasets (see the
supplementary material). Generating 20{,}000 samples takes approximately 1~ms, about 50 times less than RFM with
100 integration steps.

% ---------------------------------------------------------------------
\subsection{Images: CIFAR-10}
\label{sec:expcifar}

\noindent\textbf{Goal and setup.}
We evaluate the scalability of MorphoFM to
high-dimensional image generation on CIFAR-10
(50,000 training images of size $32\times32$,
represented in $\mathbb{R}^{3072}$).
Our goal is to assess whether the same training
and one-step generation framework extends beyond
low-dimensional distributions, and to investigate
its limitations on natural images.
We use a Gaussian source $\mathcal{N}(0,I)$ and
the quadratic transport cost ($k=2$).
The potential $f_\theta$ is parameterized by the
encoder of the CIFAR-10 U-Net from \texttt{torchcfm},
adapted to produce a scalar output with ReLU
activations (35.2M parameters, compared with
35.7M for the baseline U-Net).

\noindent\textbf{Training and evaluation.}
We follow the CIFAR-10 training protocol of
\texttt{torchcfm}, using Adam with a learning
rate of $2\cdot10^{-4}$, linear warm-up,
exponential moving average (EMA) of the weights,
and a batch size of 128.
The inner optimization follows the same procedure
as in $\mathbb{R}^2$, with $K=5$ L-BFGS steps.
After 220k iterations, we introduce memorized
extremal points as additional initialization
candidates (Sec.~\ref{sec:num}) and continue
training for 90k iterations.
We select the checkpoint with the lowest
monitoring FID, obtained at 310k iterations.
For comparison, we evaluate the official FM
and OT-CFM checkpoints (400k iterations)
using Euler integration with 1, 2, 10, and
100 steps, as well as the adaptive
\texttt{dopri5} solver.
MorphoFM generates directly using EMA weights
and the one-step map
$x_1=x_0+2\nabla f_\theta(x_0)$.
Further implementation details are provided
in the supplementary material.

\begin{table}[!tbp]
\centering
\small
\setlength{\tabcolsep}{3pt}
\caption{\textbf{Generation quality on CIFAR-10.}
FID-50K ($\downarrow$) is computed using
\texttt{clean-fid} with the same evaluation protocol
for all methods.
FM and OT-CFM use official checkpoints trained
for 400k iterations, while MorphoFM is trained
for 310k iterations, including 90k with memorized
extremal points.
Unlike ODE-based baselines, MorphoFM generates
in a single step without numerical integration.
The best one-step result is highlighted.}

%\caption{CIFAR-10, FID-50K ($\downarrow$; clean-fid, \texttt{legacy\_tensorflow} mode, training-set statistics), all
%methods evaluated with the same code. Flow Matching and OT-CFM use the official \texttt{torchcfm} checkpoints (U-Net,
%35.7M parameters, batch size 128), sampled with the Euler scheme and with the adaptive \texttt{dopri5} solver
%(tolerance $10^{-5}$). MorphoFM uses a 35.2M-parameter potential, batch size 128, trained for 220k iterations, then
%90k with memorized extremal points; it generates in a single step (Thm.~\ref{thm:gen}). Published results of other
%one-step methods are given in the supplementary material. Best one-step value highlighted.}
\label{tab:cifar}
\resizebox{\linewidth}{!}{%
\begin{tabular}{l|c|ccccc}
\toprule
\multirow{2}{*}{Method} & \multirow{2}{*}{Iter. (epochs)} & \multicolumn{5}{c}{FID-50K $\downarrow$} \\
\cmidrule(lr){3-7}
 & & 1 NFE & 2 NFE & 10 NFE & 100 NFE & dopri5 \\
\midrule
Flow Matching~\cite{lipman2023flow} & 400k (1{,}024) & 350.57 & 165.61 & 14.13 & 4.57 & 3.72 \\
OT-CFM~\cite{tong2024improving}     & 400k (1{,}024) & 231.48 & 93.16  & 11.90 & 4.52 & 3.69 \\
\midrule
\textbf{MorphoFM (ours)}            & 310k (794)     & \best{31.32} & -- & -- & -- & -- \\
\bottomrule
\end{tabular}}
\end{table}

\begin{figure}[th]
    \centering
\includegraphics[width=0.95\linewidth]{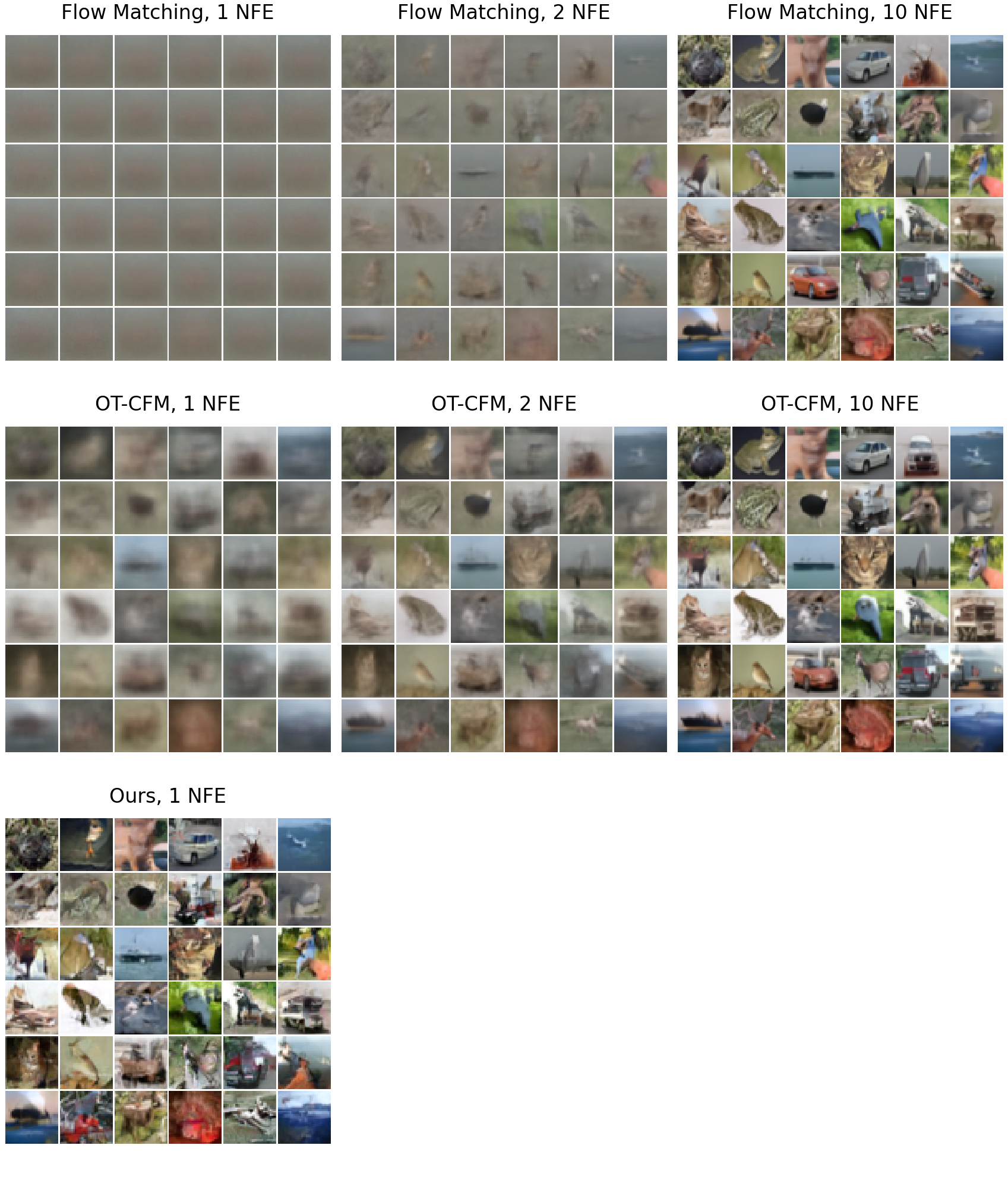}
\caption{CIFAR-10 samples from the same noise for all methods: Flow Matching and OT-CFM with 1, 2, and 10 steps, our
method with a single step. With one step, Flow Matching collapses to the mean image and OT-CFM produces blurry shapes,
whereas our map produces recognizable objects (one-step FID-50K: 350.6, 231.5, and 31.3; Table~\ref{tab:cifar}).}
\label{fig:cifar}
\end{figure}

\noindent
\textbf{Results.}
Table~\ref{tab:cifar} shows that, with a single step, our method reaches an FID of $31.3$, versus $231.5$ for OT-CFM
and $350.6$ for Flow Matching, with fewer training iterations (310k versus 400k); it also outperforms OT-CFM and Flow
Matching integrated with 2 steps ($93.2$ and $165.6$). Figure~\ref{fig:cifar} illustrates this from the same noise:
with one step, Flow Matching collapses to the mean image and OT-CFM produces blurry shapes, whereas our map produces
recognizable objects. The same algorithm thus scales to images, in dimension 3{,}072, and still generates in a single
step.

\noindent
\textbf{What limits the quality.}
Our experiments suggest that the accuracy of the inner optimization, rather than the number of training iterations,
is a major factor limiting generation quality. At 220k iterations, the FID is $63.9$, and continuing training without
memorization does not improve the monitoring FID. Memorizing the extremal points, which only gives the solver a better
starting point and changes nothing else, brings it down to $32.5$ in 80k iterations, then to $31.3$ at the selected
checkpoint; early memorization trains faster (see the supplementary material). With 10 steps or more, Flow Matching
and OT-CFM remain clearly better, and MeanFlow~\cite{geng2025mean}, whose architecture and training recipe are
designed for images (U-Net with dropout, EDM-style augmentation, batches of 1{,}024 and about 700k iterations, \ie,
about 18 times more training images than ours), reaches an FID of $2.80$ in one step (see the supplementary material).
This recipe does not transfer to low dimension: in $\mathbb{R}^2$, with the same network and budget as the other
methods, its one-step MMD is 1.8 to 3.6 times higher than ours (Table~\ref{tab:2d_mmd}), and its smooth network yields
a continuous one-step map, which cannot separate disconnected modes (Sec.~\ref{sec:mfm}). The remaining gap on images
thus lies in the inner solver and in the architecture of the potential, which we did not adapt to images (see the
limitations in the conclusion).

% ---------------------------------------------------------------------
\subsection{Ablations}
\label{sec:ablations}
Ablation studies in the supplementary material validate the main design choices. The terminal loss at $t=1$ clearly
outperforms uniform time weighting and direct CFM training, and ReLU potentials outperform smooth activations on all
20 (dataset, seed) pairs, which confirms the role of a discontinuous gradient. Further ablations assess the solver
accuracy, the starting candidates, training at equal wall-clock time, multi-step generation, and the exponent $k$, as
well as the role of the intrinsic geometry on the sphere.

\section{Conclusion and Limitations}
\label{sec:conclusion}

\noindent
\textbf{Conclusion.}
We introduced MorphoFM, which defines a generative flow through the Hopf--Lax erosion of a learned scalar potential,
in Euclidean spaces and on Riemannian manifolds. Its loss, derived from an optimality gap of the erosion, is, up to
sign, the semi-dual Kantorovich objective; at the optimal potential, trajectories are geodesics and a single step
generates exactly, without ODE integration. Non-convex ReLU potentials let the one-step map separate disconnected
modes, which no continuous one-step generator can do. With a single step, MorphoFM reaches the reference level on
synthetic and spherical data, and the same algorithm scales to CIFAR-10.

\noindent
\textbf{Limitations.}
Exactness is guaranteed only at the optimal potential: in practice, the non-convex inner problem is solved
approximately, and each training iteration costs more than a flow matching iteration. On images, this inner problem is
the bottleneck: memorizing extremal points brings the one-step FID from $63.9$ down to about $32$, which remains far
from methods designed for images. The learned map is not injective, so its density is only defined on its image, and
our manifold experiments are limited to the sphere. More accurate inner solvers and potential architectures adapted to
image generation are natural directions for future work.

%====================================================================
\newpage
\section*{LLM Usage Statement}

During the preparation of this work, the authors used ChatGPT (OpenAI) as an assistant for three purposes: (i) editing the manuscript text to improve clarity, grammar, and readability; (ii) refining and debugging parts of the source code; and (iii) helping produce the figures, with the aim of presenting high-quality visualizations. All visual results shown in the figures are produced by the authors' method and the compared baselines; the LLM was used only to improve their presentation, not to generate or alter the underlying results. The research ideas, method design, experimental setup, analysis of results, and conclusions are entirely the work of the authors. The authors reviewed and verified all AI-assisted text, code, and visual material, and take full responsibility for the content of this article.

\maketitlesupplementary

% Sections become A, B, C...; tables/figures/equations become S1, S2...
\appendix
\setcounter{table}{0}
\setcounter{figure}{0}
\setcounter{equation}{0}
\renewcommand{\thetable}{S\arabic{table}}
\renewcommand{\thefigure}{S\arabic{figure}}
\renewcommand{\theequation}{S\arabic{equation}}

\paragraph{Overview.}

This supplementary material provides theoretical proofs,
implementation details, and additional experimental results
supporting the main paper. The theorems and propositions
underpinning the development of MorphoFM are highlighted
in boxes, each followed by a detailed proof.

Section~\ref{supp:preuves} establishes the results presented
in the Morphological Flow Matching (MorphoFM) section,
following the same order and headings.
Section~\ref{supp:gen} provides a complete statement and
self-contained proof of the generation theorem for any $k>1$.
Section~\ref{supp:formules} derives explicit formulas in
$\R^n$ and on the sphere $\mathbb S^2$.

Sections~\ref{supp:details}--\ref{supp:metrics} describe the
experimental protocols for $\R^2$, $\mathbb S^2$, and CIFAR-10,
covering implementation details, datasets, baselines, and
evaluation metrics.
Section~\ref{supp:resultats} presents additional results,
while Section~\ref{supp:ablations} reports ablation studies
in $\R^2$.

Finally, Section~\ref{supp:compl} provides complementary
theoretical analyses: the incompatibility of intermediate-time
objectives with independent coupling (Sec.~3.2 of the main
paper), the convexity of the inner problem at the optimum,
and the topological obstruction motivating ReLU potentials
(Sec.~3.3).

\begin{table}[t]
\centering
\footnotesize
\setlength{\tabcolsep}{3pt}
\renewcommand{\arraystretch}{1.12}
\caption{\textbf{Notation.} Symbols used in the main
paper and supplementary material.}
\label{tab:notation}

\begin{tabularx}{\columnwidth}{
@{}>{\raggedright\arraybackslash}p{0.24\columnwidth}
>{\raggedright\arraybackslash}X@{}}
\toprule
Symbol & Meaning \\
\midrule

$(\M,\g)$ &
Complete connected Riemannian manifold,
with distance $d_\g$, gradient $\nabla_\g$,
and exponential map $\exp$ \\

$\Cut(x)$ &
Cut locus of $x$ \\

$k,k^\ast,c_k$ &
Exponent $k>1$;
$k^\ast=k/(k-1)$;
$c_k=(k-1)/k^{k^\ast}$ \\

$H$ &
Hamiltonian on $T^\star\M$ (identified
with $T\M$ via $\g$),
$H(p)=\|p\|_\g^k$ \\

$L$ &
Lagrangian on $T\M$, the Legendre transform
of $H$;
$L(w)=c_k\|w\|_\g^{k^\ast}$ \\

$b_t$ &
Structuring function, $t>0$:
$b_t(x,y)=tL(d_\g(x,y)/t)$;
equivalently,
$c_kd_\g(x,y)^{k^\ast}t^{-1/(k-1)}$ \\

$c$ &
Transport cost,
$c(x,y)=b_1(x,y)=c_kd_\g(x,y)^{k^\ast}$ \\

$u_f(t,x)$ &
Hopf--Lax erosion of $f$ at scale $t$:
$\inf_y\{f(y)+b_t(x,y)\}$ \\

$Y_t,\ y_t^\ast$ &
Set of extremal points $Y_t(x)$;
$y_t^\ast(x)$ when the minimizer is unique \\

$v_t(x)$ &
Morphological velocity:
$-\frac{1}{t}\exp_x^{-1}(y_t^\ast(x))$ \\

$T_f$ &
One-step transport map:
$T_f(x)=\exp_x(\nabla H(\nabla_\g f(x)))$ \\

$p_0,\ q$ &
Source and target probability distributions \\

$f_\theta,\ f^\ast$ &
Learned and optimal potentials, respectively \\

$\mathcal L_1$ &
Semi-dual training loss:
$\mathcal L_1(f)=\E_{p_0}f-\E_q u_f(1,\cdot)$ \\

$G_t$ &
Erosion optimality gap:
$G_t(f;x_0,x_1)=f(x_0)+t\,c(x_0,x_1)
-u_f(t,x_t)$ \\

$W_c(p_0,q)$ &
Optimal transport cost between $p_0$
and $q$ for the cost $c$ \\

$B,\ |B|$ &
Batch of $|B|$ source samples and
$|B|$ target samples \\

$J_j,\ \hat y_j$ &
Inner objective
$J_j(y)=f_\theta(y)+c(y,x_1^j)$;
best candidate found $\hat y_j$ \\

$\hat{\mathcal L}_1$ &
Batch-based estimate of the training loss \\

\bottomrule
\end{tabularx}
\end{table}

\paragraph{Notation.}
Table~\ref{tab:notation} summarizes the notation used
throughout the main paper and this supplementary material.

By a slight abuse of notation, $L$ and $H$ also denote
the scalar functions $r\mapsto c_k r^{k^\ast}$ and
$r\mapsto r^k$, respectively, for $r\in\R_+$.
Using
$tL(d/t)=c_k d^{k^\ast}t^{1-k^\ast}$
and $1-k^\ast=-1/(k-1)$ yields the alternative
expression for the structuring function $b_t$.

In the Euclidean setting $\M=\R^n$,
$d_\g(x,y)=|x-y|$,
$\exp_x(w)=x+w$,
$\exp_x^{-1}(y)=y-x$, and
$\Cut(x)=\emptyset$.

%=====================================================================
\section{Proofs of the results of the Method section}\label{supp:preuves}
%=====================================================================
%---------------------------------------------------------------------
\subsection{Preliminary lemmas}
%---------------------------------------------------------------------
%---------------------------------------------------------------------
\begin{lemma}[Conjugacy]\label{lem:conj}
On each tangent space, $H$ and $L$ are of class $C^1$ and strictly convex, $L=H^\ast$, and
\[
\begin{aligned}
\nabla H(p)&=k\|p\|^{k-2}p,\\
\nabla L(w)&=c_kk^\ast\|w\|^{k^\ast-2}w,
\end{aligned}
\]
these formulas being extended by $\nabla H(0)=0$ and $\nabla L(0)=0$. The maps $\nabla H$ and $\nabla L$ are inverse
bijections of each other, and they are odd.
\end{lemma}
\begin{proof}
For $w\neq0$, $p\mapsto\langle p,w\rangle-\|p\|^k$ is strictly concave, with zero gradient at
$p=(\|w\|/k)^{1/(k-1)}w/\|w\|$. The maximal value is
\[
\begin{aligned}
&\|w\|\left(\frac{\|w\|}{k}\right)^{\!\frac1{k-1}}
-\left(\frac{\|w\|}{k}\right)^{\!\frac{k}{k-1}}\\
&=\|w\|^{k^\ast}\left(k^{-\frac1{k-1}}-k^{-k^\ast}\right)
=c_k\|w\|^{k^\ast},
\end{aligned}
\]
since $k^{-1/(k-1)}=k\cdot k^{-k^\ast}$; for $w=0$, it equals $0=L(0)$. Since $k,k^\ast>1$, $r\mapsto r^k$ and
$r\mapsto r^{k^\ast}$ are of class $C^1$ on $\R_+$, with zero derivative at $0$: $H$ and $L$ are $C^1$, with gradients
given by the formulas (continuous at $0$, since $\|\nabla H(p)\|=k\|p\|^{k-1}\to0$ and $\|\nabla L(w)\|\to0$). With
$c_kk^\ast=k^{-1/(k-1)}$, we have $\|\nabla L(w)\|=(\|w\|/k)^{1/(k-1)}$, hence
$\nabla H(\nabla L(w))=k\|\nabla L(w)\|^{k-1}w/\|w\|=w$; similarly, $\|\nabla H(p)\|=k\|p\|^{k-1}$ and
$\nabla L(\nabla H(p))=c_kk^\ast(k\|p\|^{k-1})^{k^\ast-1}p/\|p\|=p$, since $(k-1)(k^\ast-1)=1$ and
$c_kk^\ast k^{k^\ast-1}=1$. Both compositions are the identity (also at $0$): $\nabla H$ and $\nabla L$ are inverse
bijections. Oddness is immediate.
\end{proof}
\goodbreak
\begin{lemma}[Concatenation of actions]\label{lem:concat}
For $t\in(0,1)$ and $a,m,b\in\M$,
\[
b_1(a,b)\le b_t(a,m)+b_{1-t}(m,b),
\]
with equality if and only if $d_\g(a,b)=d_\g(a,m)+d_\g(m,b)$ and $d_\g(a,m)/t=d_\g(m,b)/(1-t)$, that is, if and only
if $m$ lies on a minimizing geodesic from $a$ to $b$, at the fraction $t$ of its length.
\end{lemma}
\begin{proof}
$L$ is strictly increasing and strictly convex on $\R_+$. By the triangle inequality and then by convexity,
\[
\begin{aligned}
b_1(a,b)&=L\big(d(a,b)\big)\\
&\le L\big(d(a,m)+d(m,b)\big)\\
&=L\!\left(t\frac{d(a,m)}{t}+
(1-t)\frac{d(m,b)}{1-t}\right)\\
&\le tL\!\left(\frac{d(a,m)}{t}\right)\\
&\quad +(1-t)L\!\left(\frac{d(m,b)}{1-t}\right).
\end{aligned}
\]
and the right-hand side equals $b_t(a,m)+b_{1-t}(m,b)$. The first inequality is an equality if and only if
$d(a,b)=d(a,m)+d(m,b)$ (strict monotonicity), and the second if and only if $d(a,m)/t=d(m,b)/(1-t)$ (strict
convexity).
\end{proof}
\begin{lemma}[Localization of the extremal point in $\mathbb{R}^n$]\label{lem:loc}
Let $f$ be $\Lip$-Lipschitz on $\R^n$. For all $t>0$ and $x\in\R^n$, $Y_t(x)$ is a nonempty compact set contained in the closed ball of center $x$
and radius $R_t=(\Lip/c_k)^{k-1}t$, and $u_f(t,\cdot)$ is $\Lip$-Lipschitz.
\end{lemma}
\begin{proof}
Let $\phi_x(y)=f(y)+b_t(x,y)$ and $r=|x-y|$. Since $f$ is $\Lip$-Lipschitz, $\phi_x(y)\ge f(x)-\Lip r+c_kr^{k^\ast}t^{-1/(k-1)}$, which tends to
$+\infty$ as $r\to\infty$ since $k^\ast>1$: $\phi_x$ is continuous and coercive, hence attains its minimum on a
compact set. If $y\in Y_t(x)$, then $\phi_x(y)\le\phi_x(x)=f(x)$, hence $c_kr^{k^\ast}t^{-1/(k-1)}\le\Lip r$, i.e.,
$r^{k^\ast-1}\le(\Lip/c_k)t^{1/(k-1)}$; since $k^\ast-1=1/(k-1)$, $r\le R_t$. Finally,
$u_f(t,x)=\inf_z\{f(x-z)+b_t(z,0)\}$ is an infimum of $\Lip$-Lipschitz functions of $x$.
\end{proof}
The Lipschitz assumption holds for the MLP potentials used in $\R^2$ and on $\mathbb S^2$, since an MLP with ReLU
activations is a composition of Lipschitz maps. It need not hold for the optimal potential $f^\ast$ (in the Gaussian example of
Proposition~\ref{prop:incompat}, $f^\ast$ is quadratic); the results on $f^\ast$ (Theorems~\ref{thm:dual}
and~\ref{thm:gen}, Corollary~\ref{cor:otimplicite}) do not use it.
%---------------------------------------------------------------------
\subsection{Morphological Velocity Field}
%---------------------------------------------------------------------
\begin{tcolorbox}[colback=gray!3,colframe=gray!55]
\begin{proposition}[Velocity is the gradient of the erosion]\label{prop:vit}
Let $t>0$ ($\M$ compact and $f$ continuous, or $\M=\R^n$ and $f$ $\Lip$-Lipschitz). The function $u_f(t,\cdot)$ is Lipschitz, hence differentiable
almost everywhere. At every point $x$ where it is differentiable, $Y_t(x)=\{y^\ast_t(x)\}$ is a singleton, $x$ and
$y^\ast_t(x)$ are joined by a unique minimizing geodesic, and
\[
\begin{aligned}
v_t(x)&=\nabla H\bigl(\nabla_\g u_f(t,x)\bigr)\\
&=k\,\|\nabla_\g u_f(t,x)\|_\g^{k-2}
 \,\nabla_\g u_f(t,x).
\end{aligned}
\]
where $\exp_x^{-1}(y^\ast_t(x))$ in the definition of $v_t$ denotes the initial velocity of this geodesic. In $\R^n$
and on $\mathbb S^2$, uniqueness of the minimizing geodesic is equivalent to $y^\ast_t(x)\notin\Cut(x)$, and
$\exp_x^{-1}$ is the usual inverse of the exponential map.
\end{proposition}
\end{tcolorbox}
\begin{proof}
\emph{Regularity.} In $\R^n$, this is Lemma~\ref{lem:loc}. On a compact $\M$ of diameter $D$, the function
$x\mapsto b_t(x,y)$ is Lipschitz with constant $c_kk^\ast D^{k^\ast-1}t^{-1/(k-1)}$, uniformly in $y$; $u_f(t,\cdot)$,
an infimum of such functions, is Lipschitz. Rademacher's theorem, applied in charts, gives differentiability almost
everywhere.
\emph{Extremal point.} $Y_t(x)$ is nonempty: on a compact $\M$, $y\mapsto f(y)+b_t(x,y)$ is continuous; in $\R^n$,
this is Lemma~\ref{lem:loc}. Let $x$ be a point of differentiability and $y\in Y_t(x)$. For all $x'$,
$u_f(t,x')\le f(y)+b_t(x',y)$, with equality at $x'=x$. The function $\zeta=f(y)+b_t(\cdot,y)-u_f(t,\cdot)$ is
therefore nonnegative and vanishes at $x$: near $x$, $b_t(\cdot,y)$ is bounded from below by the function
$u_f(t,\cdot)-f(y)$, which is differentiable at $x$ and equal to $b_t(x,y)$ at this point. Hence $b_t(\cdot,y)$ admits
a subgradient at $x$.
If $y=x$, $b_t(\cdot,x)=c_kd(\cdot,x)^{k^\ast}t^{-1/(k-1)}$ is differentiable at $x$ with zero gradient since
$k^\ast>1$; hence $\nabla_\g u_f(t,x)=0$ and $v_t(x)=0=\nabla H(0)$.
If $y\neq x$, let $\gamma$ be any minimizing geodesic from $x$ to $y$, with initial velocity $w$, so that
$\exp_x(w)=y$ and $\|w\|=d(x,y)$. For $s>0$ small enough, $\gamma(s)$ lies within the injectivity radius of $x$ and
differs from $x$, so $z\mapsto d(z,\gamma(s))$ is smooth near $x$, with gradient $-w/\|w\|$ at $x$. By the triangle
inequality, $d(z,y)\le d(z,\gamma(s))+d(\gamma(s),y)$, with equality at $z=x$. Since $r\mapsto c_kr^{k^\ast}$ is
increasing, the function
\[
\psi(z)=\frac{c_k}{t^{1/(k-1)}}\big(d(z,\gamma(s))+d(\gamma(s),y)\big)^{k^\ast}
\]
is smooth near $x$, satisfies $b_t(\cdot,y)\le\psi$ with equality at $x$, and
\[
\begin{aligned}
\nabla\psi(x)
&=\frac{c_kk^\ast}{t^{1/(k-1)}}
 d(x,y)^{k^\ast-2}(-w)\\
&=c_kk^\ast\left\|\frac wt\right\|^{k^\ast-2}
 \left(-\frac wt\right)\\
&=\nabla L\!\left(-\frac wt\right).
\end{aligned}
\]
using $\|w\|=d(x,y)$ and $t^{-1/(k-1)}=t^{1-k^\ast}=t^{2-k^\ast}\cdot t^{-1}$. Near $x$, we thus have
$u_f(t,\cdot)-f(y)\le b_t(\cdot,y)\le\psi$, with equality at $x$ in both inequalities. The functions
$u_f(t,\cdot)-f(y)$ and $\psi$ are differentiable at $x$, and their difference is nonpositive and vanishes at $x$, so
their gradients at $x$ coincide:
\[
\nabla_\g u_f(t,x)=\nabla L\Big(-\frac wt\Big).
\]
By Lemma~\ref{lem:conj}, $-\frac wt=\nabla H(\nabla_\g u_f(t,x))$, hence $w=-t\,\nabla H(\nabla_\g u_f(t,x))$ and
$y=\exp_x(w)$. The initial velocity $w$ of every minimizing geodesic from $x$ to every $y\in Y_t(x)$ is therefore
determined by $\nabla_\g u_f(t,x)$: $Y_t(x)$ is a singleton, the minimizing geodesic from $x$ to $y^\ast_t(x)$ is
unique, and $v_t(x)=-w/t=\nabla H(\nabla_\g u_f(t,x))$.
\end{proof}
\paragraph{Why the CFM loss does not learn the potential.}\label{supp:cfm}
We work in $\R^n$, with the path of Lipman et al.~\cite{lipman2023flow} ($\sigma_{\min}=0$) and the independent coupling
$\pi=p_0\otimes q$. By the remark following Lemma~\ref{lem:gap}, $\Lcal_{\rm CFM}(f)=\E_{t,\pi}t^{-2}|y^\ast_t(x_t)-x_0|^2$.
With the independent coupling, $x_t$ does not determine $x_0$: there remains an irreducible residual
$\int t^{-2}\E\Var(x_0\mid x_t)\dd t$, common to all fields, including the field of Lipman et al. The issue lies
elsewhere: the marginal CFM path need not be generated by any morphological flow in the regular regime, as
Proposition~\ref{prop:contre} shows on an explicit example.
\begin{definition}[Regular regime]
A potential $f$ is in the regular regime if, for $p_0$-almost every $x_0$ and every $t\in(0,1)$,
$Y_t(x_0+tv_0(x_0))=\{x_0\}$. Its trajectories are then the segments $\psi_t(x_0)=x_0+tv_0(x_0)$, traveled at
constant speed and without crossing (two distinct starting points cannot both be the unique extremal point of the same
point). By Theorem~\ref{thm:gen}(ii), the optimal potential is in the regular regime.
\end{definition}
\begin{tcolorbox}[colback=gray!3,colframe=gray!55]
\begin{proposition}[Marginal CFM path outside the class]\label{prop:contre}
In dimension $1$, let $p_0=\mathcal N(0,1)$, $q=\mathcal N(0,\sigma^2)$ with $\sigma>0$, and the independent coupling.
The marginal CFM path is $p^{\rm CFM}_t=\mathcal N\big(0,(1-t)^2+t^2\sigma^2\big)$, and no potential in the regular
regime generates this path.
\end{proposition}
\end{tcolorbox}
\begin{proof}
By independence, $x_t$ is a centered Gaussian of variance $(1-t)^2+t^2\sigma^2$. Assume that a potential in the regular
regime generates this path. Then $T_\#p_0=q$ with $T=\mathrm{id}+v_0$ and, for all $t$,
\[
\begin{aligned}
&\Var\big((1-t)x_0+tT(x_0)\big)\\
&\quad=(1-t)^2+t^2\sigma^2\\
&\qquad+2t(1-t)\Cov\big(x_0,T(x_0)\big).
\end{aligned}
\]
hence $\Cov(x_0,T(x_0))=0$. However, the absence of crossings forces $T$ to be nondecreasing on a set of full
$p_0$-measure: for $x_0<x_0'$ in this set, the function $D(t)=(1-t)(x_0'-x_0)+t\big(T(x_0')-T(x_0)\big)$ is affine in
$t$, positive at $t=0$, and does not vanish on $[0,1)$ (no crossing), so it is positive on $[0,1)$, and letting
$t\to1$ gives $T(x_0)\le T(x_0')$. Let $x_0'$ be an independent copy of $x_0$. Then
\[
\Cov\big(x_0,T(x_0)\big)=\tfrac12\,\E\big[(x_0-x_0')\big(T(x_0)-T(x_0')\big)\big],
\]
whose integrand is nonnegative; it vanishes only if $T(x_0)=T(x_0')$ almost surely, i.e., if $T$ is $p_0$-almost
surely constant. Then $T_\#p_0$ would be a Dirac mass, whereas $q=\mathcal N(0,\sigma^2)$ with $\sigma>0$:
contradiction.
\end{proof}
\begin{remark}
For instance, the nondecreasing map $T(x)=\sigma x$ transports $p_0$ onto $q$ along the path
$\mathcal N\big(0,((1-t)+t\sigma)^2\big)$, whose variance exceeds that of $p^{\rm CFM}_t$ by $2t(1-t)\sigma>0$ for
$t\in(0,1)$. In such cases, minimizing $\Lcal_{\rm CFM}$ over the class of morphological fields amounts to projecting
the marginal CFM field onto a class that does not contain it; nothing in this loss forces the resulting flow to
transport $p_0$ onto $q$, and the ablation of Table~\ref{tab:supp_abl_poids} confirms that it degrades generation. We
keep its residual, whose vanishing is measured by the gap $G_t$ (Lemma~\ref{lem:gap}).
\end{remark}
%---------------------------------------------------------------------
\subsection{What We Minimize, and Why}
%---------------------------------------------------------------------
\begin{lemma}[The erosion gap measures the CFM residual]\label{lem:gap}
Let $(x_0,x_1)$ be joined by a unique minimizing geodesic, $x_t$ the geodesic interpolation and $t\in(0,1]$. Then
$G_t(f;x_0,x_1)=f(x_0)+t\,c(x_0,x_1)-u_f(t,x_t)\ge0$, with equality if and only if $x_0\in Y_t(x_t)$. If $Y_t(x_t)$ is
a singleton, $G_t=0$ is equivalent to $v_t(x_t)=\dot x_t$.
\end{lemma}
\begin{proof}
Since $d(x_t,x_0)=t\,d(x_0,x_1)$, $b_t(x_t,x_0)=t\,L(d(x_0,x_1))=t\,c(x_0,x_1)$. Hence $G_t$ is the value at $y=x_0$ of
$y\mapsto f(y)+b_t(x_t,y)$ minus its minimum: $G_t\ge0$, with equality if and only if $x_0$ is a minimizer.
For $t<1$, $x_t$ is an interior point of the minimizing geodesic $\gamma$ from $x_0$ to $x_1$; $x_0\notin\Cut(x_t)$ and
$\exp_{x_t}^{-1}(x_0)=-t\,\dot\gamma(t)$, so $-\frac1t\exp_{x_t}^{-1}(x_0)=\dot x_t$. For $t=1$, this is the assumption
that the geodesic is unique. If $Y_t(x_t)=\{y^\ast\}$ and $G_t=0$, then $y^\ast=x_0$ and $v_t(x_t)=\dot x_t$.
Conversely, if $v_t(x_t)=\dot x_t$, then $\exp_{x_t}^{-1}(y^\ast)=-t\dot x_t=\exp_{x_t}^{-1}(x_0)$, hence $y^\ast=x_0$
and $G_t=0$.
\end{proof}
\begin{remark}[Connection with the CFM residual in $\mathbb{R}^n$]
With the path of Lipman et al.~\cite{lipman2023flow} ($\sigma_{\min}=0$), $x_t=(1-t)x_0+tx_1$ and the conditional target is
$x_1-x_0=(x_t-x_0)/t$. For any measurable selection $y^\ast_t$,
\[
v_t(x_t)-(x_1-x_0)=\frac{x_t-y^\ast_t(x_t)}{t}-\frac{x_t-x_0}{t}=\frac{x_0-y^\ast_t(x_t)}{t},
\]
hence $\Lcal_{\rm CFM}(f)=\E_{t,\pi}\,t^{-2}|y^\ast_t(x_t)-x_0|^2$: the CFM residual vanishes if and only if the extremal
point at $x_t$ is the starting point, which is also what $G_t$ measures.
\end{remark}
\begin{tcolorbox}
\begin{theorem}[The loss is the Kantorovich semi-dual, up to sign]\label{thm:dual}
Assume that $\M$ is compact, or that $\M=\R^n$ and $p_0,q$ have finite moments of order $k^\ast$, and let
$W_c(p_0,q)=\min_{\pi\in\Pi(p_0,q)}\int c\dd\pi$. Denote by $\F$ the set of Borel functions $f:\M\to\R\cup\{+\infty\}$
with $f\in L^1(p_0)$ and $u_f(1,\cdot)\in L^1(q)$. Then $\Lcal_1(f)\ge-W_c(p_0,q)$ for every $f\in\F$, and this minimum
is attained by a $c$-convex function $f^\ast\in\F$, \ie, one of the form $f^\ast(x)=\sup_y\{\zeta(y)-c(x,y)\}$ for some
$\zeta:\M\to\R\cup\{-\infty\}$. The loss depends only on the marginal distributions, not on the coupling used to draw
$(x_0,x_1)$.
\end{theorem}
\end{tcolorbox}
\begin{proof}
We have $c(x,y)\le a(x)+b(y)$, with $a,b$ bounded if $\M$ is compact, and $a(x)=c_k2^{k^\ast-1}|x|^{k^\ast}\in L^1(p_0)$,
$b(y)=c_k2^{k^\ast-1}|y|^{k^\ast}\in L^1(q)$ in $\R^n$; an optimal plan exists and $W_c<\infty$ (\cite{villani2009},
Thm.~4.1).

\emph{Lower bound.} Let $f\in\F$ and $\pi$ be an optimal plan. For all $x,y$, $u_f(1,y)\le f(x)+c(x,y)$. The three
functions $(x,y)\mapsto f(x)$, $c(x,y)$, $u_f(1,y)$ are $\pi$-integrable (marginals $p_0$ and $q$, and
$\int c\dd\pi=W_c<\infty$); integrating against $\pi$ gives $\int u_f(1,\cdot)\dd q\le\int f\dd p_0+W_c$, \ie,
$\Lcal_1(f)\ge-W_c$.

\emph{Attainment.} Since $c$ is symmetric, $u_f(1,\cdot)$ is the $c$-transform $f^c(y)=\inf_x\{f(x)+c(x,y)\}$ of $f$
in the convention of~\cite{villani2009}, hence $-\Lcal_1(f)=\int f^c\dd q-\int f\dd p_0$. This right-hand side is the
dual functional of Theorem~5.10 of~\cite{villani2009}. Under the bound $c\le a+b$ above, item (iii) of this theorem
states that its supremum, equal to $W_c$, is attained by a $c$-convex function $f^\ast$ such that $f^\ast\in L^1(p_0)$,
$(f^\ast)^c\in L^1(q)$ and $\int(f^\ast)^c\dd q-\int f^\ast\dd p_0=W_c$. Since $(f^\ast)^c=u_{f^\ast}(1,\cdot)$, we
have $f^\ast\in\F$ and $\Lcal_1(f^\ast)=-W_c$.

Finally, $\Lcal_1$ is the sum of an expectation under $p_0$ and an expectation under $q$.
\end{proof}
\begin{corollary}[The optimal coupling comes for free]\label{cor:otimplicite}
Under the assumptions of Theorem~\ref{thm:dual}, let $f^\ast\in\F$ attain the minimum of $\Lcal_1$, and let $\pi^\ast$
be an optimal plan. Then $\pi^\ast$ is concentrated on
\[
\Gamma=\big\{(x,y):\ f^\ast(x)<+\infty,\ f^\ast(x)+c(x,y)=u_{f^\ast}(1,y)\big\},
\]
that is, $G_1(f^\ast;x_0,x_1)=0$ and $x_0\in Y_1(x_1)$ for $\pi^\ast$-almost every $(x_0,x_1)$. For $(x,y)\in\Gamma$,
$z\mapsto f^\ast(z)+c(z,y)$ attains its global minimum at $x$. If $f^\ast$ is $c$-convex, $\Gamma$ is a Borel set.
\end{corollary}
\begin{proof}
\emph{Support.} The function $g(x,y)=f^\ast(x)+c(x,y)-u_{f^\ast}(1,y)\ge0$ is $\pi^\ast$-integrable, with integral
$\Lcal_1(f^\ast)+W_c=0$; it therefore vanishes $\pi^\ast$-almost everywhere, and $f^\ast(x)<\infty$ $\pi^\ast$-almost
everywhere. For $(x,y)\in\Gamma$ and every $z$, $f^\ast(z)+c(z,y)\ge u_{f^\ast}(1,y)=f^\ast(x)+c(x,y)$. The identity
$g(x_0,x_1)=0$ is $G_1(f^\ast;x_0,x_1)=0$, and Lemma~\ref{lem:gap} gives $x_0\in Y_1(x_1)$ (its first part, for
$t=1$, does not require the geodesic to be unique).

\emph{Measurability.} If $f^\ast$ is $c$-convex, it is equal to $\sup_y\{u_{f^\ast}(1,y)-c(\cdot,y)\}$
(\cite{villani2009}, Prop.~5.8), a supremum of continuous functions: it is lower semicontinuous; $u_{f^\ast}(1,\cdot)$,
an infimum of continuous functions, is upper semicontinuous. The function $g$ is therefore lower semicontinuous with
values in $[0,+\infty]$ (where it is defined, \ie, outside the $q$-negligible Borel set $\{u_{f^\ast}(1,y)=-\infty\}$),
and $\Gamma=\{g=0\}\cap\{f^\ast<\infty\}$ is a Borel set.
\end{proof}
%---------------------------------------------------------------------
\subsection{Training in Practice}
%---------------------------------------------------------------------

\begin{tcolorbox}
\begin{proposition}[Batch candidates bound the estimated loss]\label{prop:borne}
Let $B$ be a batch of $|B|$ pairs, with sources $(x_0^i)_{1\le i\le|B|}\sim p_0$ and targets
$(x_1^j)_{1\le j\le|B|}\sim q$; let $J_j(y)=f_\theta(y)+c(y,x_1^j)$, and let $\hat y_j$ be the best point found among
candidates that include all $x_0^i$. Then
$\hat\Lcal_1=\frac1{|B|}\sum_if_\theta(x_0^i)-\frac1{|B|}\sum_jJ_j(\hat y_j)\ge-W_c(\hat p_0,\hat q)$, where
$\hat p_0,\hat q$ are the empirical distributions of the batch.
\end{proposition}
\end{tcolorbox}
\begin{proof}
Since the best point found is kept, $J_j(\hat y_j)\le\min_i\{f_\theta(x_0^i)+c(x_0^i,x_1^j)\}=\varphi^c_j$, the discrete
$c$-transform of $\varphi_i=-f_\theta(x_0^i)$. For every coupling $(\pi_{ij})$ of the empirical distributions,
$\varphi_i+\varphi^c_j\le c(x_0^i,x_1^j)$ gives
\[
\begin{aligned}
-\hat\Lcal_1
&\le\frac1{|B|}\sum_i\varphi_i+\frac1{|B|}\sum_j\varphi^c_j\\
&=\sum_{i,j}\pi_{ij}\bigl(\varphi_i+\varphi^c_j\bigr)\\
&\le\sum_{i,j}\pi_{ij}\,c(x_0^i,x_1^j),
\end{aligned}
\]
and we take $\pi$ optimal.
\end{proof}
\begin{tcolorbox}[colback=gray!3,colframe=gray!55]
\begin{proposition}[Gradient via Danskin's theorem]\label{prop:danskin}
Let $f_\theta$ be of class $C^1$ in $\theta$, with $(\theta,y)\mapsto\nabla_\theta f_\theta(y)$ continuous and (in
$\R^n$) a locally bounded Lipschitz constant $\Lip(\theta)$. If $Y_t(x)=\{y^\ast\}$, then
$\nabla_\theta u_{f_\theta}(t,x)=\nabla_\theta f_\theta(y)\big|_{y=y^\ast}$.
\end{proposition}
\end{tcolorbox}
\begin{proof}
On a compact $\M$, $u_{f_\theta}(t,x)=\min_{y\in\M}F(\theta,y)$ with $F(\theta,y)=f_\theta(y)+b_t(x,y)$ continuous and
$\nabla_\theta F=\nabla_\theta f_\theta$ continuous. In $\R^n$, if $\Lip(\theta)\le\bar\Lip$ on a neighborhood $V$ of
$\theta_0$, Lemma~\ref{lem:loc} allows us to restrict the minimum to the compact ball of center $x$ and radius
$(\bar\Lip/c_k)^{k-1}t$, for all $\theta\in V$. In both cases, Danskin's theorem~\cite{danskin1967} gives the
directional derivative $\partial_hu=\min_{y\in Y_t(x)}\langle\nabla_\theta F(\theta,y),h\rangle$; if the minimizer is
unique, $u$ is differentiable in $\theta$, with gradient $\nabla_\theta F(\theta,y^\ast)$.
\end{proof}
%---------------------------------------------------------------------
\subsection{Generation: A Single Euler Step Is Exact}\label{supp:gen}
%---------------------------------------------------------------------
We give here the complete statement of the generation theorem, for every $k>1$, and a self-contained proof. Item (i) is
classical: Gangbo--McCann~\cite{gangbo1996} in $\R^n$, McCann~\cite{mccann2001} for $k=2$ and
Fathi--Figalli~\cite{fathi2010} for any $k$ on a manifold. Items (i), (ii)--(iii) and (v) below are items (i), (ii) and
(iii) of the main paper.

\paragraph{Setting.}
In this subsection, $(\M,\g)$ is either the Euclidean space $\R^n$ or a smooth, compact, connected Riemannian manifold
without boundary; $d=d_\g$ is the distance, $\mathrm{vol}$ the volume measure (Lebesgue measure in $\R^n$), $\exp_x$ the
exponential map and $\mathrm{inj}(x)>0$ the injectivity radius at $x$ (infinite in $\R^n$). In the compact case,
$D<\infty$ is the diameter. We use two classical facts:
\begin{enumerate}[label=(R\arabic*),leftmargin=*]
\item\label{R1} if $p\neq x$ and $d(x,p)<\mathrm{inj}(x)$, then $z\mapsto d(z,p)$ is smooth near $x$, with gradient
$-\exp_x^{-1}(p)/d(x,p)$ at $x$;
\item\label{R2} a minimizing curve parametrized at constant speed is a geodesic, hence smooth; two geodesics with the
same position and velocity at one instant coincide on their common interval.
\end{enumerate}
For $f:\M\to\R\cup\{+\infty\}$, not identically $+\infty$, and $t>0$, the erosion $u_f(t,x)=\inf_y\{f(y)+b_t(x,y)\}$
takes values in $\R\cup\{-\infty\}$, and $Y_t(x)$ is the set of its minimizers. When $Y_t(x)=\{y^\ast_t(x)\}$ and $x$
and $y^\ast_t(x)$ are joined by a unique minimizing geodesic, whose initial velocity is denoted by
$\exp_x^{-1}(y^\ast_t(x))$, the morphological velocity is $v_t(x)=-\frac1t\exp_x^{-1}(y^\ast_t(x))$.
Following~\cite{villani2009}, $f$ is \emph{$c$-convex} if $f(x)=\sup_y\{\zeta(y)-c(x,y)\}$ for some
$\zeta:\M\to\R\cup\{-\infty\}$. Since $c$ is symmetric, the $c$-transform of $f$ in the sense of~\cite{villani2009},
$f^c(y)=\inf_x\{f(x)+c(x,y)\}$, is exactly $u_f(1,y)$, and $f$ is $c$-convex if and only if
$f(x)=\sup_y\{u_f(1,y)-c(x,y)\}$ (\cite{villani2009}, Prop.~5.8). As in Theorem~\ref{thm:dual}, $\F$ denotes the set of
Borel functions $f:\M\to\R\cup\{+\infty\}$ with $f\in L^1(p_0)$ and $u_f(1,\cdot)\in L^1(q)$, and, for $f\in\F$,
\[
\begin{aligned}
\Lcal_1(f)&=\int f\dd p_0-\int u_f(1,\cdot)\dd q,\\
W_c(p_0,q)&=\min_{\pi\in\Pi(p_0,q)}\int c\dd\pi .
\end{aligned}
\]

\paragraph{Assumptions (H).}
$k>1$; $p_0$ and $q$ are Borel probability measures on $\M$, with $p_0\ll\mathrm{vol}$; either $\M=\R^n$ and
$\int|x|^{k^\ast}\dd p_0+\int|y|^{k^\ast}\dd q<\infty$, or $\M$ is compact.

\begin{tcolorbox}[colback=gray!3,colframe=gray!55]
\begin{theorem}[One-step generation is exact]\label{thm:gen}
Under (H), $\min_\F\Lcal_1=-W_c(p_0,q)$, and this minimum is attained by a $c$-convex function. Let $f^\ast\in\F$ be a
$c$-convex minimizer. Then $f^\ast$ is finite and differentiable $p_0$-almost everywhere; we set, $p_0$-almost
everywhere, $v_0(x)=\nabla H(\nabla_\g f^\ast(x))$ (the initial field of the flow, which is not the morphological
velocity $v_t$ taken at $t=0$), $\psi_t(x)=\exp_x(t\,v_0(x))$ for $t\in[0,1]$, and $T=\psi_1$. Then:
\begin{enumerate}[label=(\roman*),leftmargin=*]
\item \emph{(Transport)} the optimal plan is unique and equals $(\mathrm{id},T)_\#p_0$; for $p_0$-almost every $x$,
$s\mapsto\psi_s(x)$, $s\in[0,1]$, is the unique minimizing geodesic from $x$ to $T(x)$;
\item \emph{(Extremal point along trajectories)} for $p_0$-almost every $x_0$ and every $t\in(0,1)$,
$Y_t(\psi_t(x_0))=\{x_0\}$, the minimizing geodesic from $\psi_t(x_0)$ to $x_0$ is unique, and
$v_t(\psi_t(x_0))=\frac{\mathrm d}{\mathrm dt}\psi_t(x_0)$;
\item \emph{(No crossing)} there is a Borel set $S$ with $p_0(S)=1$ such that, for every $t\in[0,1)$, $\psi_t$ is
injective on $S$;
\item \emph{(Continuity equation)} $p_t=(\psi_t)_\#p_0$ satisfies $p_1=q$ and, for every $\zeta\in C^\infty_c(\M)$,
the function $t\mapsto\int\zeta\dd p_t$ is Lipschitz on $[0,1]$, with derivative at every $t\in(0,1)$
\[
\begin{aligned}
\frac{\mathrm d}{\mathrm dt}\int\zeta\dd p_t
=\int_S\big\langle&\nabla_\g\zeta(\psi_t(x_0)),\\
&v_t(\psi_t(x_0))\big\rangle\dd p_0(x_0),
\end{aligned}
\]
\ie, $\partial_tp_t+\mathrm{div}_\g(p_tv_t)=0$ in the weak sense, where $p_tv_t$ denotes the vector measure
$(\psi_t)_\#\big((v_t\circ\psi_t)\,p_0\big)$;
\item \emph{(Exact Euler)} for every $N\ge1$ and $x_0\in S$, the scheme
$x^{(m+1)}=\exp_{x^{(m)}}\big(\frac1Nv_{m/N}(x^{(m)})\big)$, $x^{(0)}=x_0$, where $v_0$ is the initial field above, is
well defined and gives $x^{(m)}=\psi_{m/N}(x_0)$ for $m=0,\dots,N$; in particular, $x^{(N)}=T(x_0)$ follows $q$ for
every $N$, and a single step ($N=1$) suffices.
\end{enumerate}
\end{theorem}
\end{tcolorbox}

\paragraph{Lemmas used in the proof.}
Besides Lemmas~\ref{lem:conj} and~\ref{lem:concat}, Theorem~\ref{thm:dual} and Corollary~\ref{cor:otimplicite}, the
proof uses the following two lemmas.

\begin{lemma}[Partner at a point of differentiability]\label{lem:part}
Let $f:\M\to\R\cup\{+\infty\}$, $x\in\M$ such that $f$ is finite near $x$ and differentiable at $x$, and $y\in\M$ such
that $z\mapsto f(z)+c(z,y)$ has a local minimum at $x$. Then $x$ and $y$ are joined by a unique minimizing geodesic,
with initial velocity $\nabla H(\nabla_\g f(x))$; in particular, $y=\exp_x\big(\nabla H(\nabla_\g f(x))\big)$.
\end{lemma}
\begin{proof}
\emph{Case $y=x$.} $c(\cdot,x)=c_kd(\cdot,x)^{k^\ast}=o(d(\cdot,x))$ since $k^\ast>1$; hence $c(\cdot,x)$ is
differentiable at $x$ with zero gradient, and $f+c(\cdot,x)$, differentiable at $x$ and locally minimal at $x$, has a
zero gradient at $x$: $\nabla_\g f(x)=0$, and the formula indeed gives the constant geodesic, with velocity
$\nabla H(0)=0$.

\emph{Case $y\neq x$.} Let $\gamma:[0,1]\to\M$ be any minimizing geodesic from $x$ to $y$, with initial velocity $w$,
$\|w\|=d(x,y)>0$. Let $s\in(0,1)$ with $s\,d(x,y)<\mathrm{inj}(x)$, and $p=\gamma(s)\neq x$. By~\ref{R1},
$z\mapsto d(z,p)$ is smooth near $x$, with gradient $-w/\|w\|$ at $x$. By the triangle inequality,
$d(z,y)\le d(z,p)+d(p,y)$, with equality at $z=x$. The function $\Psi(z)=c_k\big(d(z,p)+d(p,y)\big)^{k^\ast}$ is
therefore smooth near $x$ (its argument is $\ge d(p,y)>0$ there), satisfies $c(\cdot,y)\le\Psi$ near $x$ with equality
at $x$, and $\nabla_\g\Psi(x)=c_kk^\ast d(x,y)^{k^\ast-1}(-w/\|w\|)=\nabla L(-w)$. Near $x$,
$f+\Psi\ge f+c(\cdot,y)\ge f(x)+c(x,y)=f(x)+\Psi(x)$: the function $f+\Psi$, differentiable at $x$, has a local minimum
there, hence $\nabla_\g f(x)=-\nabla L(-w)=\nabla L(w)$ by oddness. By Lemma~\ref{lem:conj},
$w=\nabla H(\nabla_\g f(x))$, which does not depend on $\gamma$: all minimizing geodesics from $x$ to $y$ have the same
initial velocity and coincide by~\ref{R2}, and $y=\exp_x(w)$.
\end{proof}

\begin{lemma}[Finiteness and differentiability]\label{lem:diff}
Under (H), let $f^\ast\in\F$ be $c$-convex. There is a Borel set $\mathcal D$ with $p_0(\mathcal D)=1$ such that, for
every $x\in\mathcal D$, $f^\ast$ is finite near $x$ and differentiable at $x$; $\nabla_\g f^\ast$ is Borel on
$\mathcal D$.
\end{lemma}
\begin{proof}
\emph{Compact case.} $f^\ast$ is finite $p_0$-almost everywhere (since $f^\ast\in L^1(p_0)$), hence at some point
$x_0$. For every $y$, $u_{f^\ast}(1,y)\le f^\ast(x_0)+c(x_0,y)<\infty$, and $c(\cdot,y)$ is Lipschitz with constant
$\Lambda=c_kk^\ast D^{k^\ast-1}$, uniformly in $y$. Hence $f^\ast=\sup_y\{u_{f^\ast}(1,y)-c(\cdot,y)\}$, finite at
$x_0$, is finite everywhere and $\Lambda$-Lipschitz. By Rademacher's theorem in charts, the set $\mathcal D$ of its
points of differentiability has a $\mathrm{vol}$-negligible, hence $p_0$-negligible, complement.

\emph{Case $\R^n$.} The function $\varphi=-f^\ast=\inf_y\{c(\cdot,y)-u_{f^\ast}(1,y)\}$ (infimum over the $y$ where
$u_{f^\ast}(1,y)>-\infty$) is $c$-concave in the sense of Gangbo--McCann (\cite{gangbo1996}, Def.~1.1), with values in
$\R\cup\{-\infty\}$ and not identically $-\infty$, for $c(x,y)=h(x-y)$ with $h(z)=c_k|z|^{k^\ast}$. This function
satisfies assumptions (H2)--(H4) of~\cite{gangbo1996}. (H4): $h$ is convex, since $k^\ast>1$. (H3):
$h(z)/|z|=c_k|z|^{k^\ast-1}\to+\infty$. (H2) (cone condition): let $r<\infty$, $\theta\in(0,\pi)$ and $p\neq0$; take
the direction $\hat z=-p/|p|$ and the cone $K(r,\theta,\hat z,p)$ with apex $p$, height $r$ and angle $\theta$, every
point $x$ of which satisfies $\langle\hat z,x-p\rangle\ge|x-p|\cos(\theta/2)$ and $\langle\hat z,x-p\rangle\le r$,
hence $|x-p|\le r/\cos(\theta/2)$. Then
\[
\begin{aligned}
|x|^2&=|p|^2-2|p|\langle\hat z,x-p\rangle+|x-p|^2\\
&\le|p|^2-|x-p|\big(2|p|\cos(\theta/2)-|x-p|\big)\le|p|^2
\end{aligned}
\]
as soon as $|p|\ge r/(2\cos^2(\theta/2))$; since $h$ is radial and increasing, $h(x)\le h(p)$ on the cone, and $h$
attains its maximum there at $p$. By~\cite{gangbo1996}, Thm.~3.3, there is a convex set $K$ with interior $\Omega$ such
that $\Omega\subset\{\varphi>-\infty\}\subset K$, and $\varphi$ is locally Lipschitz on $\Omega$. By Rademacher's
theorem, $\varphi$ is differentiable Lebesgue-almost everywhere on the open set $\Omega$; let $\mathcal D$ be the set
of these points of differentiability in $\Omega$. The complement of $\mathcal D$ in $\{\varphi>-\infty\}$ is contained in
$(\Omega\setminus\mathcal D)\cup\partial K$, which is Lebesgue-negligible (the boundary of a convex set is), hence
$p_0$-negligible; and $p_0(\{\varphi>-\infty\})=p_0(\{f^\ast<\infty\})=1$. For $x\in\mathcal D\subset\Omega$, $f^\ast$ is
finite on the open neighborhood $\Omega$ of $x$.

\emph{Measurability.} The set of points of differentiability of a locally Lipschitz function on an open set is a Borel
set, and its gradient there is a limit of continuous difference quotients, hence Borel.
\end{proof}

\begin{proof}[Proof of Theorem~\ref{thm:gen}]
By Theorem~\ref{thm:dual}, $\min_\F\Lcal_1=-W_c$, attained by a $c$-convex function. Let $f^\ast$ be a $c$-convex
minimizer in $\F$, and $\mathcal D$ be given by Lemma~\ref{lem:diff}: $f^\ast$ is finite and differentiable on
$\mathcal D$, which has full $p_0$-measure, and $v_0$, $\psi_t$, $T$ are defined and Borel on $\mathcal D$ ($\exp$ is
continuous).

\emph{(i).} Let $x\in\mathcal D$ and $y$ with $(x,y)\in\Gamma$, with $\Gamma$ given by Corollary~\ref{cor:otimplicite}. By this corollary, $z\mapsto f^\ast(z)+c(z,y)$ is
minimal at $x$; by Lemma~\ref{lem:part}, $x$ and $y$ are joined by a unique minimizing geodesic, with initial velocity
$v_0(x)$, and $y=T(x)$. Let $\pi$ be an optimal plan: it is concentrated on $\Gamma$ (Corollary~\ref{cor:otimplicite}), and its
first marginal $p_0$ is concentrated on $\mathcal D$; it is therefore concentrated on the Borel graph
$\{(x,T(x)):x\in\mathcal D\}$, which forces $\pi=(\mathrm{id},T)_\#p_0$. The optimal plan is thus unique, and
$T_\#p_0=q$.

Let $S=\{x\in\mathcal D:(x,T(x))\in\Gamma\}$, a Borel set; since $(\mathrm{id},T)_\#p_0$ is optimal, hence concentrated
on $\Gamma$, $p_0(S)=1$. For $x_0\in S$, the curve $\gamma(s)=\psi_s(x_0)$, $s\in[0,1]$, is the unique minimizing
geodesic from $x_0$ to $x_1=T(x_0)$; we set $x_t=\gamma(t)$, so that $d(x_0,x_t)=t\,d(x_0,x_1)$ and
$d(x_t,x_1)=(1-t)\,d(x_0,x_1)$. The membership $(x_0,x_1)\in\Gamma$ reads $f^\ast(x_0)+c(x_0,x_1)=u_{f^\ast}(1,x_1)$.

\emph{(ii).} Let $x_0\in S$ and $t\in(0,1)$. For every $y$, by definition of the erosion and then by
Lemma~\ref{lem:concat},
\[
\begin{aligned}
u_{f^\ast}(1,x_1)&\le f^\ast(y)+c(y,x_1)\\
&\le f^\ast(y)+b_t(y,x_t)+b_{1-t}(x_t,x_1),
\end{aligned}
\]
with equality for $y=x_0$ (the first because $(x_0,x_1)\in\Gamma$, the second by the equality case of
Lemma~\ref{lem:concat}, $x_t$ being at the fraction $t$ of a minimizing geodesic). Here $u_{f^\ast}(1,x_1)$ is finite.
Hence, for every $y$,
\[
\begin{aligned}
f^\ast(x_0)+b_t(x_t,x_0)&=u_{f^\ast}(1,x_1)-b_{1-t}(x_t,x_1)\\
&\le f^\ast(y)+b_t(x_t,y),
\end{aligned}
\]
and $x_0\in Y_t(x_t)$.

\emph{Uniqueness of the extremal point.} Let $y'\in Y_t(x_t)$: the chain of inequalities is an equality for $y=y'$,
and Lemma~\ref{lem:concat} gives $d(y',x_1)=d(y',x_t)+d(x_t,x_1)$ and $d(y',x_t)=\frac t{1-t}d(x_t,x_1)=d(x_0,x_t)$. If
$x_1=x_0$, then $x_t=x_0$ and $y'=x_0$. Otherwise, let $\sigma$ be a minimizing geodesic from $y'$ to $x_t$ and
$\eta:[0,1]\to\M$ the concatenation of $\sigma$ and $\gamma_{|[t,1]}$, parametrized at constant speed. Its length is
$d(y',x_1)$: $\eta$ is minimizing, hence a geodesic by~\ref{R2}, with speed $d(y',x_1)=d(x_0,x_1)$, that of $\gamma$,
and it passes through $x_t$ at the parameter $d(y',x_t)/d(y',x_1)=t$. It therefore coincides with $\gamma$ on $[t,1]$,
then on $[0,1]$ by~\ref{R2}, hence $y'=\eta(0)=x_0$. Thus $Y_t(x_t)=\{x_0\}$.

\emph{Uniqueness of the geodesic from $x_t$ to $x_0$.} Let $\sigma$ be a minimizing geodesic from $x_0$ to $x_t$, of
length $d(x_0,x_t)=t\,d(x_0,x_1)$. The concatenation of $\sigma$ and $\gamma_{|[t,1]}$, at constant speed, has length
$d(x_0,x_1)$: it is minimizing, hence a geodesic from $x_0$ to $x_1$; by uniqueness of the minimizing geodesic from
$x_0$ to $x_1$ (item (i)), it equals $\gamma$, and $\sigma=\gamma_{|[0,t]}$. The minimizing geodesic from $x_t$ to
$x_0$ is therefore unique ($\gamma_{|[0,t]}$ traversed backwards), with initial velocity $-t\,\dot\gamma(t)$; by
definition, $v_t(x_t)=-\frac1t\big(-t\,\dot\gamma(t)\big)=\dot\gamma(t)=\frac{\mathrm d}{\mathrm dt}\psi_t(x_0)$.

\emph{(iii).} If $x_0,x_0'\in S$ and $\psi_t(x_0)=\psi_t(x_0')=x$ with $t\in(0,1)$, then $\{x_0\}=Y_t(x)=\{x_0'\}$ by
(ii). For $t=0$, $\psi_0=\mathrm{id}$.

\emph{(iv).} $p_1=T_\#p_0=q$ by (i). Let $\zeta\in C^\infty_c(\M)$. For $x_0\in S$, $t\mapsto\zeta(\psi_t(x_0))$ is of
class $C^1$ on $[0,1]$, with derivative $\langle\nabla_\g\zeta(\psi_t(x_0)),\dot\gamma(t)\rangle$, bounded in absolute
value by $\|\nabla_\g\zeta\|_\infty\,d(x_0,T(x_0))$ since $\|\dot\gamma(t)\|=d(x_0,x_1)$. This bound is
$p_0$-integrable: it is $\le\|\nabla_\g\zeta\|_\infty D$ in the compact case; in $\R^n$, $d(x_0,T(x_0))\le|x_0|+|T(x_0)|$
and $\int|x_0|\dd p_0+\int|T(x_0)|\dd p_0=\int|x|\dd p_0+\int|y|\dd q<\infty$ under the moment assumption, since
$k^\ast>1$. By the mean value theorem and dominated convergence, $t\mapsto\int\zeta\dd p_t=\int_S\zeta(\psi_t(x_0))\dd
p_0(x_0)$ is Lipschitz on $[0,1]$ and differentiable at every $t\in(0,1)$, with derivative
$\int_S\langle\nabla_\g\zeta(\psi_t(x_0)),\dot\gamma(t)\rangle\dd p_0(x_0)$, and $\dot\gamma(t)=v_t(\psi_t(x_0))$ by
(ii).

\emph{(v).} Let $x_0\in S$ and $N\ge1$. By induction on $m\in\{0,\dots,N-1\}$: if $x^{(m)}=\gamma(m/N)$, then the
velocity $v_{m/N}(x^{(m)})$ is defined and equals $\dot\gamma(m/N)$ (by definition of $v_0$ if $m=0$, by (ii) if
$m\ge1$), and $x^{(m+1)}=\exp_{\gamma(m/N)}\big(\frac1N\dot\gamma(m/N)\big)=\gamma\big(\frac{m+1}N\big)$, since $\gamma$
is a geodesic. Hence $x^{(m)}=\psi_{m/N}(x_0)$ for $m=0,\dots,N$, and $x^{(N)}=T(x_0)$; since $p_0(S)=1$ and
$T_\#p_0=q$, $x^{(N)}$ follows $q$ when $x_0\sim p_0$, for every $N\ge1$.
\end{proof}
\begin{remark}[Zero CFM loss on the optimal coupling, at every $t$]\label{rem:cfmot}
Item (ii) extends Corollary~\ref{cor:otimplicite} to every $t\in(0,1)$: for $(x_0,x_1)$ drawn from the optimal plan
$(\mathrm{id},T)_\#p_0$, the geodesic interpolation $x_t$ satisfies $Y_t(x_t)=\{x_0\}$, hence $v_t(x_t)=\dot x_t$.
At the optimal potential, the morphological field thus cancels the conditional flow matching loss
$\E_{t,\pi^\ast}\|v_t(x_t)-\dot x_t\|_\g^2$ computed on the optimal coupling $\pi^\ast$, \ie, the OT-CFM loss with the
exact plan, even though training only used the independent coupling. This answers the failure of the CFM loss
(Sec.~3.1 of the main paper): under the independent coupling, the residual cannot vanish; the loss $\mathcal L_1$ finds by
itself the coupling on which it does.
\end{remark}
\begin{remark}[No inner problem at generation time]
Training requires solving the extremal-point problem, but one-step generation only uses $\nabla_\g f_\theta(x_0)$, an
approximation of $\nabla_\g f^\ast(x_0)$. A multi-step generation would have to compute $y^\ast_t$ at every step; by
item (v), it brings nothing at the optimal potential, and we verify numerically that it yields the same samples as a
single step at a 100 to 1{,}000 times higher cost (Table~\ref{tab:supp_abl_multipas}). Moreover, in $\R^n$, if $\hat y$
approximates $y^\ast_t(x)$, the computed velocity $\hat v_t=(x-\hat y)/t$ satisfies $|\hat v_t-v_t|=|\hat y-y^\ast_t(x)|/t$:
multi-step generation amplifies the error of the inner problem near $t=0$.
\end{remark}
%=====================================================================
\section{Explicit formulas for Euclidean spaces and the sphere}\label{supp:formules}
%=====================================================================
\paragraph{Euclidean case $\M=\mathbb{R}^n$, $k=2$.}
\[
\begin{aligned}
c(x,y)&=\tfrac14|x-y|^2,\\
u_f(1,x)&=\min_y\!\left\{f(y)+\tfrac14|x-y|^2\right\},\\
T(x)&=x+2\nabla f(x).
\end{aligned}
\]
The inner problem is minimized over the whole of $\R^n$; its gradient is $\nabla f(y)+\frac12(y-x)$.
\paragraph{Sphere $\M=\mathbb S^2\subset\mathbb{R}^3$, $k=2$.}
Points are unit vectors of $\R^3$ and $T_x\mathbb S^2=\{w:\langle w,x\rangle=0\}$. We use
\[
\begin{aligned}
d(x,y)&=\arccos\langle x,y\rangle\\
&=2\arcsin\!\left(\tfrac12|x-y|\right),\\
\exp_x(w)&=\cos(|w|)\,x
+\sin(|w|)\,\frac{w}{|w|}.
\end{aligned}
\]
\[
\begin{aligned}
\exp_x^{-1}(y)
&=\frac{\theta}{\sin\theta}
\bigl(y-\cos\theta\,x\bigr),\\
\theta&=d(x,y)\in[0,\pi),\qquad
\Cut(x)=\{-x\}.
\end{aligned}
\]
The second expression of $d$ is numerically stable for $y$ close to $x$; it is the one we use. The potential is a
network $\tilde f:\R^3\to\R$ evaluated on the ambient coordinates, and its Riemannian gradient is the tangent
projection of the ambient gradient:
\[
\nabla_\g f(x)=(I-xx^\top)\,\nabla\tilde f(x).
\]
The cost is $c(x,y)=d(x,y)^2/4$ and one-step generation reads
\[
T(z)=\exp_z\!\big(2\nabla_\g f(z)\big),\qquad z\sim\mathcal U(\mathbb S^2),
\]
the uniform distribution being sampled by normalizing a standard Gaussian of $\R^3$. The inner problem
$y\mapsto f(y)+d(x,y)^2/4$ is solved by Riemannian gradient descent, $y\leftarrow\exp_y(-a\,\mathrm{grad})$, where
$\mathrm{grad}$ is the tangent projection of the ambient gradient (Section~\ref{supp:details}). The geodesic
interpolation used by RFM is
\[
\begin{aligned}
x_t&=\cos(t\theta)\,x_0+\sin(t\theta)\,e,\\
\dot x_t&=\theta\bigl(-\sin(t\theta)\,x_0
+\cos(t\theta)\,e\bigr),\\
\theta&=d(x_0,x_1),\qquad
 e=\frac{\exp_{x_0}^{-1}(x_1)}{\theta}.
\end{aligned}
\]
%=====================================================================
\section{Implementation details}\label{supp:details}
%=====================================================================
\paragraph{Hardware.}
The experiments in $\R^2$ and on $\mathbb S^2$, including the official RFM run, as well as all timing measurements and
FID-50K evaluations, use one NVIDIA RTX~4090 Laptop GPU (16~GB). On CIFAR-10, our model is trained on 3 NVIDIA A100
GPUs for the first 220k iterations, then on one NVIDIA Quadro RTX~6000 GPU (24~GB) from 220k to 310k iterations, the
phase in which extremal points are memorized. All CIFAR-10 timings (Table~\ref{tab:cifar_cost}) are measured on the
RTX~4090 Laptop GPU. % TODO (author): GPU of the ablation with memorization from the first iteration.
%---------------------------------------------------------------------
\subsection{Synthetic distributions in the Euclidean plane}
%---------------------------------------------------------------------
\begin{itemize}
\item \emph{Potential $f_\theta$}: MLP with 3 hidden layers of width 256, ReLU activations, scalar output.
\item \emph{Inner problem}: best of the three candidates $\{x_1,\ \argmin_i[f(x_0^i)+c(x_0^i,x_1)],\ g_\phi(x_1)\}$, the
second one being taken over the current batch of $x_0$; then $K=5$ L-BFGS iterations (memory at most $K$), with the
direction replaced by $-\nabla$ if it is not a descent direction, Armijo line search (constant $10^{-4}$, initial
step $1$, at most 8 halvings), keeping the best point found. The inner problem is solved without gradients with
respect to $\theta$; the gradient of the loss is given by Danskin's theorem.
\item \emph{Loss}: the estimated loss $\hat{\mathcal L}_1$ of Proposition~\ref{prop:borne}, denoted $\ell_f$ in
Algorithm~1 of the main paper.
\item \emph{Predictor $g_\phi$} (optional candidate): MLP $3\times256$, SiLU activations, in residual form
$g_\phi(x)=x+\mathrm{MLP}(x)$, trained in parallel by regression, $\ell_g=|g_\phi(z_1)-z_0|^2$, on a second batch
$z_0\sim p_0$ with detached targets $z_1=z_0+2\nabla f_\theta(z_0)$; these pairs require no extremal-point computation.
Its ablation shows no measurable effect (Table~\ref{tab:supp_abl_candidats}), and it becomes unnecessary once extremal
points are memorized.
\item \emph{Optimization}: Adam, learning rate $10^{-3}$, cosine decay over 5{,}000 iterations, batches of 512; no
per-dataset tuning.
\end{itemize}
%---------------------------------------------------------------------
\subsection{Earth data on the sphere}
%---------------------------------------------------------------------
\begin{itemize}
\item \emph{Potential $f_\theta$}: MLP $3\times256$ with ReLU on the ambient coordinates in $\R^3$; Riemannian gradient
by tangent projection (Section~\ref{supp:formules}); cost $c=d^2/4$; generation $T(z)=\exp_z(2\nabla_\g f(z))$.
\item \emph{Predictor $g_\phi$}: MLP $3\times256$ with SiLU, $g_\phi(x)=(x+\mathrm{MLP}(x))/|x+\mathrm{MLP}(x)|$, chordal
loss $|g_\phi(z_1)-z_0|^2$.
\item \emph{Inner problem}: the same three candidates as in $\R^2$, then $K=10$ steps of Riemannian gradient descent,
$y\leftarrow\exp_y(-a\,\mathrm{grad})$, with Armijo line search along the exponential map (constant $10^{-4}$, at
most 8 halvings); the initial step of each iteration is the step accepted at the previous iteration multiplied by 2,
capped at 4.
\item \emph{Optimization}: Adam $10^{-3}$, cosine decay, 5{,}000 iterations, batches of 512.
\end{itemize}
%---------------------------------------------------------------------
\subsection{CIFAR-10}
%---------------------------------------------------------------------
\begin{itemize}
\item \emph{Potential $f_\theta$}: encoder half of the CIFAR-10 U-Net of \texttt{torchcfm} (\texttt{EncoderUNetModel},
160 channels, 4 residual blocks, channel multipliers $(1,2,2,2)$, attention at resolution 16, dropout 0), with ReLU
activations and a scalar output; 35.2M parameters, versus 35.7M for the U-Net of the baselines. Cost $c=|x-y|^2/4$
($k=2$) in $\R^{3072}$.
\item \emph{Inner problem}: best of the candidates $x_1$, best point of the batch and, from iteration 220k on, the
extremal point memorized for the same training image at its previous pass; then $K=5$ L-BFGS iterations, keeping the
best point found (L-BFGS reaches a better minimum than the fixed-point iteration, Table~\ref{tab:cifar_inner}). The
batch candidate is taken among the noise samples of the same GPU. The predictor $g_\phi$ (U-Net of \texttt{torchcfm},
35.75M parameters, trained as in $\R^2$) was among the candidates during the first 300k iterations; once extremal points
were memorized, it was no longer selected, and it was removed at 300k iterations.
\item \emph{Optimization}: training protocol of \texttt{torchcfm}: Adam, learning rate $2\cdot10^{-4}$, linear warm-up
over 5{,}000 iterations, then constant up to 300k iterations, then cosine decay (planned from $2\cdot10^{-4}$ at 300k
to $10^{-5}$ at 340k); gradient clipping at 1; exponential moving average of the weights with decay $0.9999$; global
batch of 128, split over 3 GPUs up to 220k iterations (about 43 images and 43 noise samples per GPU), then on one GPU,
with gradient accumulation over micro-batches when memory is short (same gradient). Protocol of the final run:
iterations 0--220k without memorization; 220k--300k with memorization, nothing else changed; 300k--310k without $g_\phi$
and with the learning-rate decay. We report the checkpoint with the best monitoring FID (10{,}000 images), reached at
310k iterations.
\item \emph{Sampling}: averaged weights, one step $x_1=x_0+2\nabla f_\theta(x_0)$, $x_0\sim\mathcal N(0,I)$.
\end{itemize}
%=====================================================================
\section{Datasets}\label{supp:data}
%=====================================================================
%---------------------------------------------------------------------
\subsection{Synthetic distributions in the Euclidean plane}
%---------------------------------------------------------------------
The source is $\mathcal N(0,I_2)$. Targets are sampled on the fly at each iteration:
\begin{itemize}
\item \emph{8 Gaussians}: centers $5(\cos\frac{2\pi j}8,\sin\frac{2\pi j}8)$, $j=0,\dots,7$, isotropic variance $0.1$;
\item \emph{Two Moons}: two interleaved half-circles, Gaussian noise with standard deviation $0.08$, scaled by $3$;
\item \emph{Checkerboard}: uniform distribution on 8 cells of a $4\times4$ checkerboard with unit cells, scaled by $2$;
\item \emph{Spirals}: two symmetric spirals with parameter $\sqrt{U}\cdot3\pi$, $U\sim\mathcal U[0,1]$, uniform noise of
width $0.5$, then scaling by $0.4$ and Gaussian noise with standard deviation $0.1$.
\end{itemize}
%---------------------------------------------------------------------
\subsection{Earth data on the sphere}
%---------------------------------------------------------------------
Files and coordinate parsing from the official RFM code; conversion
\[
\begin{aligned}
(\mathrm{lat},\mathrm{lon})\longmapsto\bigl(
&\cos\mathrm{lat}\cos\mathrm{lon}, \cos\mathrm{lat}\sin\mathrm{lon},\sin\mathrm{lat}\bigr).
\end{aligned}
\]
80/10/10 split with \texttt{random\_split} and seeds 0 to 4, checked to be identical to the split of their code.  
%---------------------------------------------------------------------
\subsection{CIFAR-10}
%---------------------------------------------------------------------
The 50{,}000 training images of CIFAR-10 ($32\times32\times3$, \ie, points of $\R^{3072}$), preprocessed as in
\texttt{torchcfm}, with random horizontal flips; the source is $\mathcal N(0,I)$.
%=====================================================================
\section{Baselines}\label{supp:baselines}
%=====================================================================
%---------------------------------------------------------------------
\subsection{Synthetic distributions in the Euclidean plane}
%---------------------------------------------------------------------
\begin{itemize}
\item \emph{Setting shared by Flow Matching, OT-CFM and 2-Rectified Flow}: field $v(x,t)$, MLP $3\times256$ with SiLU
on $(x,t)$; Adam $10^{-3}$ with cosine decay, 5{,}000 iterations, batches of 512; Euler sampling.
\item \emph{Flow Matching}~\cite{lipman2023flow}: \texttt{AffineProbPath} and \texttt{CondOTScheduler} from the \texttt{flow\_matching}
library~\cite{lipman2024guide}.
\item \emph{OT-CFM}~\cite{tong2024improving}: \texttt{ExactOptimalTransport}\allowbreak\texttt{ConditionalFlowMatcher} from \texttt{torchcfm}, $\sigma=0$.
\item \emph{2-Rectified Flow}~\cite{liu2023flow}: a first model trained for 5{,}000 iterations, then 20{,}000 pairs $(z_0,\hat x_1)$
generated by Euler integration and a second training run of 5{,}000 iterations on these pairs.
\item \emph{MeanFlow}~\cite{geng2025mean}: \texttt{MeanFlow} class from the official \texttt{py-meanflow} repository, network
$u(z,(t,t-r))$, MLP $3\times256$ with SiLU on $(z,t,t-r)$. Selected setting (repository defaults): time sampler
\texttt{v1}, $P_{\text{mean},t}=-0.6$, $P_{\text{std},t}=1.6$, $P_{\text{mean},r}=-4$, $P_{\text{std},r}=1.6$,
fraction $r\neq t$ of $0.75$, adaptive weighting $p=0.75$; Adam $6\cdot10^{-4}$, $\beta=(0.9,\,0.999)$, linear warm-up
over 5\% of the iterations then constant learning rate. Only adaptation: exponential moving average of the weights
with decay $0.999$ instead of $0.9999$, the latter assuming hundreds of thousands of iterations. Multi-step sampling:
$z_r=z_t-(t-r)\,u(z_t,(t,t-r))$, identical to their \texttt{sample} function for one step. The two other published
settings (CIFAR-10 and ImageNet) were tried and perform worse on our data.
\item \emph{W2GN}~\cite{korotin2021wasserstein}: official code (\texttt{DenseICNN}, rank 3, layers $[128,128,64]$, data normalization), procedure of
their 2D notebook: pretraining towards the identity (at most 10{,}000 iterations), then 30{,}000 iterations, batches
of 1{,}024, Adam $10^{-3}$, $\lambda=1$ for the cycle penalty.
\item \emph{OFM}~\cite{kornilov2024optimal}: official code (\texttt{ofmsrc} package, convex potential \texttt{ICNN2CPF}
of width 1{,}024 with 2 hidden layers), procedure of their 2D notebook: pretraining towards the identity
(200 iterations), OFM loss with inversion by L-BFGS (5 outer iterations, tolerance $10^{-2}$) and exact
inverse-Hessian-vector product, Adam $10^{-2}$, batches of 1{,}024, 30{,}000 iterations, independent coupling (the
authors' default setting); one-step generation $x_1=\nabla\Psi(x_0)$. As for W2GN, the target is whitened (zero mean,
identity covariance, with the \texttt{StandardNormalScaler} transform of their repository) and the samples are mapped
back to the data scale: without this normalization, training diverges on our data. Even so, with the authors' learning
rate, training diverges on 5 of the 20 runs (exploding loss, or a singular Hessian in their inversion). Every 1{,}000
iterations, we therefore evaluate the MMD on a validation set of 2{,}000 points drawn with a different seed from the
test set, keep the best checkpoint, and stop training as soon as the loss diverges. The selected iteration ranges from
3{,}000 to 29{,}000 depending on the run. This selection, analogous to the early stopping of the official RFM, can only
favor OFM.
\end{itemize}
%---------------------------------------------------------------------
\subsection{Earth data on the sphere}
%---------------------------------------------------------------------
\begin{itemize}
\item \emph{RFM, same budget}~\cite{chen2024flow}: field $v(x,t)$, MLP $3\times256$ with SiLU on $(x,t)\in\R^4$, projected onto the tangent
space; loss $\E|v(x_t,t)-\dot x_t|^2$ with the geodesic interpolation (Section~\ref{supp:formules}),
$t\sim\mathcal U[0,1]$, independent coupling; Adam $10^{-3}$, cosine decay, 5{,}000 iterations, batches of 512;
sampling by geodesic Euler $x\leftarrow\exp_x(h\,v(x,t))$.
\item \emph{Flow Matching in $\R^3$ + projection}~\cite{lipman2023flow} (ablation): same network,
\texttt{flow\_matching} library~\cite{lipman2024guide}, integration
in $\R^3$ followed by a projection onto the sphere at the end of integration.
\item \emph{RFM, official}~\cite{chen2024flow}: the authors' code and configuration for the Volcano dataset, seed 0 (MLP $2048\times12$,
batches of 8{,}192, exponential moving average of the weights, early stopping on the validation negative log-likelihood (NLL)). Training took
about 45~h on an RTX~4090 Laptop GPU; the best validation checkpoint is reached at epoch 6{,}463, i.e., about
808{,}000 iterations. Samples are generated with 1{,}000 steps and the exact NLL is computed with their evaluation
code.
\end{itemize}
%---------------------------------------------------------------------
\subsection{CIFAR-10}
%---------------------------------------------------------------------
\begin{itemize}
\item \emph{Flow Matching~\cite{lipman2023flow} and OT-CFM~\cite{tong2024improving}}: official \texttt{torchcfm}
checkpoints (U-Net with 35.7M parameters, 400k
iterations, batch of 128), sampled with the Euler scheme at 1, 2, 10 and 100 steps and with the adaptive
\texttt{dopri5} solver (tolerance $10^{-5}$).
\item \emph{MeanFlow}~\cite{geng2025mean}: official \texttt{py-meanflow} checkpoint (batch of 1{,}024, about 700k iterations), sampled with
1 and 2 steps.
\end{itemize}
With our evaluation code, we recover the published FID-50K of OT-CFM (3.69, versus 3.58 published) and of MeanFlow in
one step (2.80, versus 2.92 published; the repository reports about 2.80 for this checkpoint); no published value
corresponds to the Flow Matching checkpoint of \texttt{torchcfm}, for which we obtain 3.72 with dopri5. The other
methods of Table~\ref{tab:cifar_full}, 2-Rectified Flow~\cite{liu2023flow} (with and without distillation),
OTM~\cite{rout2022generative}, and UOTM~\cite{choi2023generative}  , were not
re-evaluated: their values are taken from the original papers, with training protocols and budgets that differ from
ours, and are given for reference only.
%=====================================================================
\section{Metrics}\label{supp:metrics}
%=====================================================================
%---------------------------------------------------------------------
\subsection{Synthetic distributions in the Euclidean plane}
%---------------------------------------------------------------------
We compare $N_a=2{,}000$ generated samples $\{a_i\}_{i\le N_a}$ with $N_b=2{,}000$ test samples $\{b_j\}_{j\le N_b}$,
independent of training.
\begin{itemize}
\item \emph{Maximum Mean Discrepancy (MMD)}: with the sum-of-Gaussians kernel
\[
\kappa(x,y)=\sum_{\sigma\in\{0.1,\,0.5,\,1,\,2,\,5\}}\exp\!\Big(-\frac{\|x-y\|^2}{2\sigma^2}\Big),
\]
we use the square root of the biased estimator (V-statistic)
\[
\begin{aligned}
\mathrm{MMD}=\Bigg[&\frac1{N_a^2}
 \sum_{i,i'=1}^{N_a}\kappa(a_i,a_{i'})\\
&+\frac1{N_b^2}
 \sum_{j,j'=1}^{N_b}\kappa(b_j,b_{j'})\\
&-\frac2{N_aN_b}\sum_{i=1}^{N_a}\sum_{j=1}^{N_b}
 \kappa(a_i,b_j)\Bigg]_+^{1/2}.
\end{aligned}
\]
where $[\cdot]_+=\max(\cdot,0)$; summing over several bandwidths makes the metric sensitive both to the fine structure
and to the global distribution of the mass.
\item \emph{Wasserstein-2 distance}: between the uniform empirical distributions, it is computed exactly by linear
programming (POT, \texttt{ot.emd2}):
\[
\begin{aligned}
W_2&=\left(\min_{P\in\Pi_{N_a,N_b}}
 \sum_{i=1}^{N_a}\sum_{j=1}^{N_b}
 P_{ij}\|a_i-b_j\|^2\right)^{1/2},\\
\Pi_{N_a,N_b}&=\Big\{P\ge0:\ 
 P\mathbf 1_{N_b}=\tfrac1{N_a}\mathbf 1_{N_a},\\
&\hspace{26mm}P^\top\mathbf 1_{N_a}
=\tfrac1{N_b}\mathbf 1_{N_b}\Big\}.
\end{aligned}
\]
\item \emph{Noise floor}: both metrics are also computed between two independent samples of 2{,}000 data points; this
value, nonzero because of the finite number of samples, is the best achievable value.
\item \emph{Hardware and timing}: one NVIDIA RTX~4090 Laptop GPU; sampling times are measured after CUDA
synchronization, for 2{,}000 samples.
\end{itemize}
%---------------------------------------------------------------------
\subsection{Earth data on the sphere}
%---------------------------------------------------------------------
\begin{itemize}
\item \emph{MMD}: sum of Gaussian kernels on the chordal distance, bandwidths $\{0.02,\,0.05,\,0.1,\,0.2,\,0.5\}$,
between 2{,}000 generated samples and the whole test set.
\item \emph{Kernel NLL}: von Mises--Fisher estimator built on 20{,}000 generated samples,
\[
\begin{aligned}
\hat q_\kappa(x)
&=\frac1N\sum_{i=1}^N
\frac{\kappa}{2\pi(1-e^{-2\kappa})}\\
&\quad\times e^{\kappa(\langle x,s_i\rangle-1)}.
\end{aligned}
\]
with concentration $\kappa$ chosen from:
\[
\begin{gathered}
\{10,30,100,300,1{,}000,3{,}000,10{,}000\}
\end{gathered}
\]
by maximum likelihood on the
validation set, then evaluated on the test set.
\item \emph{Exact NLL, RFM at the same budget}: backward integration of the log-density equation,
$\frac{\mathrm d}{\mathrm dt}\log p_t(x_t)=-\mathrm{div}_\g v_t(x_t)$, with 200 Euler steps, the divergence being
computed exactly in an orthonormal basis of the tangent space.
\item \emph{Exact NLL, our method}: change of variables $\log q(x_1)=\log p_0(x_0)-\log|\det\mathrm dT(x_0)|$, with
$\log p_0=-\log4\pi$, where $\mathrm dT(x_0)$ is the tangent map $T_{x_0}\mathbb S^2\to T_{T(x_0)}\mathbb S^2$ (ambient
Jacobian expressed in two tangent bases) and $x_0$ is obtained by solving the inner problem at $x_1$
(by Lemma~\ref{lem:part}, a minimizer $x_0$ at which $f$ is differentiable satisfies $T(x_0)=x_1$): best of 2{,}048 uniform candidates, then 60 steps of Riemannian gradient descent. A point
is considered inverted if $|T(x_0)-x_1|<10^{-3}$, and the NLL is averaged over these points only. Since $T$ is not
injective, the formula only accounts for the preimage found; it is not the density of $T_\#p_0$ where several
preimages exist.
\end{itemize}
%---------------------------------------------------------------------
\subsection{CIFAR-10}
%---------------------------------------------------------------------
\begin{itemize}
\item \emph{FID-50K}: \texttt{clean-fid} in \texttt{legacy\_tensorflow} mode, 50{,}000 generated images against the
statistics of the training set, as in the protocol of \texttt{torchcfm}; the same code is used for all methods.
\end{itemize}
\FloatBarrier
%=====================================================================
% Wide numerical tables use table* and max-width adjustbox in the two-column CVPR layout.
\section{Additional results}\label{supp:resultats}
%=====================================================================
%---------------------------------------------------------------------
\subsection{Synthetic distributions in the Euclidean plane}
\begin{figure*}[!tp]
\centering
\begin{subfigure}[t]{0.40\textwidth}
\centering
\includegraphics[width=\linewidth]{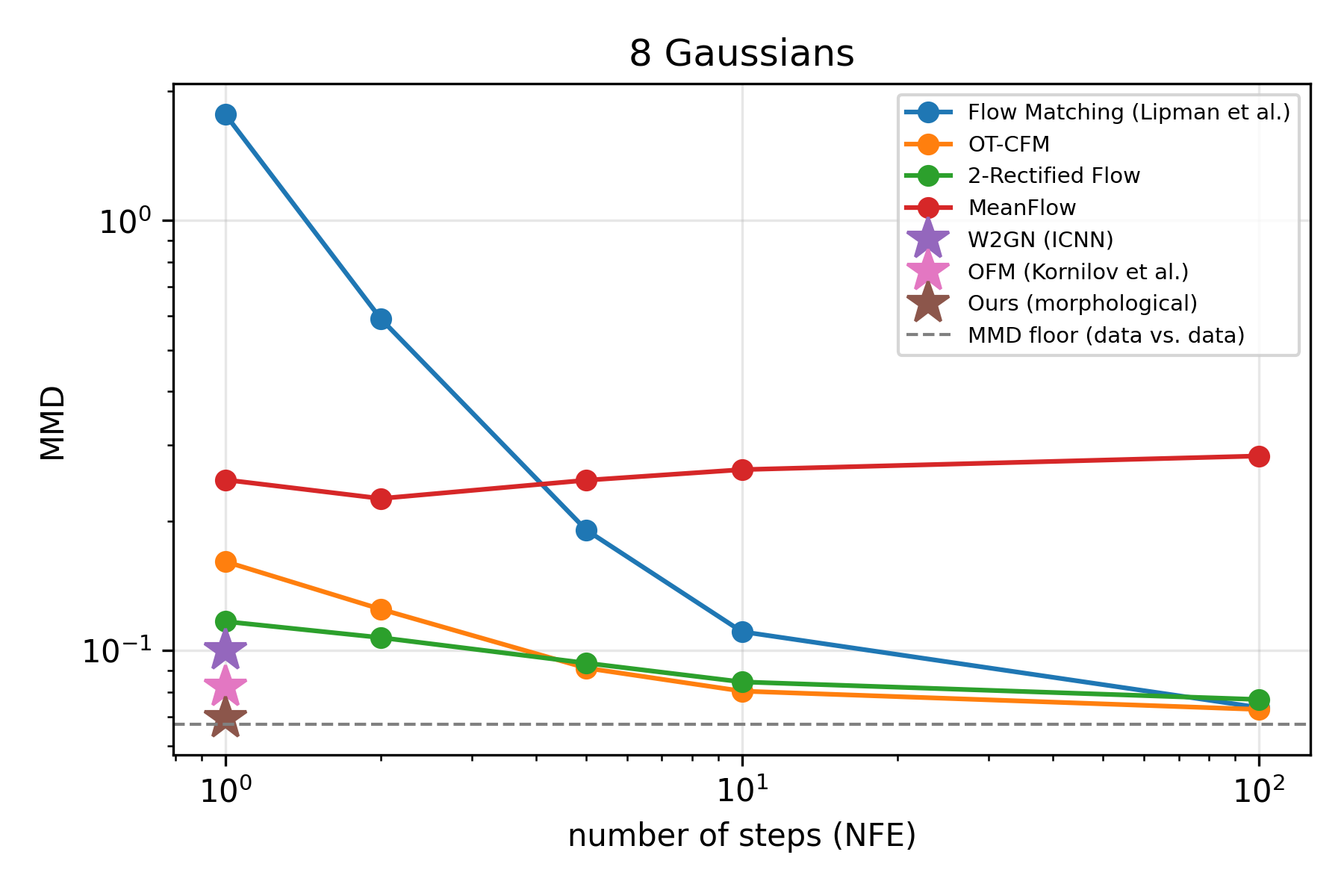}
\caption{8 Gaussians}
\end{subfigure}\hfill
\begin{subfigure}[t]{0.40\textwidth}
\centering
\includegraphics[width=\linewidth]{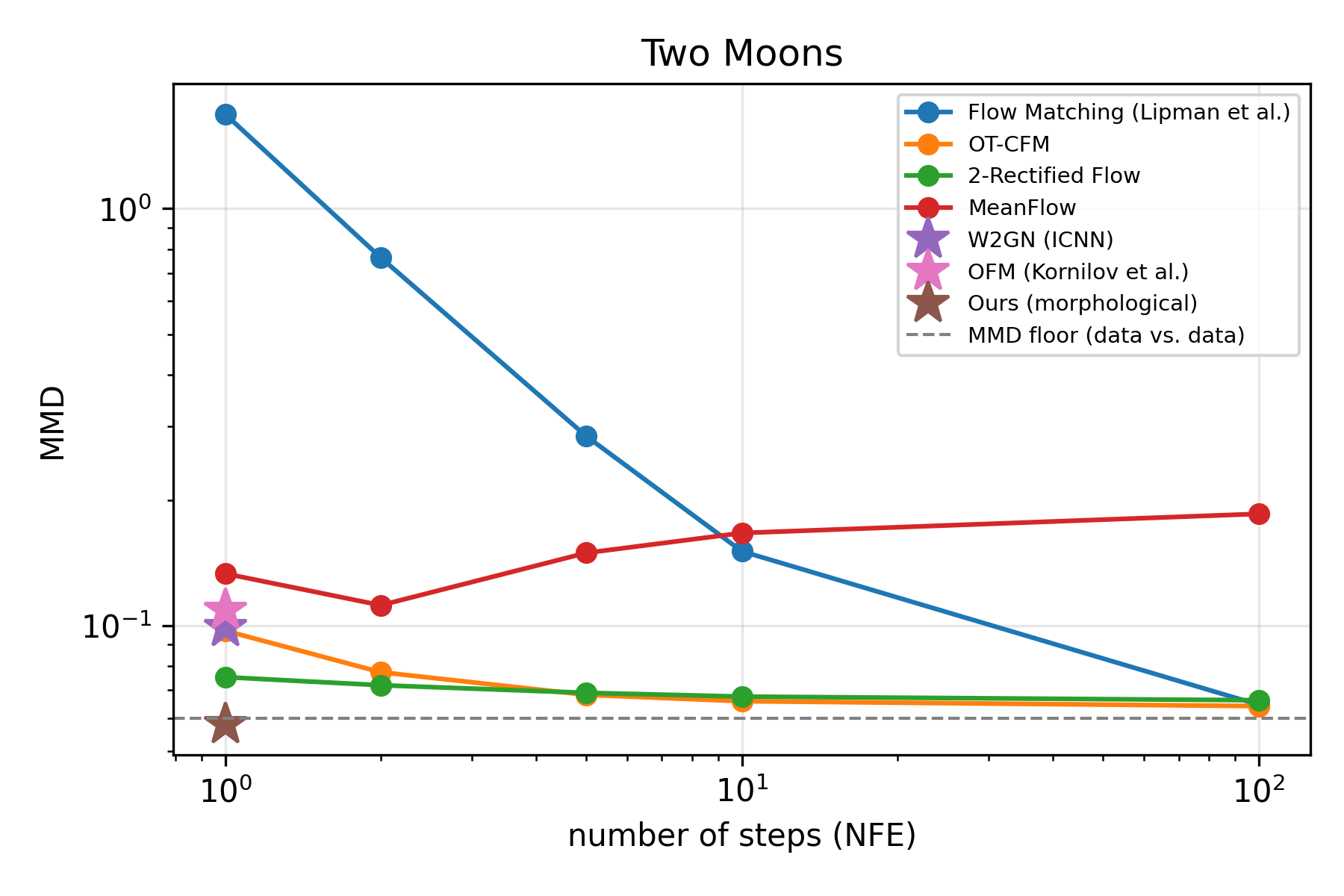}
\caption{Two Moons}
\end{subfigure}\\[4pt]
\begin{subfigure}[t]{0.40\textwidth}
\centering
\includegraphics[width=\linewidth]{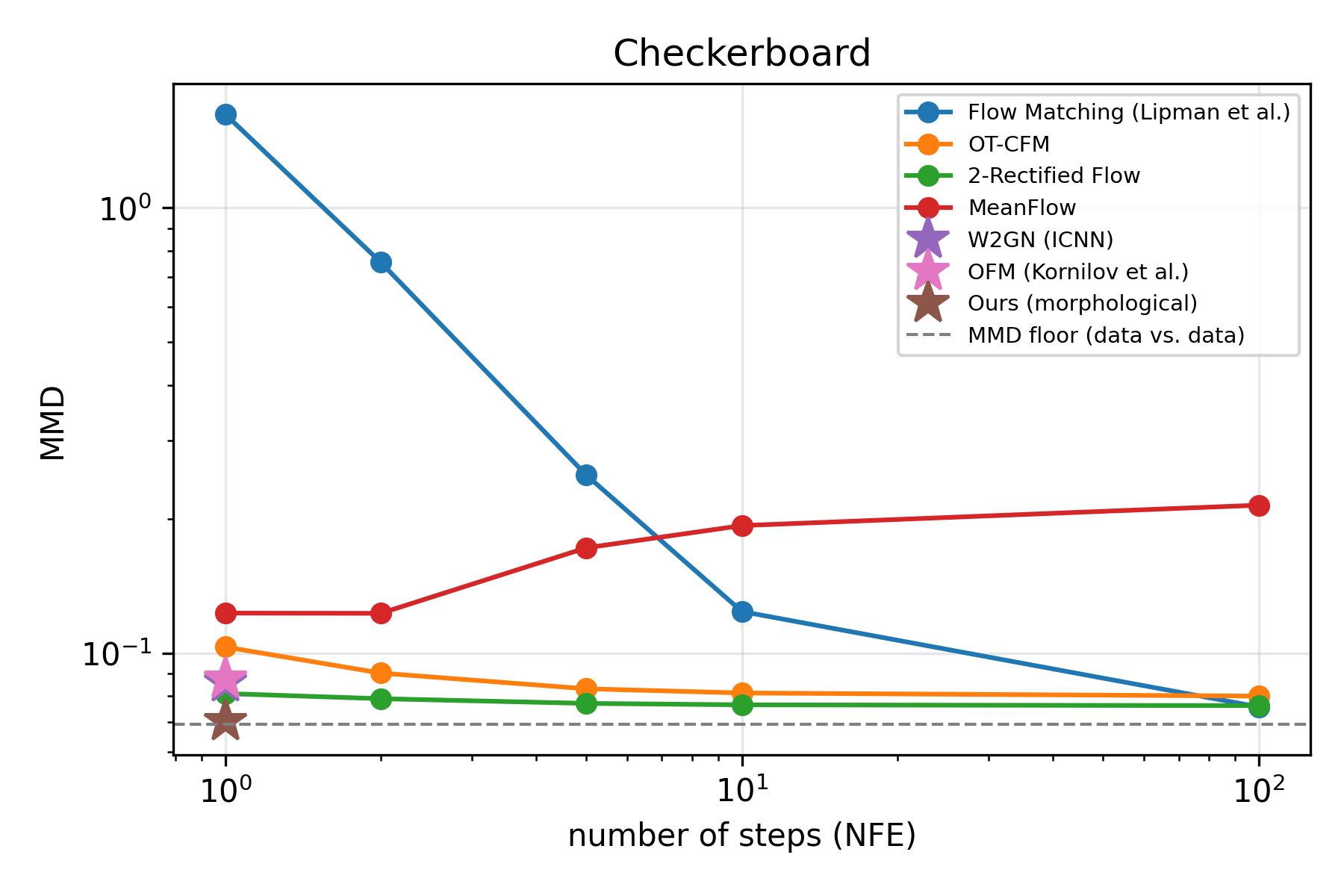}
\caption{Checkerboard}
\end{subfigure}\hfill
\begin{subfigure}[t]{0.40\textwidth}
\centering
\includegraphics[width=\linewidth]{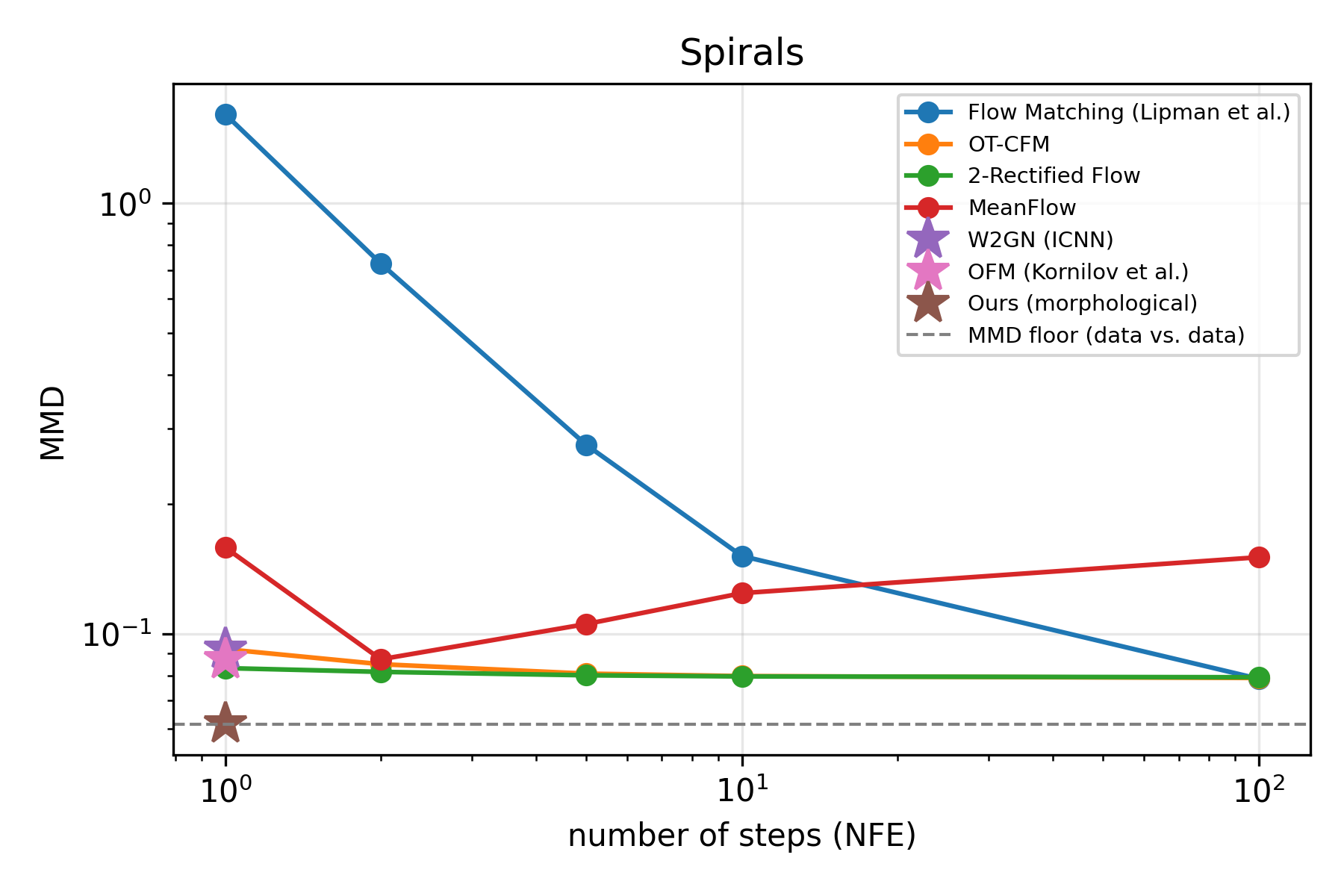}
\caption{Spirals}
\end{subfigure}
\caption{MMD as a function of the number of sampling steps (mean over 5 seeds, log-log scale). Stars denote one-step methods. The dashed line is the noise floor (two independent data samples).}
\label{fig:2d_nfe}
\end{figure*}

%---------------------------------------------------------------------
% Images: those of resultats_2d_avec_ofm/ (figures_2d.py), which include OFM.
This subsection complements the 2D benchmark of the main paper. Figure~\ref{fig:2d_samples} shows the samples generated
by each method at 1, 10 and 100 steps on the four datasets, and Figure~\ref{fig:2d_nfe} shows how the MMD evolves with
the number of steps. Table~\ref{tab:2d_full} reports the MMD for all numbers of steps: with a single step, our method
reaches the noise floor on all four datasets, within one standard deviation. Table~\ref{tab:2d_w2} reports the
Wasserstein-2 distance, and Table~\ref{tab:2d_cost} the training and sampling costs.
% ---- part 1: (a) 8 Gaussians, (b) Two Moons
\begin{figure*}[p]
\begin{minipage}[c][0.95\textheight][c]{\textwidth}
\centering
\begin{subfigure}[t]{0.49\textwidth}
\centering
\includegraphics[width=\linewidth,height=0.78\textheight,keepaspectratio]{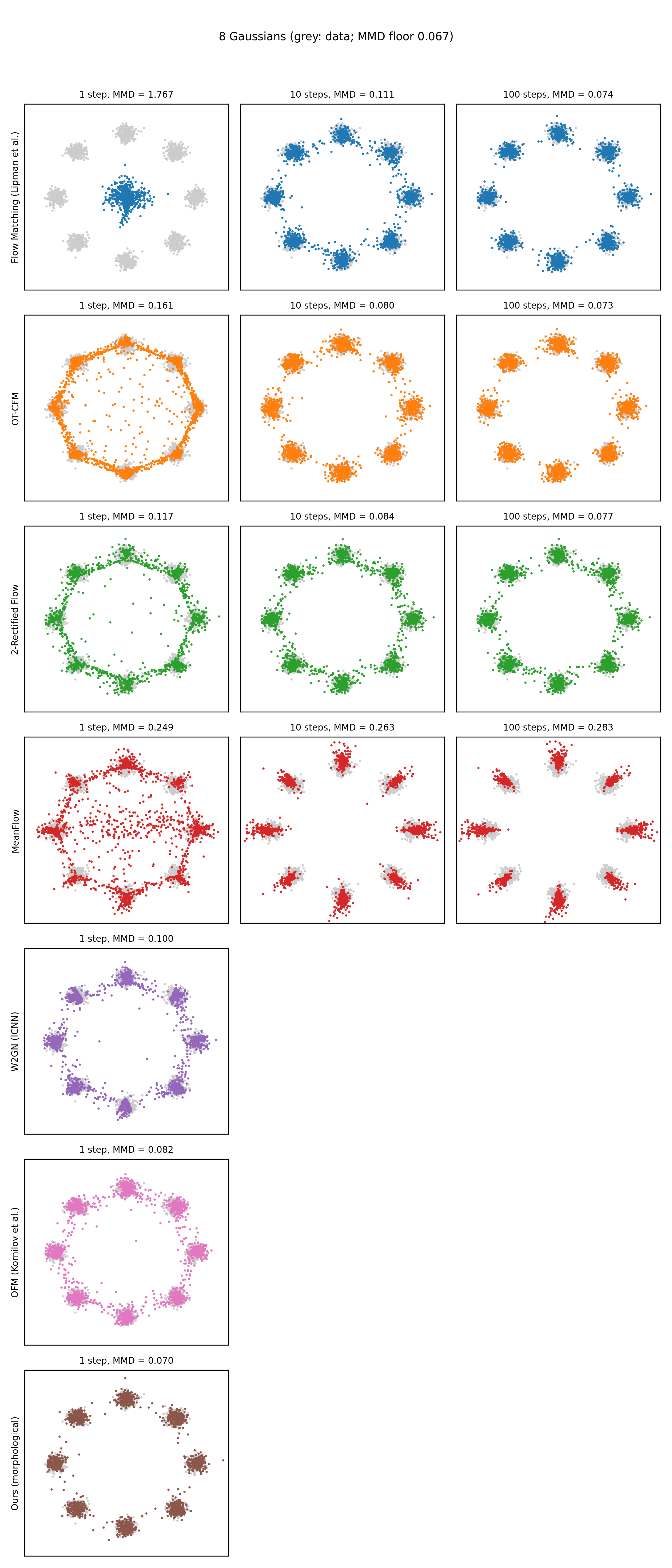}
\caption{8 Gaussians}
\end{subfigure}\hfill
\begin{subfigure}[t]{0.49\textwidth}
\centering
\includegraphics[width=\linewidth,height=0.78\textheight,keepaspectratio]{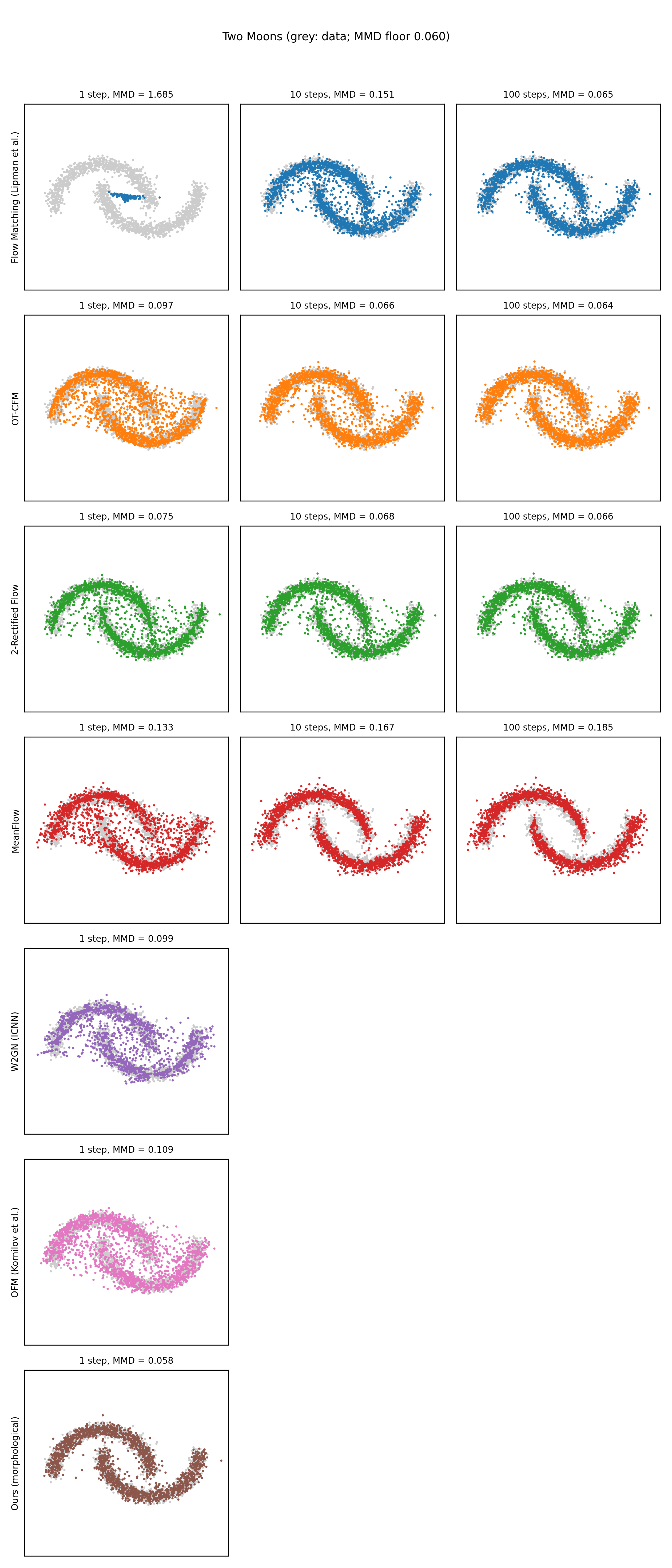}
\caption{Two Moons}
\end{subfigure}
\caption{Generated samples (seed 0) for each method at 1, 10 and 100 steps; grey: data. The reported MMD is the mean over 5 seeds. Methods whose transport map is continuous (W2GN, OFM, and OT-CFM and 2-Rectified Flow at one step) connect the modes with ``bridges'' of points, as predicted by the topological obstruction; our ReLU potential, whose gradient can jump, separates the modes in a single step.}
\label{fig:2d_samples}
\end{minipage}
\end{figure*}
\begin{figure*}[p]
\ContinuedFloat
\begin{minipage}[c][0.95\textheight][c]{\textwidth}
\centering
\begin{subfigure}[t]{0.49\textwidth}
\centering
\includegraphics[width=\linewidth,height=0.78\textheight,keepaspectratio]{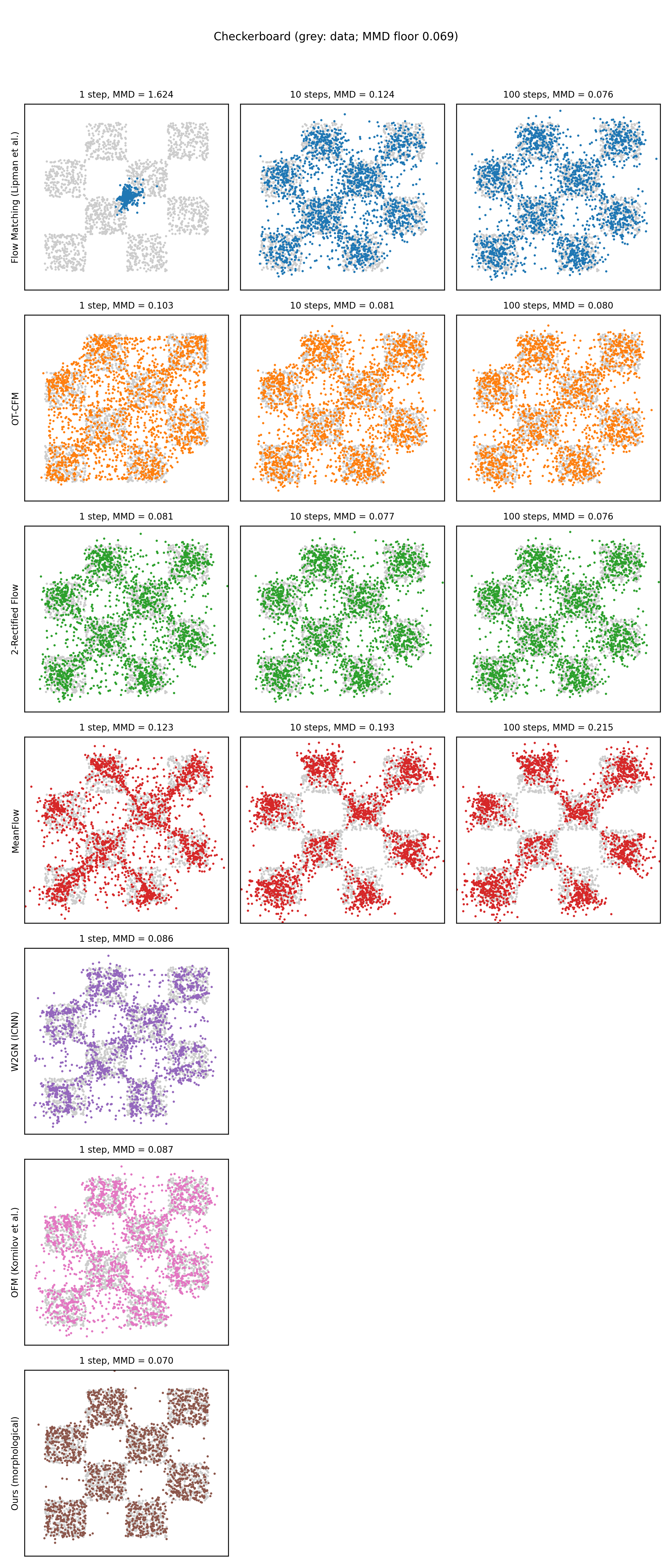}
\caption{Checkerboard}
\end{subfigure}\hfill
\begin{subfigure}[t]{0.49\textwidth}
\centering
\includegraphics[width=\linewidth,height=0.78\textheight,keepaspectratio]{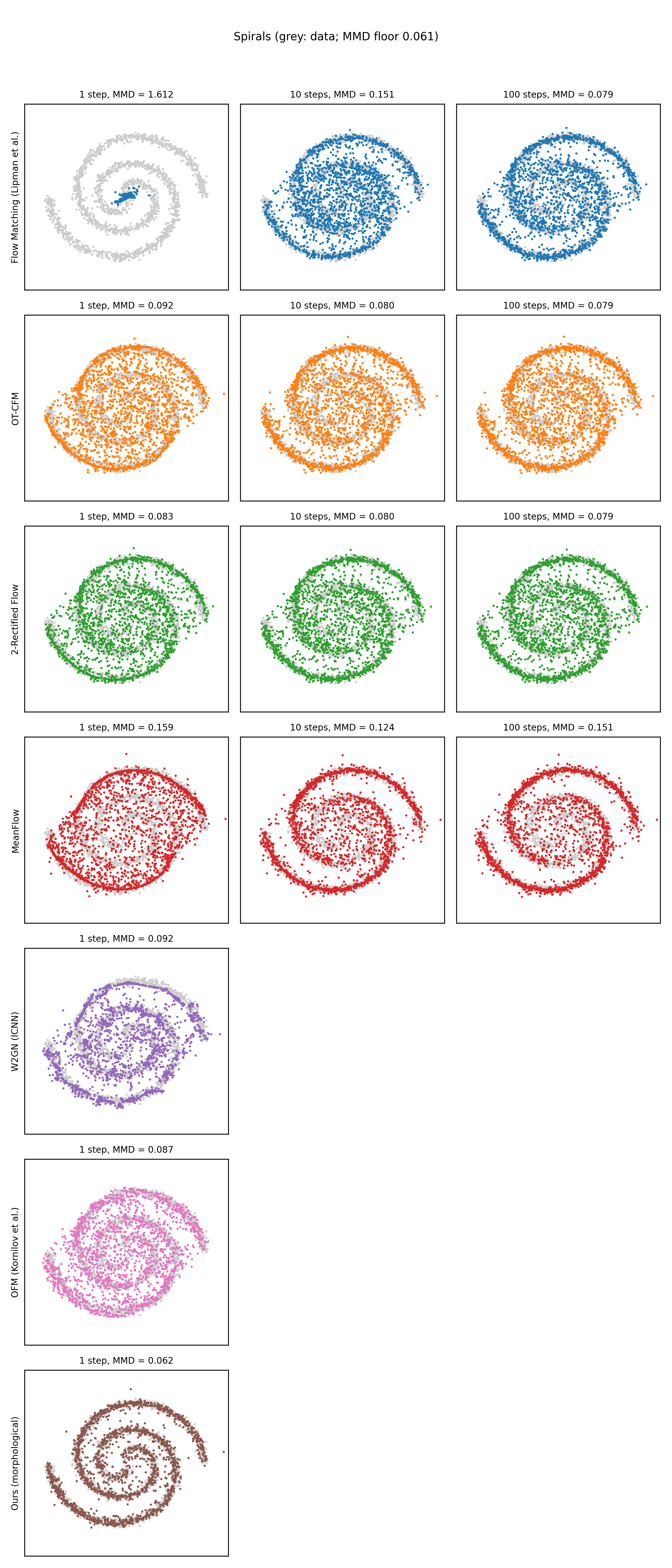}
\caption{Spirals}
\end{subfigure}
\caption[]{Generated samples (continued).}
\end{minipage}
\end{figure*}
\begin{table*}[!tp]
\centering
\footnotesize
\setlength{\tabcolsep}{3.5pt}
\caption{Full MMD results ($\downarrow$) on the 2D benchmark for all numbers of sampling steps (mean $\pm$ std over 5 seeds). The gray row is the MMD between two independent data samples and indicates the sampling-noise level; values slightly below it are within sampling noise. Best value per column highlighted; best one-step value underlined.}
\label{tab:2d_full}
\begin{adjustbox}{max width=\textwidth,center}
\begin{tabular}{l|c|cccc}
\toprule
\multirow{2}{*}{Method} & \multirow{2}{*}{NFE} & \multicolumn{4}{c}{MMD $\downarrow$} \\
\cmidrule(lr){3-6}
 & & 8 Gaussians & Two Moons & Checkerboard & Spirals \\
\midrule
\rowcolor{refgray}
Data (independent sample) & -- & 0.0672 & 0.0599 & 0.0691 & 0.0615 \\
\midrule
Flow Matching~\cite{lipman2023flow} & 1 & $1.7671 \pm 0.0093$ & $1.6853 \pm 0.0257$ & $1.6237 \pm 0.0160$ & $1.6125 \pm 0.0326$ \\
 & 2 & $0.5908 \pm 0.0044$ & $0.7631 \pm 0.0108$ & $0.7545 \pm 0.0048$ & $0.7248 \pm 0.0068$ \\
 & 5 & $0.1905 \pm 0.0068$ & $0.2845 \pm 0.0041$ & $0.2510 \pm 0.0070$ & $0.2746 \pm 0.0063$ \\
 & 10 & $0.1106 \pm 0.0085$ & $0.1507 \pm 0.0068$ & $0.1240 \pm 0.0061$ & $0.1515 \pm 0.0073$ \\
 & 100 & $0.0737 \pm 0.0106$ & $0.0646 \pm 0.0053$ & $0.0757 \pm 0.0049$ & $0.0785 \pm 0.0057$ \\
\midrule
OT-CFM~\cite{tong2024improving} & 1 & $0.1610 \pm 0.0073$ & $0.0972 \pm 0.0045$ & $0.1032 \pm 0.0035$ & $0.0921 \pm 0.0043$ \\
 & 2 & $0.1247 \pm 0.0074$ & $0.0773 \pm 0.0049$ & $0.0902 \pm 0.0034$ & $0.0849 \pm 0.0044$ \\
 & 5 & $0.0910 \pm 0.0078$ & $0.0682 \pm 0.0050$ & $0.0831 \pm 0.0035$ & $0.0807 \pm 0.0044$ \\
 & 10 & $0.0804 \pm 0.0083$ & $0.0658 \pm 0.0050$ & $0.0814 \pm 0.0037$ & $0.0797 \pm 0.0043$ \\
 & 100 & $0.0729 \pm 0.0087$ & $0.0641 \pm 0.0050$ & $0.0801 \pm 0.0039$ & $0.0789 \pm 0.0043$ \\
\midrule
2-Rectified Flow~\cite{liu2023flow} & 1 & $0.1168 \pm 0.0048$ & $0.0752 \pm 0.0044$ & $0.0811 \pm 0.0050$ & $0.0832 \pm 0.0053$ \\
 & 2 & $0.1071 \pm 0.0056$ & $0.0719 \pm 0.0046$ & $0.0790 \pm 0.0049$ & $0.0815 \pm 0.0054$ \\
 & 5 & $0.0935 \pm 0.0076$ & $0.0690 \pm 0.0049$ & $0.0771 \pm 0.0048$ & $0.0800 \pm 0.0056$ \\
 & 10 & $0.0845 \pm 0.0088$ & $0.0675 \pm 0.0050$ & $0.0765 \pm 0.0048$ & $0.0795 \pm 0.0056$ \\
 & 100 & $0.0769 \pm 0.0097$ & $0.0662 \pm 0.0051$ & $0.0762 \pm 0.0049$ & $0.0792 \pm 0.0057$ \\
\midrule
MeanFlow~\cite{geng2025mean} & 1 & $0.2494 \pm 0.0091$ & $0.1332 \pm 0.0062$ & $0.1230 \pm 0.0052$ & $0.1586 \pm 0.0126$ \\
 & 2 & $0.2256 \pm 0.0104$ & $0.1119 \pm 0.0137$ & $0.1228 \pm 0.0076$ & $0.0872 \pm 0.0065$ \\
 & 5 & $0.2489 \pm 0.0078$ & $0.1494 \pm 0.0142$ & $0.1724 \pm 0.0131$ & $0.1053 \pm 0.0106$ \\
 & 10 & $0.2634 \pm 0.0060$ & $0.1667 \pm 0.0145$ & $0.1934 \pm 0.0143$ & $0.1242 \pm 0.0121$ \\
 & 100 & $0.2833 \pm 0.0052$ & $0.1853 \pm 0.0146$ & $0.2149 \pm 0.0150$ & $0.1505 \pm 0.0124$ \\
\midrule
W2GN~\cite{korotin2021wasserstein} & 1 & $0.0999 \pm 0.0132$ & $0.0992 \pm 0.0109$ & $0.0858 \pm 0.0067$ & $0.0923 \pm 0.0096$ \\
\midrule
OFM~\cite{kornilov2024optimal} & 1 & $0.0821 \pm 0.0142$ & $0.1088 \pm 0.0136$ & $0.0875 \pm 0.0054$ & $0.0873 \pm 0.0060$ \\
\midrule
\textbf{MorphoFM (ours)} & 1 & \best{\underline{0.0695 \pm 0.0112}} & \best{\underline{0.0579 \pm 0.0064}} & \best{\underline{0.0701 \pm 0.0052}} & \best{\underline{0.0618 \pm 0.0054}} \\
\bottomrule
\end{tabular}
\end{adjustbox}
\end{table*}
\begin{table}[!htbp]
\centering
\small
\setlength{\tabcolsep}{3pt}
\caption{Computational cost on the 2D benchmark (one NVIDIA RTX~4090 Laptop GPU, mean over the 4 datasets and
5 seeds). Sampling time is for 2{,}000 samples. The last block gives, for each dataset (8G: 8 Gaussians, TM: Two
Moons, CB: Checkerboard, Sp: Spirals), the smallest number of steps among $\{1,2,5,10,100\}$ at which the mean MMD is
within 10\% of the noise floor (Table~\ref{tab:2d_full}); $>$100: not reached; --: one-step method that does not
reach it.}
\label{tab:2d_cost}
\resizebox{\linewidth}{!}{%
\begin{tabular}{l|cc|cc|cccc}
\toprule
\multirow{2}{*}{Method} & \multicolumn{2}{c|}{Training} & \multicolumn{2}{c|}{Sampling (ms)} & \multicolumn{4}{c}{NFE to noise floor $\downarrow$} \\
\cmidrule(lr){2-3}\cmidrule(lr){4-5}\cmidrule(lr){6-9}
 & Iterations & Time (s) & 1 NFE & 100 NFE & 8G & TM & CB & Sp \\
\midrule
Flow Matching~\cite{lipman2023flow} & 5{,}000 & 14 & 2.2 & 110.0 & 100 & 100 & 100 & $>$100 \\
OT-CFM~\cite{tong2024improving} & 5{,}000 & 103 & 1.3 & 108.8 & 100 & 10 & $>$100 & $>$100 \\
2-Rectified Flow~\cite{liu2023flow} & $2\times5{,}000$ (reflow) & 23 & 1.3 & 107.7 & $>$100 & $>$100 & $>$100 & $>$100 \\
MeanFlow~\cite{geng2025mean} & 5{,}000 & 11 & 0.2 & 11.5 & $>$100 & $>$100 & $>$100 & $>$100 \\
W2GN~\cite{korotin2021wasserstein} & 30{,}000 (+ pretraining) & 252 & 1.1 & -- & -- & -- & -- & -- \\
OFM~\cite{kornilov2024optimal} & $\le$30{,}000 (+ pretraining) & 1{,}096 & 2.5 & -- & -- & -- & -- & -- \\
\midrule
\textbf{MorphoFM (ours)} & 5{,}000 & 194 & 2.4 & -- & \best{1} & \best{1} & \best{1} & \best{1} \\
\bottomrule
\end{tabular}}
\end{table}
\paragraph{Computational cost (Table~\ref{tab:2d_cost}).}
Training our method takes longer than training flow matching methods (about 190~s versus 10 to 110~s), owing to the
solution of the inner problem at each iteration; it nevertheless remains faster than W2GN and more than five times
faster than OFM. In return, sampling requires only a single evaluation of the gradient of the potential, \ie, about
2~ms for 2{,}000 samples, versus about 110~ms for ODE-based methods with 100 steps, and our method is the only one that
reaches the noise floor with a single network evaluation on all four datasets.
\begin{table*}[!tp]
\centering
\footnotesize
\setlength{\tabcolsep}{3.5pt}
\caption{Wasserstein-2 distance ($\downarrow$) between 2{,}000 generated samples and 2{,}000 held-out data samples (mean $\pm$ std over 5 seeds), computed exactly with POT. The gray row is the distance between two independent data samples and indicates the sampling-noise level; values slightly below it are within sampling noise. Best value per column highlighted; best one-step value underlined.}
\label{tab:2d_w2}
\begin{adjustbox}{max width=\textwidth,center}
\begin{tabular}{l|c|cccc}
\toprule
\multirow{2}{*}{Method} & \multirow{2}{*}{NFE} & \multicolumn{4}{c}{Wasserstein-2 $\downarrow$} \\
\cmidrule(lr){3-6}
 & & 8 Gaussians & Two Moons & Checkerboard & Spirals \\
\midrule
\rowcolor{refgray}
Data (independent sample) & -- & 0.735 & 0.345 & 0.269 & 0.239 \\
\midrule
Flow Matching~\cite{lipman2023flow} & 1 & $4.690 \pm 0.011$ & $2.905 \pm 0.012$ & $3.052 \pm 0.010$ & $2.587 \pm 0.022$ \\
 & 10 & $0.713 \pm 0.106$ & $0.525 \pm 0.033$ & $0.450 \pm 0.018$ & $0.432 \pm 0.017$ \\
 & 100 & $0.747 \pm 0.115$ & \best{0.225 \pm 0.038} & $0.303 \pm 0.023$ & $0.260 \pm 0.020$ \\
\midrule
OT-CFM~\cite{tong2024improving} & 1 & $0.766 \pm 0.072$ & $0.332 \pm 0.019$ & $0.379 \pm 0.014$ & $0.300 \pm 0.012$ \\
 & 10 & $0.700 \pm 0.096$ & $0.239 \pm 0.031$ & $0.321 \pm 0.012$ & $0.274 \pm 0.015$ \\
 & 100 & $0.724 \pm 0.101$ & $0.234 \pm 0.033$ & $0.317 \pm 0.012$ & $0.273 \pm 0.016$ \\
\midrule
2-Rectified Flow~\cite{liu2023flow} & 1 & $0.681 \pm 0.088$ & $0.271 \pm 0.026$ & $0.324 \pm 0.019$ & $0.276 \pm 0.014$ \\
 & 10 & $0.681 \pm 0.109$ & $0.243 \pm 0.032$ & $0.308 \pm 0.021$ & $0.265 \pm 0.017$ \\
 & 100 & $0.703 \pm 0.116$ & $0.235 \pm 0.033$ & $0.306 \pm 0.022$ & $0.264 \pm 0.018$ \\
\midrule
MeanFlow~\cite{geng2025mean} & 1 & $1.343 \pm 0.099$ & $0.425 \pm 0.039$ & $0.397 \pm 0.020$ & $0.443 \pm 0.032$ \\
 & 10 & $1.313 \pm 0.036$ & $0.569 \pm 0.075$ & $0.628 \pm 0.035$ & $0.429 \pm 0.041$ \\
 & 100 & $1.351 \pm 0.036$ & $0.644 \pm 0.075$ & $0.700 \pm 0.037$ & $0.506 \pm 0.037$ \\
\midrule
W2GN~\cite{korotin2021wasserstein} & 1 & $0.687 \pm 0.122$ & $0.348 \pm 0.058$ & $0.327 \pm 0.026$ & $0.301 \pm 0.033$ \\
\midrule
OFM~\cite{kornilov2024optimal} & 1 & \best{\underline{0.654 \pm 0.146}} & $0.382 \pm 0.056$ & $0.338 \pm 0.028$ & $0.299 \pm 0.022$ \\
\midrule
\textbf{MorphoFM (ours)} & 1 & $0.732 \pm 0.114$ & \best{\underline{0.225 \pm 0.049}} & \best{\underline{0.282 \pm 0.022}} & \best{\underline{0.228 \pm 0.032}} \\
\bottomrule
\end{tabular}
\end{adjustbox}
\end{table*}
\paragraph{Wasserstein-2 distance (Table~\ref{tab:2d_w2}).}
With a single step, our method reaches the noise floor, within one standard deviation, or falls below it on all four
datasets. On Two Moons, Checkerboard, and Spirals, it is the best one-step method, $0.04$ to $0.05$ below the best
one-step competitor (2-Rectified Flow), and ODE-based methods do not do better even with 100 steps. On 8 Gaussians, all
reasonable methods lie within one standard deviation of the noise floor ($0.735$), so this metric does not discriminate
between methods on this dataset.
%---------------------------------------------------------------------
\subsection{Earth data on the sphere}
\begin{figure*}[!tp]
\centering
\includegraphics[width=\textwidth,height=0.6\textheight,keepaspectratio]{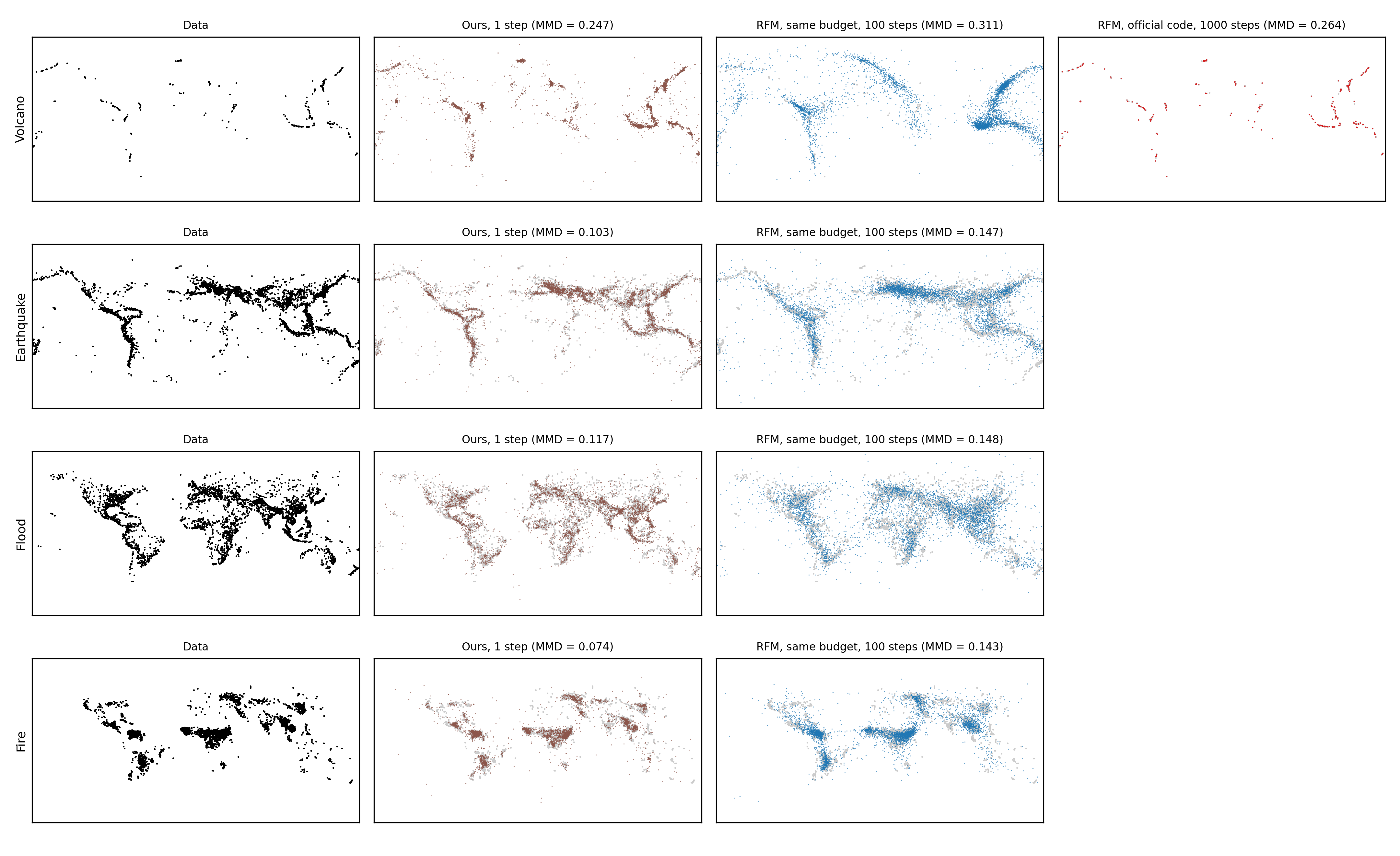}
\caption{Generated samples on the four Earth datasets (seed 0, longitude--latitude projection). From left to right:
data; our method with one step; RFM at the same budget with 100 steps; official RFM with 1{,}000 steps (Volcano
only). Grey: data; the reported MMD is that of seed 0. Our method recovers the fine structure of the data (volcanic
arcs, seismic ridges, continental outlines) in a single step, whereas RFM at the same budget spreads the mass and
creates structures absent from the data.}
\label{fig:sphere}
\end{figure*}

%---------------------------------------------------------------------
This subsection complements the sphere experiments of the main paper. Table~\ref{tab:sphere_cost} reports the training
and sampling costs, Table~\ref{tab:sphere_ablation} the ablation on the role of the geometry, and
Figure~\ref{fig:sphere} the samples generated on the four Earth datasets (the main paper only shows the Volcano
dataset). Table~\ref{tab:volcano_budget} studies the training budget of RFM, and Table~\ref{tab:sphere_nll}
reports the exact negative log-likelihoods.
\begin{table}[!htbp]
\centering
\footnotesize
\caption{Computational cost on the Earth datasets (one NVIDIA RTX~4090 Laptop GPU). Training time is the mean over datasets and seeds; sampling time is for 20{,}000 samples. The official RFM figure is for the Volcano run.}
\label{tab:sphere_cost}
\begin{adjustbox}{max width=\columnwidth,center}
\begin{tabular}{l|c|c|c|c|c}
\toprule
Method & Network & Iter. & Batch & Training & Sampling (ms) \\
\midrule
RFM, same budget & $3\times256$ & 5k & 512 & 6 s & 58.3 (100 NFE) \\
RFM, official & $2048\times12$ & $\approx$808k & 8{,}192 & $\approx$45 h & 139{,}906 (1000 NFE) \\
\midrule
\textbf{MorphoFM (ours)} & $3\times256$ & 5k & 512 & 164 s & 1.2 (1 NFE) \\
\bottomrule
\end{tabular}
\end{adjustbox}
\end{table}
\begin{table*}[!tp]
\centering
\footnotesize
\setlength{\tabcolsep}{3pt}
\caption{Ablation on $\mathbb{S}^2$: ignoring the geometry. Euclidean flow matching~\cite{lipman2023flow} in $\mathbb{R}^3$ followed by a projection onto the sphere, compared with RFM at the same budget (MMD $\downarrow$, mean $\pm$ std over 5 seeds). The gray row replaces generated samples by training samples (main paper). Best generative value per column highlighted.}
\label{tab:sphere_ablation}
\begin{adjustbox}{max width=\textwidth,center}
\begin{tabular}{l|c|c|cccc}
\toprule
\multirow{2}{*}{Method} & \multirow{2}{*}{Geometry} & \multirow{2}{*}{NFE} & \multicolumn{4}{c}{MMD $\downarrow$} \\
\cmidrule(lr){4-7}
 & & & Volcano & Earthquake & Flood & Fire \\
\midrule
\rowcolor{refgray}
Reference (training data) & -- & -- & $0.246 \pm 0.023$ & $0.102 \pm 0.007$ & $0.117 \pm 0.004$ & $0.073 \pm 0.009$ \\
\midrule
Flow Matching~\cite{lipman2023flow} & $\mathbb{R}^3$ + proj. & 1 & $1.108 \pm 0.038$ & $1.169 \pm 0.042$ & $1.166 \pm 0.014$ & $1.175 \pm 0.042$ \\
 & & 10 & $0.369 \pm 0.030$ & $0.203 \pm 0.018$ & $0.208 \pm 0.009$ & $0.281 \pm 0.020$ \\
 & & 100 & $0.346 \pm 0.023$ & $0.195 \pm 0.016$ & $0.191 \pm 0.015$ & $0.248 \pm 0.016$ \\
\midrule
RFM~\cite{chen2024flow}, same budget & $\mathbb{S}^2$ & 1 & $0.594 \pm 0.042$ & $0.493 \pm 0.008$ & $0.421 \pm 0.014$ & $0.645 \pm 0.011$ \\
 & & 10 & $0.328 \pm 0.026$ & $0.155 \pm 0.004$ & $0.151 \pm 0.004$ & $0.181 \pm 0.009$ \\
 & & 100 & $0.318 \pm 0.022$ & $0.147 \pm 0.004$ & $0.148 \pm 0.003$ & $0.145 \pm 0.005$ \\
\midrule
\textbf{MorphoFM (ours)} & $\mathbb{S}^2$ & 1 & \best{0.252 \pm 0.019} & \best{0.107 \pm 0.008} & \best{0.116 \pm 0.006} & \best{0.073 \pm 0.006} \\
\bottomrule
\end{tabular}
\end{adjustbox}
\end{table*}
\begin{table}[!htbp]
\centering
\small
\setlength{\tabcolsep}{3pt}
\caption{Effect of the training budget on the Volcano dataset (seed 0, identical test split). The official RFM is trained with the authors' code and configuration~\cite{chen2024flow} ($2048\times12$ MLP, batch size 8{,}192) and evaluated at its best validation checkpoint ($\approx$808k iterations); its exact NLL is computed with the authors' evaluation code. For our method, the exact NLL is computed on the inverted test points (fraction in parentheses). Best generative value per column highlighted.}
\label{tab:volcano_budget}
\resizebox{\linewidth}{!}{%
\begin{tabular}{l|ccc|cc|c}
\toprule
\multirow{2}{*}{Method} & \multicolumn{3}{c|}{Setup} & \multicolumn{2}{c|}{Sample quality} & Likelihood \\
\cmidrule(lr){2-4}\cmidrule(lr){5-6}\cmidrule(lr){7-7}
 & Network & Iterations & NFE & MMD $\downarrow$ & Kernel NLL $\downarrow$ & Exact NLL $\downarrow$ \\
\midrule
\rowcolor{refgray}
Reference (training data) & -- & -- & -- & $0.244$ & $-1.92$ & -- \\
\midrule
RFM, same budget & MLP $3\times256$ & 5k & 1 & $0.556$ & $2.56$ & -- \\
 & & & 10 & $0.318$ & $0.48$ & -- \\
 & & & 100 & $0.311$ & $0.45$ & $0.26$ \\
\midrule
RFM, official & MLP $2048\times12$ & $\approx$808k & 1 & $0.530$ & $2.54$ & -- \\
 & & & 10 & $0.281$ & $-0.94$ & -- \\
 & & & 100 & $0.261$ & \best{-2.00} & -- \\
 & & & 1000 & $0.264$ & $-1.92$ & \best{-6.68} \\
\midrule
\textbf{MorphoFM (ours)} & MLP $3\times256$ & 5k & 1 & \best{0.247} & $-1.46$ & $0.99$ (40\%) \\
\bottomrule
\end{tabular}}
\end{table}
\begin{table}[!htbp]
\centering
\small
\setlength{\tabcolsep}{4pt}
\caption{Exact test negative log-likelihood ($\downarrow$) on the Earth datasets. The upper block reports values from~\cite{chen2024flow} (5 runs). For our method, the one-step map is not injective (it is the gradient of a non-convex ReLU potential), so the exact density is only defined on the image of the map; we report the NLL on the fraction of test points that are successfully inverted (in parentheses). RFM (same budget) is integrated with 200 Euler steps. Best value per column highlighted.}
\label{tab:sphere_nll}
\resizebox{\linewidth}{!}{%
\begin{tabular}{l|cccc}
\toprule
\multirow{2}{*}{Method} & \multicolumn{4}{c}{Exact NLL $\downarrow$} \\
\cmidrule(lr){2-5}
 & Volcano & Earthquake & Flood & Fire \\
\midrule
\multicolumn{5}{l}{\textit{Reported in~\cite{chen2024flow}}} \\
Riemannian CNF~\cite{mathieu2020riemannian} & $-6.05 \pm 0.61$ & $0.14 \pm 0.23$ & $1.11 \pm 0.19$ & $-0.80 \pm 0.54$ \\
Moser Flow~\cite{rozen2021moser} & $-4.21 \pm 0.17$ & $-0.16 \pm 0.06$ & $0.57 \pm 0.10$ & $-1.28 \pm 0.05$ \\
CNF Matching~\cite{benhamu2022matching} & $-2.38 \pm 0.17$ & $-0.38 \pm 0.01$ & \best{0.25 \pm 0.02} & $-1.40 \pm 0.02$ \\
Riemannian Score-Based~\cite{debortoli2022riemannian} & $-4.92 \pm 0.25$ & $-0.19 \pm 0.07$ & $0.48 \pm 0.17$ & $-1.33 \pm 0.06$ \\
Riemannian Diffusion~\cite{huang2022riemannian} & $-6.61 \pm 0.96$ & \best{-0.40 \pm 0.05} & $0.43 \pm 0.07$ & $-1.38 \pm 0.05$ \\
RFM~\cite{chen2024flow} & \best{-7.93 \pm 1.67} & $-0.28 \pm 0.08$ & $0.42 \pm 0.05$ & \best{-1.86 \pm 0.11} \\
\midrule
\multicolumn{5}{l}{\textit{Our runs}} \\
RFM, same budget & $0.12 \pm 0.16$ & $0.86 \pm 0.06$ & $1.37 \pm 0.02$ & $0.01 \pm 0.05$ \\
\textbf{MorphoFM (ours)} & $0.66 \pm 0.42$ {\scriptsize(35\%)} & $1.12 \pm 0.14$ {\scriptsize(42\%)} & $1.35 \pm 0.16$ {\scriptsize(47\%)} & $0.48 \pm 0.13$ {\scriptsize(39\%)} \\
\bottomrule
\end{tabular}}
\end{table}
\paragraph{Training budget (Table~\ref{tab:volcano_budget}).}
To verify that our conclusions do not stem from an undertrained competitor, we trained the official RFM with its full
protocol on Volcano. Its exact NLL ($-6.68$) lies within the published range ($-7.93\pm1.67$), which confirms that the
reproduction is correct. With a training budget more than a thousand times larger (about 160 times more iterations and
16 times larger batches, \ie, about 2{,}600 times more points seen, with a much larger network) and 1{,}000 integration
steps, RFM achieves a better likelihood than our method, but a worse MMD ($0.264$ vs.\ $0.247$, for a reference of
$0.244$). Moreover, the sample quality of RFM no longer improves with training: its MMD is the same at about 322{,}000
and 808{,}000 iterations, whereas its exact NLL keeps decreasing.

\paragraph{Exact likelihood (Table~\ref{tab:sphere_nll}).}
Our method is not a density model in the usual sense: the generative map is the Riemannian gradient of a non-convex
ReLU potential; it is not injective and its image does not cover the whole sphere. This is precisely what makes it
possible to separate modes in a single step, but the density is then only defined on the image of the map. We compute
the exact NLL with the change-of-variables formula, inverting the map by solving the erosion (Section~\ref{supp:metrics});
only 35--47\% of the test points are successfully inverted, and the reported value is computed on these points only.
On this metric, continuous flows, which are designed to provide exact likelihoods and use hundreds of integration
steps, perform better: our method targets sampling quality and cost, not density estimation.
%---------------------------------------------------------------------
\subsection{CIFAR-10}
\begin{figure*}[!tp]
\centering
\includegraphics[width=\textwidth,height=0.7\textheight,keepaspectratio]{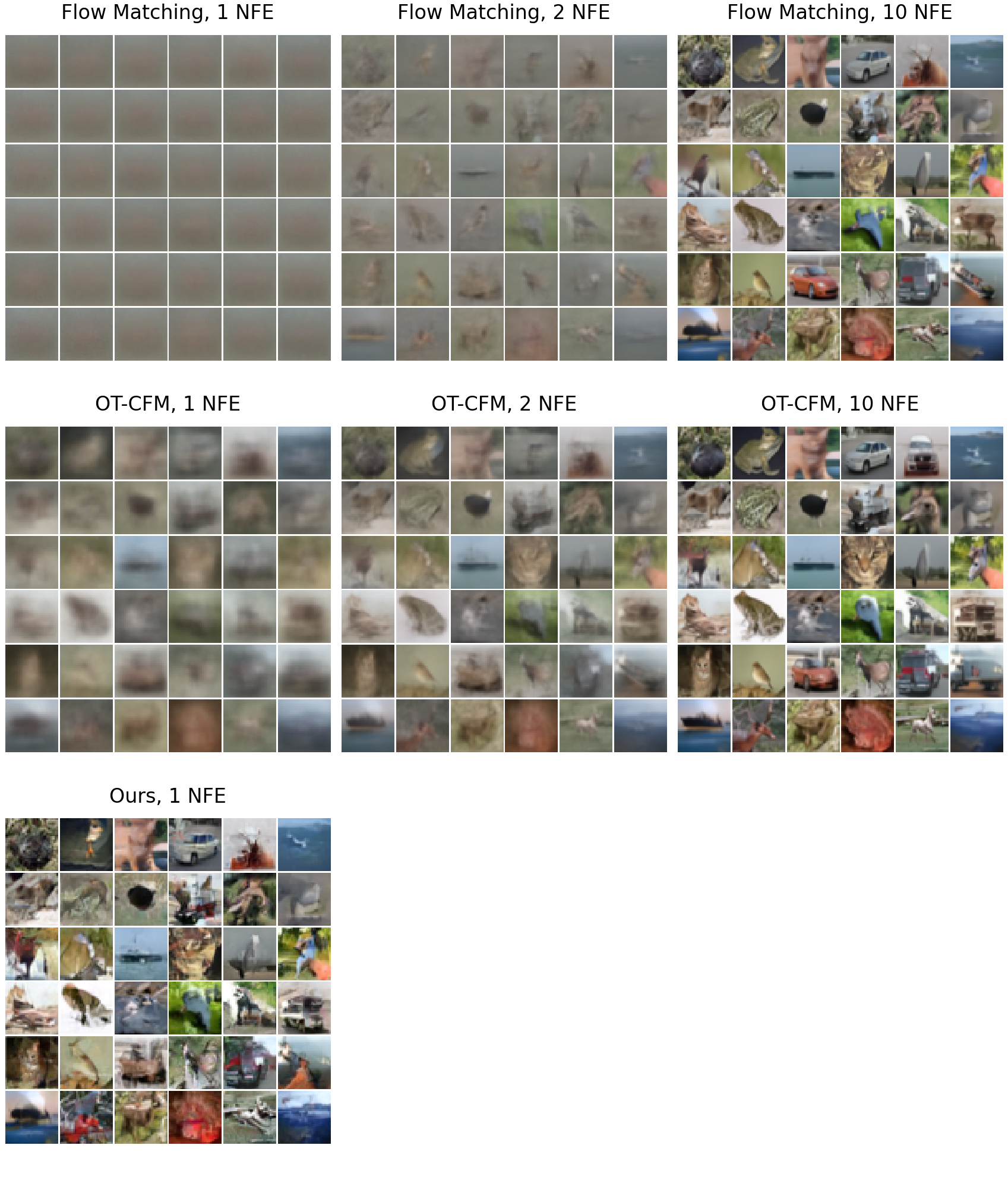}
\caption{CIFAR-10 samples from the same noise for all methods: Flow Matching and OT-CFM with 1, 2 and 10 Euler steps,
MorphoFM with a single step. With one step, Flow Matching collapses to the mean image and OT-CFM produces blurry
shapes, whereas MorphoFM produces recognizable objects (one-step FID-50K: 350.6, 231.5 and 31.3;
Table~\ref{tab:cifar_full}).}
\label{fig:cifar_pas}
\end{figure*}

\begin{figure}[!tp]
\centering
\includegraphics[width=\linewidth]{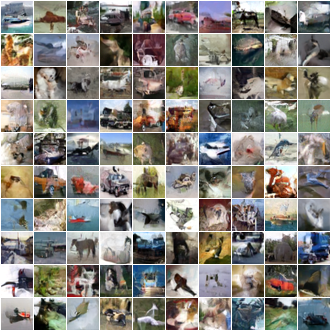}
\caption{Uncurated CIFAR-10 samples of MorphoFM: the first 100 noise samples of seed 2024, each mapped in a single
step, $x_1=x_0+2\nabla f_\theta(x_0)$, by the final model (310k iterations, averaged weights, FID-50K of $31.32$). No
sample is selected or discarded.}
\label{fig:cifar_samples}
\end{figure}

\begin{figure*}[!tp]
\centering
\includegraphics[width=\textwidth]{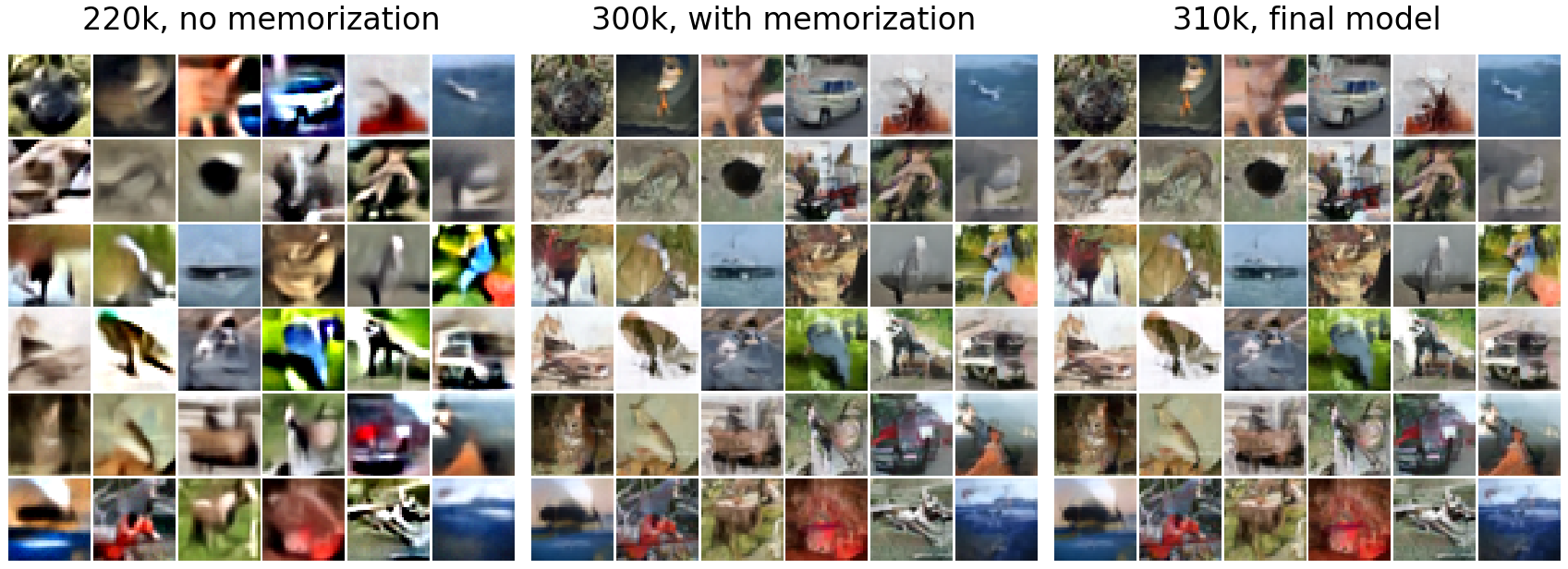}
\caption{CIFAR-10 samples of MorphoFM generated from the same noise, in a single step with averaged weights, at three
checkpoints of the final run: 220k iterations, without memorization (FID-50K of $63.9$); 300k iterations, after 80k
iterations with memorized extremal points and no other change ($32.5$); 310k iterations, final model ($31.3$).
Memorization turns blurry, saturated shapes into objects with textures and details, while each sample keeps its
layout and colors.}
\label{fig:cifar_memo_samples}
\end{figure*}

\begin{figure}[!tp]
\centering
\includegraphics[width=\linewidth]{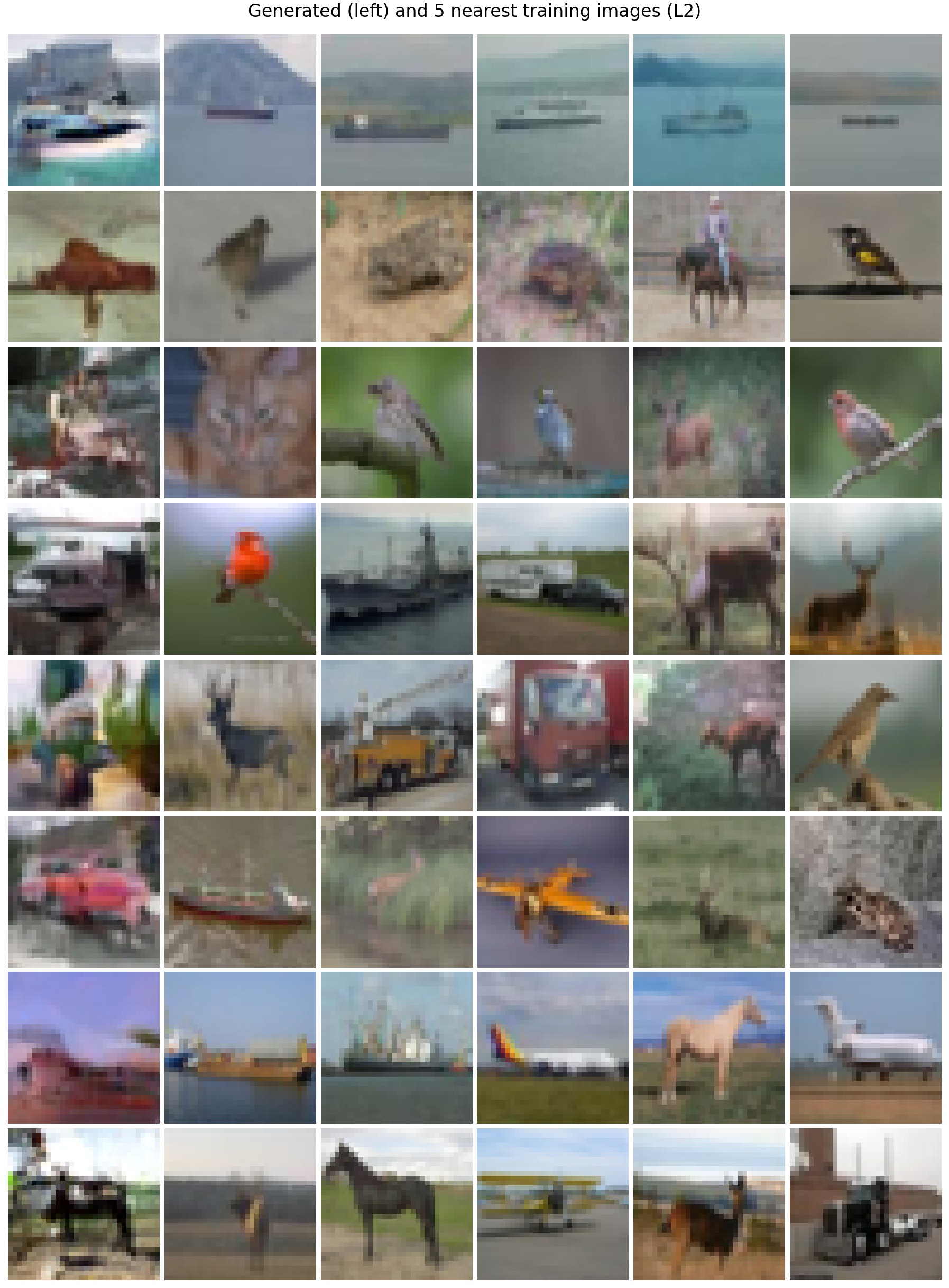}
\caption{Nearest training images of MorphoFM samples. Left column: the first 8 samples of
Figure~\ref{fig:cifar_samples}, not selected. Other columns: the 5 nearest of the 50{,}000 training images in
Euclidean distance on pixels, each training image being compared in its original and horizontally flipped orientation
(flips are used during training). The root-mean-square pixel difference to the nearest image ranges from $0.26$ to
$0.43$ on the $[-1,1]$ scale.}
\label{fig:cifar_nn}
\end{figure}

%---------------------------------------------------------------------
This subsection complements the CIFAR-10 experiment of the main paper. Table~\ref{tab:cifar_full} gathers all one-step
results, our evaluations and published values; Table~\ref{tab:cifar_memo} shows the effect of memorizing extremal
points; Table~\ref{tab:cifar_inner} compares inner solvers; Table~\ref{tab:cifar_cost} reports the costs; and
Table~\ref{tab:cifar_memo0} gives an ablation with memorization from the first iteration. Figure~\ref{fig:cifar_pas}
compares samples at equal numbers of steps; Figure~\ref{fig:cifar_memo_samples} shows samples from the same noise at
three checkpoints of our run; Figure~\ref{fig:cifar_samples} shows uncurated samples of our model, and
Figure~\ref{fig:cifar_nn} their nearest training images.

\begin{table*}[!tp]
\centering
\footnotesize
\setlength{\tabcolsep}{3.2pt}
\caption{CIFAR-10, FID-50K ($\downarrow$). Upper block: our evaluations, all computed with the same code
(\texttt{clean-fid}, \texttt{legacy\_tensorflow} mode, training-set statistics, 50{,}000 images, seed 2024, averaged
weights); Flow Matching, OT-CFM and MeanFlow use the official checkpoints. Lower block: values published by the
authors, with their own protocols and training budgets, given for reference. dopri5: adaptive solver (tolerance
$10^{-5}$), mean number of network evaluations in parentheses. Batch: training batch size.}
\label{tab:cifar_full}
\begin{adjustbox}{max width=\textwidth,center}
\begin{tabular}{l|cc|ccccc}
\toprule
\multirow{2}{*}{Method} & \multirow{2}{*}{Iterations} & \multirow{2}{*}{Batch} & \multicolumn{5}{c}{FID-50K $\downarrow$} \\
\cmidrule(lr){4-8}
 & & & 1 NFE & 2 NFE & 10 NFE & 100 NFE & dopri5 \\
\midrule
\multicolumn{8}{l}{\textit{Our evaluations}} \\
Flow Matching~\cite{lipman2023flow} & 400k & 128 & 350.57 & 165.61 & 14.13 & 4.57 & 3.72 (142) \\
OT-CFM~\cite{tong2024improving} & 400k & 128 & 231.48 & 93.16 & 11.90 & 4.52 & 3.69 (136) \\
MeanFlow~\cite{geng2025mean} & $\approx$700k & 1{,}024 & 2.80 & 2.85 & -- & -- & -- \\
\textbf{MorphoFM (ours)} & 310k & 128 & 31.32 & -- & -- & -- & -- \\
\midrule
\multicolumn{8}{l}{\textit{Published values}} \\
MeanFlow~\cite{geng2025mean} & & & 2.92 & -- & -- & -- & -- \\
2-Rectified Flow~\cite{liu2023flow} & & & 12.21 & -- & -- & -- & -- \\
2-Rectified Flow + distillation~\cite{liu2023flow} & & & 4.85 & -- & -- & -- & -- \\
OTM~\cite{rout2022generative} & & & 21.78 & -- & -- & -- & -- \\
UOTM~\cite{choi2023generative} & & & 2.97 & -- & -- & -- & -- \\
\bottomrule
\end{tabular}
\end{adjustbox}
\end{table*}
\paragraph{One-step results (Table~\ref{tab:cifar_full}).}
With our evaluation code, we recover the published values of the official checkpoints (Section~\ref{supp:baselines}).
The methods of the lower block reach a lower one-step FID with image-specific designs: MeanFlow learns an average
velocity with a U-Net trained with EDM-style augmentation, batches of 1{,}024 and about 700k iterations, \ie, about
18 times more training images than ours; 2-Rectified Flow needs a reflow stage and, for its best value, a distillation
stage; OTM and UOTM train a generator adversarially, the $c$-transform being replaced by a second network
trained by a min-max game. Our method uses neither an adversarial network nor a time-dependent field: its one-step map
is the gradient of the learned potential.
\begin{table}[!htbp]
\centering
\small
\setlength{\tabcolsep}{3pt}
\caption{Effect of memorizing extremal points on CIFAR-10. Top: FID-50K of the final run (Table~\ref{tab:cifar_full}).
Bottom: monitoring FID (10{,}000 images), restarting from the 220k checkpoint either with memorization alone or with a
cosine learning-rate decay alone (control, without memorization).}
\label{tab:cifar_memo}
\resizebox{\linewidth}{!}{%
\begin{tabular}{l|ccc}
\toprule
\multicolumn{4}{c}{FID-50K $\downarrow$ of the final run} \\
\midrule
Iterations & 220k & 300k & 310k \\
Setting & no memorization & + memorization & + $g_\phi$ removed, decay \\
FID-50K & 63.88 & 32.50 & \best{31.32} \\
\bottomrule
\end{tabular}}

\vspace{4pt}
\resizebox{\linewidth}{!}{%
\begin{tabular}{l|ccccc}
\toprule
Iterations & 222.5k & 225k & 230k & 235k & 237.5k \\
\midrule
Memorization only & 62.2 & 58.7 & 50.6 & 44.7 & \best{41.8} \\
Learning-rate decay only & -- & 65.0 & 65.2 & 64.9 & -- \\
\bottomrule
\end{tabular}}
\end{table}
\paragraph{Memorization of extremal points (Table~\ref{tab:cifar_memo}).}
At 220k iterations, the inner problem is poorly solved: within the right basin, $K=5$ L-BFGS steps leave a large gap
to the minimum (Table~\ref{tab:cifar_inner}). Restarting from this checkpoint, memorization alone lowers the monitoring
FID from about 64 to 41.8 in 17.5k iterations, whereas a cosine decay of the learning rate alone leaves it at about 65.
After 800 iterations, the memorized point is the best candidate for 80\% of the images, and the value of the inner
objective after 5 steps decreases from about 289 to 240. Between 220k and 300k iterations, memorization is the only
change, and the FID-50K goes from 63.9 to 32.5; the last 10k iterations, with $g_\phi$ removed and the learning rate
decaying, bring it to 31.3. Figure~\ref{fig:cifar_memo_samples} shows the effect on samples generated from the
same noise.
\begin{table}[!htbp]
\centering
\small
\setlength{\tabcolsep}{4pt}
\caption{Inner solver on CIFAR-10 (checkpoint at 220k iterations, 128 training images, same starting points and same
Armijo line search): mean gap $J_j(\hat y_j)-J_j(y_{\rm ref})$ after $K$ steps, where $y_{\rm ref}$ is obtained with
100 L-BFGS steps. Fixed point: $y\leftarrow x_1-2\nabla f_\theta(y)$, a Newton step for a ReLU MLP.}
\label{tab:cifar_inner}
\begin{tabular}{l|cccc}
\toprule
Solver & $K=5$ & $K=10$ & $K=20$ & $K=40$ \\
\midrule
Fixed point & 102 & 83 & 66 & 53 \\
L-BFGS (ours) & \best{97} & \best{70} & \best{41} & \best{16} \\
\bottomrule
\end{tabular}
\end{table}
\paragraph{Inner solver (Table~\ref{tab:cifar_inner}).}
The CIFAR-10 potential is not piecewise linear (group normalization and attention), and the fixed-point iteration,
which is a Newton step for a ReLU MLP, converges slowly; L-BFGS reaches a better minimum at every budget, about twice as
fast. The batch candidate is essential: starting from $x_1$ alone, or from the prediction of $g_\phi$, leads to a poor
local minimum, whereas a larger pool of noise samples (up to 4{,}096) brings no gain.
\begin{table*}[!tp]
\centering
\footnotesize
\setlength{\tabcolsep}{3.2pt}
\caption{Cost on CIFAR-10 (one NVIDIA RTX~4090 Laptop GPU, all methods measured in the same session). Training: time
of one iteration (batch of 128, forward and backward passes, Adam step, moving average of the weights), and total
training time estimated from it for the iterations actually performed; the predictor $g_\phi$ was among the candidates
during the first 300k iterations (3.91~s per iteration), then removed (3.33~s). Our gradient is accumulated over
micro-batches of 32 to fit in 16~GB; it is identical to the gradient of the full batch, but this slightly penalizes our
timing. Sampling: time for 1{,}000 images (batches of 250, after warm-up), followed by the FID-50K of the main paper;
dopri5: adaptive solver, with the mean number of network evaluations in parentheses.}
\label{tab:cifar_cost}
\begin{adjustbox}{max width=\textwidth,center}
\begin{tabular}{l|ccc|ccccc}
\toprule
\multirow{2}{*}{Method} & \multicolumn{3}{c|}{Training} & \multicolumn{5}{c}{Sampling: time for 1{,}000 images (s) / FID-50K $\downarrow$} \\
\cmidrule(lr){2-4}\cmidrule(lr){5-9}
 & s/iteration & Iterations & Total & 1 NFE & 2 NFE & 10 NFE & 100 NFE & dopri5 \\
\midrule
Flow Matching~\cite{lipman2023flow} & 0.32 & 400k & $\approx$36 h & 0.81 / 350.6 & 1.62 / 165.6 & 8.2 / 14.1 & 82.9 / 4.57 & 119 (142 NFE) / 3.72 \\
OT-CFM~\cite{tong2024improving} & 0.32 & 400k & $\approx$36 h & 0.83 / 231.5 & 1.66 / 93.2 & 8.3 / 11.9 & 82.9 / 4.52 & 115 (136 NFE) / 3.69 \\
\midrule
\textbf{MorphoFM (ours)} & 3.33 (3.91 with $g_\phi$) & 310k & $\approx$335 h & \best{1.51\text{ / }31.3} & -- & -- & -- & -- \\
\bottomrule
\end{tabular}
\end{adjustbox}
\end{table*}
\paragraph{Computational cost (Table~\ref{tab:cifar_cost}).}
One training iteration of our method costs about ten times a flow matching iteration, since it runs $K=5$ L-BFGS steps
on the inner problem, each with a forward and a backward pass through the potential; our total training time is about
nine times larger, for fewer iterations. Sampling is the converse. One step of our method costs about twice an Euler
step of flow matching, because it evaluates the gradient of the potential, but a single step suffices. At about the same
sampling time (2 NFE), Flow Matching and OT-CFM reach an FID of $165.6$ and $93.2$, versus $31.3$ for our method; they
need about 10 steps, \ie, about five times our sampling time, to do better, and 55 to 79 times our sampling time for
their best quality. Our numbers of dopri5 evaluations (142 and 136) agree with those reported by the authors (142 and
134).

\begin{table}[!htbp]
\centering
\small
\setlength{\tabcolsep}{3pt}
\caption{Ablation: memorization from the first iteration (monitoring FID on 10{,}000 images, $\downarrow$). This run uses
one GPU, so the batch candidate is taken among the 128 noise samples of the batch, versus about 43 per GPU for the
reference run (3 GPUs, memorization from 220k on). Reference values at the same iterations are given when available.}
\label{tab:cifar_memo0}
\resizebox{\linewidth}{!}{%
\begin{tabular}{l|cccccccccc}
\toprule
Iterations & 10k & 20k & 30k & 40k & 50k & 60k & 70k & 80k & 90k & 100k \\
\midrule
Memorization from 0 & 447.5 & 443.2 & 289.2 & 295.0 & 152.1 & 97.6 & 84.6 & 71.5 & 55.9 & \best{48.3} \\
Reference run & -- & -- & -- & -- & -- & -- & -- & -- & -- & 78.1 \\
\bottomrule
\end{tabular}}
\end{table}
\paragraph{Memorization from the first iteration (Table~\ref{tab:cifar_memo0}).}
With memorization from the first iteration, the monitoring FID reaches 48.3 after 100k iterations, versus 78.1 for our
reference run at the same stage. The two runs also differ by the size of the batch candidate pool (128 versus about 43
noise samples), but a larger pool does not improve the inner problem (Table~\ref{tab:cifar_inner}); this confirms that
the accuracy of the inner problem, rather than the number of iterations, limits the quality.

\paragraph{Memorizing extremal points does not memorize images (Figure~\ref{fig:cifar_nn}).}
The memory stores, for each training image, a point of the source space that serves as a starting point for the inner
solver; it stores no image, and sampling only evaluates the gradient of the potential. The nearest training images of
our samples share their global colors and layout, but none is a copy of a training image.
%=====================================================================
\section{Ablations in the Euclidean plane}\label{supp:ablations}
%=====================================================================
This section reports the ablations in $\R^2$: Table~\ref{tab:supp_abl_K} studies the number $K$ of local-solver
steps, Table~\ref{tab:supp_abl_candidats} the starting candidates, Table~\ref{tab:supp_abl_activation} the activation of
the potential, Table~\ref{tab:supp_abl_temps} the baselines at equal training time, Table~\ref{tab:supp_abl_poids} the
training objective of the potential, Table~\ref{tab:supp_abl_multipas} multi-step generation, and
Table~\ref{tab:supp_abl_k} the exponent $k$. All ablations use the protocol of the 2D benchmark (MLP $3\times256$, 5{,}000 iterations, batches of 512, 5 seeds, MMD
between 2{,}000 generated samples and 2{,}000 test samples). The reference variant is exactly our method ($K=5$,
candidates $\{x_1,\text{batch},g_\phi\}$, ReLU potential); the variants are only used for analysis, and the model is
not modified. The ablations come from a single run: comparisons between variants are paired (same seeds, same sampling
noise). Absolute values differ slightly from those of the main table, which comes from a separate run; the difference
is of the order of the standard deviation across seeds. The best value is highlighted only where the
difference is significant.
\paragraph{Number $K$ of local-solver steps (Table~\ref{tab:supp_abl_K}).}
Without refinement ($K=0$), the best candidate is far from the true minimum for 54\% of the points; with $K=5$, this
fraction drops to 4.6\%, and $K=10$ or $K=20$ bring almost nothing more (3.4\%) at a two to four times higher cost. The
MMD follows the same trend, but more weakly: $K=0$ is worse than $K=5$ for 19 seeds out of 20, and the quality no
longer changes from $K=2$ on. $K=5$ is therefore an accuracy--cost trade-off, and the final quality is not very
sensitive to this choice.
\begin{table*}[!tp]
\centering
\footnotesize
\setlength{\tabcolsep}{3.2pt}
\caption{Number $K$ of local-solver steps in the inner problem (MMD $\downarrow$, one-step generation). Inner gap: mean of $\phi(\hat y_K)-\phi(y_{\rm ref})$ on test points at the learned potential, where $y_{\rm ref}$ uses 2{,}048 starting points and 100 L-BFGS steps; fail: fraction of points with gap $>10^{-3}$. Mean $\pm$ std over 5 seeds, 2{,}000 generated vs. held-out samples.}
\label{tab:supp_abl_K}
\begin{adjustbox}{max width=\textwidth,center}
\begin{tabular}{l|cccc|cc|c}
\toprule
\multirow{2}{*}{Variant} & \multicolumn{4}{c|}{MMD $\downarrow$} & \multicolumn{2}{c|}{Inner problem} & \multirow{2}{*}{Training (s)} \\
\cmidrule(lr){2-5}\cmidrule(lr){6-7}
 & 8 Gaussians & Two Moons & Checkerboard & Spirals & Inner gap $\downarrow$ & Fail (\%) $\downarrow$ & \\
\midrule
$K=0$ & $0.0736 \pm 0.0119$ & $0.0733 \pm 0.0111$ & $0.0764 \pm 0.0057$ & $0.0650 \pm 0.0032$ & 3.3e-03 & 53.6 & 18 \\
$K=1$ & $0.0719 \pm 0.0118$ & $0.0723 \pm 0.0113$ & $0.0741 \pm 0.0061$ & $0.0648 \pm 0.0036$ & 1.5e-03 & 21.7 & 28 \\
$K=2$ & $0.0705 \pm 0.0122$ & $0.0717 \pm 0.0115$ & $0.0742 \pm 0.0064$ & $0.0648 \pm 0.0036$ & 6.8e-04 & 10.5 & 40 \\
$K=5$ (ours) & $0.0703 \pm 0.0123$ & $0.0717 \pm 0.0112$ & $0.0739 \pm 0.0063$ & $0.0646 \pm 0.0036$ & 2.5e-04 & 4.6 & 78 \\
$K=10$ & $0.0703 \pm 0.0118$ & $0.0716 \pm 0.0113$ & $0.0741 \pm 0.0062$ & $0.0646 \pm 0.0037$ & 1.6e-04 & 3.4 & 151 \\
$K=20$ & $0.0703 \pm 0.0121$ & $0.0715 \pm 0.0113$ & $0.0742 \pm 0.0062$ & $0.0649 \pm 0.0035$ & 1.6e-04 & 3.4 & 304 \\
\bottomrule
\end{tabular}
\end{adjustbox}
\end{table*}
\paragraph{Starting candidates (Table~\ref{tab:supp_abl_candidats}).}
All variants remain within $0.0015$ of the reference. Removing the batch candidate slightly degrades the MMD
(8 Gaussians and Spirals: 5 seeds out of 5); removing the predictor $g_\phi$ has no measurable effect in dimension 2,
where the best point of the batch already provides a good starting point. On CIFAR-10, memorized extremal points make it
unnecessary (Section~\ref{supp:details}).
\begin{table*}[!tp]
\centering
\footnotesize
\setlength{\tabcolsep}{3.2pt}
\caption{Starting candidates of the inner problem (MMD $\downarrow$, $K=5$, one-step generation). ``Batch'': best point of the current batch of source samples; $g$: learned predictor. Mean $\pm$ std over 5 seeds, 2{,}000 generated vs. held-out samples.}
\label{tab:supp_abl_candidats}
\begin{adjustbox}{max width=\textwidth,center}
\begin{tabular}{l|cccc|c}
\toprule
\multirow{2}{*}{Variant} & \multicolumn{4}{c|}{MMD $\downarrow$} & \multirow{2}{*}{Training (s)} \\
\cmidrule(lr){2-5}
 & 8 Gaussians & Two Moons & Checkerboard & Spirals & \\
\midrule
all (ours) & $0.0703 \pm 0.0123$ & $0.0717 \pm 0.0112$ & $0.0739 \pm 0.0063$ & $0.0646 \pm 0.0036$ & 80 \\
w/o batch & $0.0718 \pm 0.0121$ & $0.0717 \pm 0.0112$ & $0.0740 \pm 0.0062$ & $0.0653 \pm 0.0035$ & 76 \\
w/o predictor $g$ & $0.0707 \pm 0.0123$ & $0.0718 \pm 0.0113$ & $0.0740 \pm 0.0064$ & $0.0643 \pm 0.0038$ & 73 \\
w/o $x_1$ & $0.0705 \pm 0.0121$ & $0.0717 \pm 0.0113$ & $0.0741 \pm 0.0062$ & $0.0646 \pm 0.0035$ & 77 \\
predictor $g$ only & $0.0717 \pm 0.0123$ & $0.0717 \pm 0.0114$ & $0.0741 \pm 0.0062$ & $0.0654 \pm 0.0037$ & 74 \\
batch only & $0.0706 \pm 0.0122$ & $0.0717 \pm 0.0112$ & $0.0741 \pm 0.0063$ & $0.0644 \pm 0.0037$ & 73 \\
\bottomrule
\end{tabular}
\end{adjustbox}
\end{table*}
\paragraph{Activation of the potential (Table~\ref{tab:supp_abl_activation}).}
This is the clearest effect: the ReLU potential outperforms SiLU and Softplus on all 20 (dataset, seed) pairs, with a
mean gap of $0.022$ (SiLU) and $0.057$ (Softplus), and up to $0.085$ on 8 Gaussians. This is what the topological
obstruction predicts (Section~\ref{supp:obstr}): a smooth potential has a continuous gradient, hence a continuous
generation map, which cannot separate the modes in one step.
\begin{table*}[!tp]
\centering
\footnotesize
\setlength{\tabcolsep}{3.2pt}
\caption{Activation of the potential network (MMD $\downarrow$, one-step generation). Smooth activations give a continuous transport map, which cannot separate disconnected modes. Mean $\pm$ std over 5 seeds, 2{,}000 generated vs. held-out samples. Best value per column highlighted.}
\label{tab:supp_abl_activation}
\begin{adjustbox}{max width=\textwidth,center}
\begin{tabular}{l|cccc|c}
\toprule
\multirow{2}{*}{Variant} & \multicolumn{4}{c|}{MMD $\downarrow$} & \multirow{2}{*}{Training (s)} \\
\cmidrule(lr){2-5}
 & 8 Gaussians & Two Moons & Checkerboard & Spirals & \\
\midrule
ReLU (ours) & \best{0.0703 \pm 0.0123} & \best{0.0717 \pm 0.0112} & \best{0.0739 \pm 0.0063} & \best{0.0646 \pm 0.0036} & 79 \\
SiLU & $0.1216 \pm 0.0064$ & $0.0823 \pm 0.0120$ & $0.0844 \pm 0.0049$ & $0.0819 \pm 0.0026$ & 70 \\
Softplus & $0.1551 \pm 0.0076$ & $0.1134 \pm 0.0112$ & $0.1397 \pm 0.0036$ & $0.0986 \pm 0.0043$ & 69 \\
\bottomrule
\end{tabular}
\end{adjustbox}
\end{table*}
\paragraph{Baselines at equal training time (Table~\ref{tab:supp_abl_temps}).}
One iteration of our method is more expensive than one iteration of Flow Matching, because of the inner problem. At
equal training time, Flow Matching performs 58{,}000 to 69{,}000 iterations; OT-CFM, whose iteration is more expensive
than ours, keeps the 5{,}000 iterations of the main benchmark. Our one-step method remains better than OT-CFM at 1, 10
and 100 steps (20 seeds out of 20) and than Flow Matching at 1 and 10 steps (20 out of 20). Flow Matching integrated
with 100 steps, however, reaches a quality comparable to ours: it is slightly better on Checkerboard and 8 Gaussians,
and slightly worse on Two Moons and Spirals. At equal training time, our one-step generation therefore matches the
quality of Flow Matching with 100 times fewer network evaluations at sampling.
\begin{table*}[!tp]
\centering
\footnotesize
\setlength{\tabcolsep}{3.2pt}
\caption{Baselines trained for the same wall-clock time as our method (MMD $\downarrow$). The number of training iterations of each baseline is calibrated on the same GPU and in the same run as our method, and is never smaller than in the main benchmark (5{,}000). Absolute times differ from those of Table~\ref{tab:2d_cost}, which come from a different run: absolute times on this laptop GPU vary between runs (our method takes 79\,s here and 194\,s there, and Flow Matching is also about twice as fast per iteration here); only times measured within the same table should be compared. Mean $\pm$ std over 5 seeds, 2{,}000 generated vs. held-out samples. Best value per column highlighted; best one-step value underlined.}
\label{tab:supp_abl_temps}
\begin{adjustbox}{max width=\textwidth,center}
\begin{tabular}{l|c|c|cccc|c}
\toprule
\multirow{2}{*}{Method} & \multirow{2}{*}{NFE} & \multirow{2}{*}{Iterations} & \multicolumn{4}{c|}{MMD $\downarrow$} & \multirow{2}{*}{Training (s)} \\
\cmidrule(lr){4-7}
 & & & 8 Gaussians & Two Moons & Checkerboard & Spirals & \\
\midrule
Flow Matching (same time) & 1 & 64{,}217 & $2.0121 \pm 0.0229$ & $1.8153 \pm 0.0082$ & $1.7912 \pm 0.0033$ & $1.7807 \pm 0.0024$ & 76 \\
 & 10 & 64{,}217 & $0.1049 \pm 0.0064$ & $0.1454 \pm 0.0084$ & $0.1133 \pm 0.0070$ & $0.1337 \pm 0.0083$ & 76 \\
 & 100 & 64{,}217 & \best{0.0688 \pm 0.0107} & $0.0731 \pm 0.0103$ & \best{0.0687 \pm 0.0055} & $0.0650 \pm 0.0040$ & 76 \\
\midrule
OT-CFM (same time) & 1 & 5{,}000 & $0.1588 \pm 0.0042$ & $0.1061 \pm 0.0102$ & $0.1030 \pm 0.0071$ & $0.0917 \pm 0.0052$ & 98 \\
 & 10 & 5{,}000 & $0.0810 \pm 0.0097$ & $0.0789 \pm 0.0108$ & $0.0835 \pm 0.0062$ & $0.0817 \pm 0.0046$ & 98 \\
 & 100 & 5{,}000 & $0.0736 \pm 0.0109$ & $0.0776 \pm 0.0107$ & $0.0826 \pm 0.0062$ & $0.0811 \pm 0.0046$ & 98 \\
\midrule
\textbf{MorphoFM (ours)} & 1 & 5{,}000 & $\underline{0.0703 \pm 0.0123}$ & \best{\underline{0.0717 \pm 0.0112}} & $\underline{0.0739 \pm 0.0063}$ & \best{\underline{0.0646 \pm 0.0036}} & 79 \\
\bottomrule
\end{tabular}
\end{adjustbox}
\end{table*}
\paragraph{Training objective of the potential (Table~\ref{tab:supp_abl_poids}).}
With the independent coupling, a uniform weighting $w$ on $(0,1]$ strongly degrades generation (MMD of $0.27$ to
$0.35$, versus $0.065$ to $0.074$), as predicted by the incompatibility of the terms $t<1$
(Section~\ref{supp:incompat}). The substituted CFM loss, where the morphological velocity is regressed onto the
conditional target $x_1-x_0$, is also clearly worse (MMD of $0.10$ to $0.21$). In both cases, the gap with our loss is
observed on all 20 (dataset, seed) pairs. Among the objectives tested, only the $t=1$ term, that is, the semi-dual
Kantorovich formulation, yields correct generation, which justifies the construction of the loss.
\begin{table*}[!tp]
\centering
\footnotesize
\setlength{\tabcolsep}{3.2pt}
\caption{Training objective of the potential (MMD $\downarrow$, one-step generation, independent coupling). Mean $\pm$ std over 5 seeds, 2{,}000 generated vs. held-out samples. Best value per column highlighted.}
\label{tab:supp_abl_poids}
\begin{adjustbox}{max width=\textwidth,center}
\begin{tabular}{l|cccc|c}
\toprule
\multirow{2}{*}{Variant} & \multicolumn{4}{c|}{MMD $\downarrow$} & \multirow{2}{*}{Training (s)} \\
\cmidrule(lr){2-5}
 & 8 Gaussians & Two Moons & Checkerboard & Spirals & \\
\midrule
$w=\delta_1$ (ours) & \best{0.0703 \pm 0.0123} & \best{0.0717 \pm 0.0112} & \best{0.0739 \pm 0.0063} & \best{0.0646 \pm 0.0036} & 77 \\
$w$ uniform on $(0,1]$ & $0.3444 \pm 0.0038$ & $0.3453 \pm 0.0038$ & $0.2714 \pm 0.0048$ & $0.3333 \pm 0.0053$ & 77 \\
CFM loss & $0.1038 \pm 0.0075$ & $0.2089 \pm 0.0086$ & $0.1906 \pm 0.0055$ & $0.1404 \pm 0.0057$ & 81 \\
\bottomrule
\end{tabular}
\end{adjustbox}
\end{table*}
\paragraph{Multi-step generation (Table~\ref{tab:supp_abl_multipas}).}
With the learned potential, integrating the morphological field with 2, 5 or 10 Euler steps, recomputing the extremal
point at each step, gives the same samples as a single step: the MMD changes by less than $4\cdot10^{-4}$ on all
seeds. This is the numerical verification of the exact Euler scheme (Theorem~\ref{thm:gen}(v)): the learned
trajectories are indeed straight. Several steps therefore bring nothing, and cost 100 to 1{,}000 times more, since each
step solves an inner problem.
\begin{table*}[!tp]
\centering
\footnotesize
\setlength{\tabcolsep}{3.2pt}
\caption{Multi-step generation with our trained potential: geodesic Euler with $v_t(x)=(x-y^\ast_t(x))/t$, the extremal point $y^\ast_t$ being recomputed at each step (candidates and 20 L-BFGS steps). MMD $\downarrow$, mean $\pm$ std over 5 seeds; sampling time for 2{,}000 samples. Exact Euler (Theorem~\ref{thm:gen}(v)) predicts identical samples for any number of steps.}
\label{tab:supp_abl_multipas}
\begin{adjustbox}{max width=\textwidth,center}
\begin{tabular}{c|cccc|c}
\toprule
\multirow{2}{*}{NFE} & \multicolumn{4}{c|}{MMD $\downarrow$} & \multirow{2}{*}{Sampling time (ms) $\downarrow$} \\
\cmidrule(lr){2-5}
 & 8 Gaussians & Two Moons & Checkerboard & Spirals & \\
\midrule
1 (ours) & $0.0703 \pm 0.0123$ & $0.0717 \pm 0.0112$ & $0.0739 \pm 0.0063$ & $0.0646 \pm 0.0036$ & 0.5 \\
2 & $0.0703 \pm 0.0122$ & $0.0717 \pm 0.0113$ & $0.0740 \pm 0.0063$ & $0.0647 \pm 0.0035$ & 62.8 \\
5 & $0.0702 \pm 0.0122$ & $0.0717 \pm 0.0113$ & $0.0740 \pm 0.0063$ & $0.0647 \pm 0.0035$ & 244.6 \\
10 & $0.0703 \pm 0.0123$ & $0.0717 \pm 0.0113$ & $0.0740 \pm 0.0063$ & $0.0647 \pm 0.0036$ & 544.4 \\
\bottomrule
\end{tabular}
\end{adjustbox}
\end{table*}
\paragraph{Exponent $k$ (Table~\ref{tab:supp_abl_k}).}
The exponents $k=1.5$, $2$ and $3$ give equivalent results, within the variability across seeds (mean difference
below $5\cdot10^{-4}$). The method is therefore not sensitive to this choice; we use $k=2$, for which the cost is
$d^2/4$ and generation reads $x_0+2\nabla f(x_0)$.
\begin{table*}[!tp]
\centering
\footnotesize
\setlength{\tabcolsep}{3.2pt}
\caption{Exponent $k$ of the Hamiltonian $H(p)=|p|^k$ (cost $c_k|x-y|^{k^\ast}$, MMD $\downarrow$, one-step generation). Mean $\pm$ std over 5 seeds, 2{,}000 generated vs. held-out samples. All differences are within one standard deviation.}
\label{tab:supp_abl_k}
\begin{adjustbox}{max width=\textwidth,center}
\begin{tabular}{l|cccc|c}
\toprule
\multirow{2}{*}{Variant} & \multicolumn{4}{c|}{MMD $\downarrow$} & \multirow{2}{*}{Training (s)} \\
\cmidrule(lr){2-5}
 & 8 Gaussians & Two Moons & Checkerboard & Spirals & \\
\midrule
$k=1.5$ & $0.0701 \pm 0.0119$ & $0.0717 \pm 0.0096$ & $0.0751 \pm 0.0073$ & $0.0650 \pm 0.0037$ & 89 \\
$k=2$ (ours) & $0.0703 \pm 0.0123$ & $0.0717 \pm 0.0112$ & $0.0739 \pm 0.0063$ & $0.0646 \pm 0.0036$ & 79 \\
$k=3$ & $0.0707 \pm 0.0120$ & $0.0708 \pm 0.0118$ & $0.0729 \pm 0.0055$ & $0.0644 \pm 0.0042$ & 87 \\
\bottomrule
\end{tabular}
\end{adjustbox}
\end{table*}
%=====================================================================
\section{Additional theoretical results}\label{supp:compl}
%=====================================================================
%---------------------------------------------------------------------
\subsection{Incompatibility of intermediate-time terms with the independent coupling}\label{supp:incompat}
%---------------------------------------------------------------------
For $t\in(0,1)$, the term $\E_{p_0}f-\E_{p_t}u_f(t,\cdot)$ is, by the same argument as in Theorem~\ref{thm:dual}, the
Kantorovich dual of the transport from $p_0$ to the distribution $p_t$ of $x_t$ for the cost $b_t$. With the
independent coupling, $p_t$ is not McCann's interpolation between $p_0$ and $q$, and the optimal potentials change
with $t$.
\begin{proposition}\label{prop:incompat}
In the Gaussian setting of Proposition~\ref{prop:contre}, with $k=2$ and $\sigma\neq1$:
\begin{enumerate}[label=(\roman*)]
\item the optimal potential of the term $t$ is $f_t(x)=\frac{s_t-1}{4t}x^2$, with $s_t=\sqrt{(1-t)^2+t^2\sigma^2}$, and
$f_t\neq f_1$ for all $t\in(0,1)$;
\item within the family $f_a(x)=ax^2$, if $w$ puts mass on $(0,1)$, the minimizer of $\Lcal_w$ satisfies
$a<a_1=(\sigma-1)/4$: one-step generation $x\mapsto x+2f_a'(x)=(1+4a)x$ then has standard deviation $1+4a<\sigma$, i.e., it is too
concentrated.
\end{enumerate}
\end{proposition}
\begin{proof}
For $a>-1/(4t)$, $y\mapsto ay^2+(x-y)^2/(4t)$ is minimal at $y=x/(1+4at)$, with value $ax^2/(1+4at)$. The term $t$,
as a function of $a$, is therefore
\[
\begin{aligned}
\Lambda_t(a)&=a-\frac{a\,s_t^2}{1+4at},\\
\Lambda_t'(a)&=1-\frac{s_t^2}{(1+4at)^2},\\
\Lambda_t''(a)&=\frac{8t\,s_t^2}{(1+4at)^3}>0.
\end{aligned}
\]
$\Lambda_t$ is strictly convex, with minimizer given by $1+4at=s_t$, which proves (i) within this family. This minimizer
is also the unrestricted optimal potential: $x\mapsto x+2tf_t'(x)=s_tx$ is the monotone optimal transport map from
$\mathcal N(0,1)$ to $\mathcal N(0,s_t^2)$. The identity $f_t=f_1$ requires $s_t=1+t(\sigma-1)=:m_t$; but
$m_t^2-s_t^2=2t(1-t)\sigma>0$ for $t\in(0,1)$, so $s_t<m_t$. For (ii), at $a=a_1$, $1+4a_1t=m_t$ and
$\Lambda_t'(a_1)=1-s_t^2/m_t^2>0$ for all $t\in(0,1)$, whereas $\Lambda_1'(a_1)=0$. The derivative of
$\Lcal_w=\int w\Lambda_t\dd t$ at $a_1$ is therefore strictly positive as soon as $w$ puts mass on $(0,1)$ and, by
strict convexity, the minimizer is strictly smaller than $a_1$.
\end{proof}
\begin{remark}
This is the reason for the choice $w=\delta_1$: the term $t=1$ alone is well posed (Theorem~\ref{thm:dual}), and
Theorem~\ref{thm:gen} shows that it suffices, since the whole dynamics for $t\in(0,1)$ follows from it.
\end{remark}
%---------------------------------------------------------------------
\subsection{Convexity of the inner problem}\label{supp:interieur}
%---------------------------------------------------------------------
\begin{proposition}[Inner problem convex at optimum]\label{prop:k2} % (Case $k=2$ in $\mathbb{R}^n$)
Let $f^\ast$ be $c$-convex for $c(x,y)=|x-y|^2/4$ (for instance, the $c$-convex minimizer of Theorem~\ref{thm:dual}
for $k=2$ in $\R^n$). Then $\psi(x)=\frac12|x|^2+2f^\ast(x)$ is convex (Brenier's condition), and for all $x$: the
function $y\mapsto f^\ast(y)+|x-y|^2/(4t)$ is $\frac{1-t}{2t}$-strongly convex for $t\in(0,1)$, and convex for $t=1$.
\end{proposition}
\begin{proof}
By $c$-convexity, $f^\ast(x)=\sup_y\{\zeta(y)-|x-y|^2/4\}$ for some function $\zeta$, hence
\[
f^\ast(x)+\frac{|x|^2}4=\sup_y\Big\{\frac{x\cdot y}2-\frac{|y|^2}4+\zeta(y)\Big\}
\]
is a supremum of affine functions of $x$, hence convex, and so is $\psi=2\big(f^\ast+\frac{|\cdot|^2}4\big)$. Then
\[
\begin{aligned}
f^\ast(y)+\frac{|x-y|^2}{4t}
&=\frac{\psi(y)}2+\frac{1-t}{4t}|y|^2\\
&\quad-\frac{x\cdot y}{2t}+\frac{|x|^2}{4t},
\end{aligned}
\]
the sum of a convex function, a $\frac{1-t}{2t}$-strongly convex function, and a function that is affine in $y$.
\end{proof}
\begin{remark}
At the optimal potential, the inner problem at $t=1$ is therefore convex, and a quasi-Newton solver converges to its
global minimum. For a learned potential far from the optimum, $\psi$ is not convex and the inner problem may have
several local minima, a local solver converging within the basin where it starts; this motivates combining several
starting candidates ($x_1$ itself, the best point of the batch and, when available, the memorized extremal point).
\end{remark}
\begin{table}[!htbp]
\centering
\small
\caption{Potential architecture (8 Gaussians embedded in $\mathbb{R}^{10}$, 5{,}000 iterations, inner solver: candidates + L-BFGS with $K=5$, one-step generation; MMD $\downarrow$, mean $\pm$ std over 3 seeds). ``Max-of-$M$-heads'': SiLU trunk $3\times256$ with $M$ scalar heads, $f=\max_j h_j$. Active heads: number of heads that attain the maximum on 2{,}000 source samples at the end of training. Best value highlighted.}
\label{tab:supp_maxheads}
\begin{adjustbox}{max width=\columnwidth,center}
\begin{tabular}{l|c|c}
\toprule
Potential & MMD $\downarrow$ & Active heads \\
\midrule
ReLU MLP $3\times256$ (ours) & \best{0.075 \pm 0.009} & -- \\
Max of 8 heads & $0.107 \pm 0.010$ & 2--3 \\
Max of 16 heads & $0.104 \pm 0.012$ & 2--4 \\
\bottomrule
\end{tabular}
\end{adjustbox}
\end{table}
%---------------------------------------------------------------------
\subsection{Topological obstruction and potential architecture}\label{supp:obstr}
%---------------------------------------------------------------------
\begin{proposition}[Topological obstruction]\label{prop:obstr}
Let $\M$ be connected, $p_0$ a distribution with a continuous, strictly positive density on $\M$ (for instance
$\mathcal N(0,I)$ in $\R^n$ or the uniform distribution on $\mathbb S^2$), and $q$ a distribution whose support is the
union of two disjoint nonempty closed sets $A$ and $B$ of positive mass. Then no continuous map $T:\M\to\M$ satisfies
$T_\#p_0=q$.
\end{proposition}
\begin{proof}
Let $T$ be continuous with $T_\#p_0=q$. If $T(x)\notin\supp q$ for some $x$, then, since $\supp q$ is closed, an open
neighborhood $U$ of $x$ is mapped outside $\supp q$, and $q(\M\setminus\supp q)\ge p_0(U)>0$: a contradiction. Hence
$T(\M)\subset A\cup B$. Since $\M$ is connected, so is $T(\M)$, which is therefore contained in $A$ or in $B$, say
$A$. Then $q(B)=p_0(T^{-1}(B))=0$, which contradicts $q(B)>0$.
\end{proof}
\begin{corollary}\label{cor:discontinu}
If $\supp q$ is not connected, no map that coincides $p_0$-almost everywhere with the map
$T=\exp(\nabla H(\nabla_\g f^\ast))$ of Theorem~\ref{thm:gen} is continuous; since $\exp$ and $\nabla H$ are continuous,
$f^\ast$ is not of class $C^1$. A potential of class $C^1$ (a network
with smooth activations) therefore cannot achieve exact one-step generation. This obstruction does not apply to
Lipschitz potentials whose gradient can jump, such as ReLU networks; we do not claim that a ReLU network represents
$f^\ast$ exactly, and the ablation of Table~\ref{tab:supp_abl_activation} assesses this choice empirically.
\end{corollary}
\begin{remark}
The flow of a field that is Lipschitz in space is a homeomorphism, and the same obstruction holds for any ODE with a
Lipschitz field, including flow matching: the latter moves the stiffness to time (the learned field becomes stiff near
$t=1$), whereas one-step generation places it in space (a gradient that jumps between modes). For a target with
well-separated modes but connected support (a Gaussian mixture), the optimal map is continuous but very steep between
the modes, and the same conclusions hold qualitatively.
\end{remark}
\paragraph{A tested alternative: the ``max-of-heads'' potential.}
For $k=2$, if $\frac12|x|^2+2h_j$ is convex for every $j$, then so is $\frac12|x|^2+2\max_jh_j$ (a maximum of convex
functions is convex). This suggests a potential that is itself morphological, $f_\theta=\max_{j\le M}h_{\theta,j}$,
where the maximum (a dilation) creates the jumps of the gradient and each smooth head ensures the contraction within its
mode. We compared it with the ReLU potential (Table~\ref{tab:supp_maxheads}): it performs worse, and only 2 to 4 heads
remain active at the end of training. We therefore keep the ReLU potential.
% Protocol of diag_maxheads.py (results_maxheads/results.csv): 8 Gaussians embedded in dimension 10,
% 3 seeds, 5,000 iterations, g + L-BFGS K = 5, MMD of experiment_8gaussians.mmd_rbf on 2,000 points.
% This protocol differs from that of the main 2D table: do not compare values across tables.

% =====================================================================

%\input{sec/2_formatting}
%\input{sec/3_finalcopy}
%\input{sec/X_suppl}
{
    \small
    \bibliographystyle{ieeenat_fullname}
    \bibliography{main,biblio}
}

%%%%%%%%% SUPPLEMENTARY MATERIAL + ITS BIBLIOGRAPHY
%\clearpage
%\setcounter{page}{1}
%\maketitlesupplementary

% Sections become A, B, C...; tables/figures/equations become S1, S2...
%\appendix
%\setcounter{table}{0}
%\setcounter{figure}{0}
%\setcounter{equation}{0}
%\renewcommand{\thetable}{S\arabic{table}}
%\renewcommand{\thefigure}{S\arabic{figure}}
%\renewcommand{\theequation}{S\arabic{equation}}

%\begin{bibunit}
%\input{sec/sm}

%{
%    \small
%    \putbib
%}
%\end{bibunit}

\end{document}